%% file: Bandit_Labor_Training.tex
\documentclass[mnsc]{informs3}
\usepackage{booktabs} 
\usepackage[ruled,linesnumbered]{algorithm2e} 

\SetAlFnt{\small}
\SetAlCapFnt{\small}
\SetAlCapNameFnt{\small}
\SetAlCapHSkip{0pt}
\IncMargin{-\parindent}

\usepackage{titlesec}

\newskip\smallskipamount \smallskipamount=1pt plus 1pt minus 1pt

\usepackage[numbers]{natbib}
\def\bibfont{\small}

\DoubleSpacedXI

\usepackage{caption}
\usepackage{subcaption}
\usepackage{amsfonts,bbm}
\usepackage {booktabs}
\usepackage{mathrsfs}
\usepackage{mathtools}
\usepackage{verbatim}
\usepackage{multirow}
\usepackage{nicefrac}       
\usepackage{amsmath, amssymb}

\usepackage{enumitem,kantlipsum}
\usepackage[ruled]{algorithm2e} 
\usepackage{float}
\usepackage{graphicx} 
\newcommand{\E}{\textup{E}}
\newcommand{\ucb}{\textup{UCB}}

\newcommand{\lcb}{\textup{LCB}}
\newcommand{\p}{\textup{P}}

\newcommand{\cR}{\mathcal{R}}

\newcommand{\bnu}{\boldsymbol{\nu}}
\newcommand{\bmu}{\boldsymbol{\mu}}
\newcommand{\PSR}{\mathcal{\textup{Reg}}}
\newcommand{\PPSR}{\mathcal{\textup{Reg}}}
\newcommand{\rew}{\mathsf{R}}
\newcommand{\rewR}{\mathsf{R}}
\newcommand{\triangleqq}{\overset{\Delta}{=}}

\newcommand{\boldi}{\textbf{I}} 
\newcommand{\bolde}{\textbf{E}}
\newcommand{\bx}{\textbf{X}}
\newcommand{\obj}{\Gamma^m}

\newcommand{\bestm}{\mu_{\overline{m}}}
\newcommand{\Tminline}{\nicefrac{T}{m}}
\newcommand{\Kminline}{\nicefrac{K}{m}}

\usepackage{natbib}
\bibpunct[, ]{(}{)}{,}{a}{}{,}%
\def\bibfont{\small}%
\usepackage{amsmath} 
\newtheorem{prop}{Proposition} 
\usepackage{amssymb}
\usepackage{enumitem}

\usepackage{xcolor} 

\graphicspath{{NumericalExp/Finalfigs/}}

\TheoremsNumberedThrough     

\EquationsNumberedThrough    

\begin{document}
\RUNAUTHOR{Ozbay and Kamble}
\RUNTITLE{Maximal Objectives in the Multi-armed Bandit}

\TITLE{Maximal Objectives in the Multi-armed Bandit with Applications}

\ARTICLEAUTHORS{
		\AUTHOR{Eren Ozbay}
		\AFF{Department of Information and Decision Sciences\\The University of Illinois Chicago\\ \EMAIL{eozbay3@uic.edu}}
		\AUTHOR{Vijay Kamble}
		\AFF{Department of Information and Decision Sciences\\The University of Illinois Chicago\\ \EMAIL{kamble@uic.edu}} 
}

\ABSTRACT{
In several applications of the stochastic multi-armed bandit problem, the traditional objective of maximizing the expected total reward can be inappropriate. In this paper, motivated by certain operational concerns in online platforms, we consider a new objective in the classical setup. Given $K$ arms, instead of maximizing the expected total reward from $T$ pulls (the traditional ``sum" objective), we consider the vector of total rewards earned from each of the $K$ arms at the end of $T$ pulls and aim to maximize the expected highest total reward across arms (the ``max" objective). For this objective, we show that any policy must incur an instance-dependent asymptotic regret of $\Omega(\log T)$ (with a higher instance-dependent constant compared to the traditional objective) and a worst-case regret of $\Omega(K^{1/3}T^{2/3})$. We then design an adaptive explore-then-commit policy featuring exploration based on appropriately tuned confidence bounds on the mean reward and an adaptive stopping criterion, which adapts to the problem difficulty and achieves these bounds (up to logarithmic factors). We then generalize our algorithmic insights to the problem of maximizing the expected value of the average total reward of the top $m$ arms with the highest total rewards. Our numerical experiments demonstrate the efficacy of our policies compared to several natural alternatives in practical parameter regimes. We discuss applications of these new objectives to the problem of grooming an adequate supply of value-providing market participants (workers/sellers/service providers) in online platforms.

}
\maketitle

\input{Intro}

\input{setup}
\input{lower-bound}
\input{algo}
\input{gen_setup}

\input{gen_algo}
\input{numerics}
\input{gen_numerics}
\input{numerics_amazon}

\input{numerics_market}
\input{conclusion}

\bibliographystyle{apalike}
\bibliography{BanditLaborTraining,bibfile}

\newpage
\input{appendix}
\end{document}

%% file: Intro.tex
\section{Introduction}\label{sec:intro}
The stochastic multi-armed bandit (MAB) problem \citep{lai1985asymptotically, auer2002finite} presents a formal framework to study the exploration vs. exploitation tradeoff fundamental to sequential resource allocation in uncertain settings, with wide-ranging applications in areas such as artificial intelligence, adaptive control, economics, marketing, and healthcare. In this problem, given a set of $K$ arms, each of which yields independent and identically distributed (i.i.d.) rewards over successive pulls, the goal is to adaptively choose a sequence of arms to maximize the expected value of the total reward attained at the end of $T$ pulls. The critical aspect of the problem is that the reward distributions of the different arms are a priori unknown. Any good policy must hence, over time, optimize the tradeoff between choosing arms that are known to yield high rewards (exploitation) and choosing arms whose reward distributions are yet relatively unknown (exploration). Over several years of extensive analysis, this classical problem is now well understood (see \cite{lattimore2018bandit}, \cite{slivkins2019introduction}, and \cite{bubeck2012regret} for a survey). 

However, there are several sequential allocation problems arising in practice where the classical objective of maximizing the expected total reward is inappropriate. The main contribution of this paper is the introduction and analysis of a new objective in the classical MAB setup: we consider the vector of cumulative rewards that have been earned from the different arms at the end of $T$ pulls, and instead of maximizing the expectation of their {\it sum}, we aim to maximize the expected value of their {\it maximum} (max). We also address algorithm design for a generalization of this objective, in which we pull $m$ arms in each time period, and we are interested in maximizing the average of the top $m$ cumulative rewards across all arms, where $1\leq m\leq K$ (the max objective corresponding to the case of $m=1$). 

These problems are motivated by the operational concern of what we refer to as {\it supply grooming} in online platforms and marketplaces, which is the concern of ensuring that an adequate fraction of the incoming supply of value-providing market entities gets established as good-quality using limited onboarding resources. What is an adequate fraction depends on the (limited) demand for these entities. Because expending onboarding resources on grooming a higher fraction of the supply than necessary is wasteful, the maximal objectives we consider naturally arise. Consider the following examples.



\begin{enumerate}[labelwidth=!, labelindent=1pt]
	\item {\bf Supply grooming in online service platforms.} An important operational objective of online service platforms is to maintain a pool of well-rated (i.e., with a large rating volume as well as a high average rating) workers or service providers to satisfy the demand for jobs from a discerning clientele. The problem of maintaining such a pool is challenging since (a) workers continuously leave the platform, and hence the quality of new workers must be learned and publicly established through their ratings on an ongoing basis, and (b) only a limited capacity of jobs can be allocated for the risky proposition of obtaining ratings and learning the quality of new workers.\footnote{Generating reviews for new market-entrants is an important and well-recognized challenge in online marketplaces; see, e.g., \cite{luca2017designing, luca2019designing}.} Because the demand is limited, the goal is to ensure that the limited capacity of jobs available for learning gets utilized in obtaining ratings for only the highest quality incoming service workers sufficient to serve the demand for regular jobs.  
	
At the core of this challenging operational question is the following problem. Given a limited number of jobs available for learning, the platform must determine a policy to allocate them to a set of new workers to maximize some appropriate functional of their terminal publicly-observable quality levels. The first question is, what defines a quality level? Since customers care about both the average rating as well as the number of received ratings, one natural definition is the product of the two, i.e., the sum of all ratings received by a provider. The second question is, what functional of the terminal total ratings is appropriate? For a platform that seeks to serve a limited client demand, maximizing the average total rating across {\it all} workers may not be necessary. A more appropriate objective is to maximize the average total rating of the top $q^{\textup{th}}$ percentile workers ordered by their terminal total ratings, where $q$ is determined by the volume of demand for regular jobs: higher the demand for jobs, the higher the $q$ needed. Essentially, the ratings of the lower-rated workers at the end of the onboarding period do not matter since there is not enough demand for regular jobs to assign to them anyway. 


To address this problem, we can use the MAB framework: the set of arms is the set of new service workers, the reward of an arm is the random increment in the worker's total rating (i.e., the rating received) after performing a job, and the number of jobs available for learning is $T$. Given $K$ workers, the goal is to adaptively allocate the jobs to them to maximize the average terminal total rating amongst the top $1\leq m\leq K$ (where $m\approx qK$) best-rated workers. This motivates the model and the general objective we consider. While our primary focus is on the max objective, i.e., the case of $m=1$, we use our insights from this setting to design a well-performing algorithm for the case where $m>1$. In Section~\ref{sec:numerics_market}, we discuss how we can implement the resulting algorithms in a dynamic market model where service workers continuously arrive and depart, and there is a finite capacity of incoming jobs, using a ``cohorting" approach.


\item {\bf Training in online service platforms.} 
Another goal of supply grooming in online service platforms could be to {\it train} novice workers in settings where worker skills improve with experience. Skill improvement of a worker can be tracked, for example, by observing the number of satisfactory job completions or the increase in the total rating, which may be reasonable proxies in the early phase of a worker's lifetime. 
Given limited training resources, e.g., the capacity of jobs available for training, the goal then is to adaptively allocate these resources to a set of novice workers to maximize some functional of their terminal skill levels. As discussed above, since the demand for regular jobs is limited, it may not be necessary to train all workers. A more appropriate objective is to maximize the average skill level of the top $q^{\textup{th}}$ percentile workers ordered by their terminal skill levels, where the volume of demand for regular jobs determines $q$.

	\item {\bf Product grooming on e-commerce platforms.} E-commerce platforms often feature similar substitutes within a product category. For instance, consider a product like a tablet cover (e.g., for an iPad). Once the utility of a new product of this type becomes established (e.g., the size specifications of a new version of the iPad becomes available), several brands offering close to identical products serving the same purpose proliferate the marketplace. This proliferation is problematic to the platform for two reasons: (a) customers are inundated by choices and may unnecessarily delay their purchase decision, thereby increasing the possibility of leaving the platform altogether \citep{settle1974consumer, gourville2005overchoice}, and (b) the heterogeneity in the purchase behavior resulting from the lack of a clear choice may complicate the problem of effectively managing inventory and order-fulfillment decisions. Given a budget for incentivizing customers to pick different products in the early exploratory phase when the qualities of the different products are being discovered, a natural objective for the platform is to groom a product to have the highest publicly observable quality level at the end of this phase. This product then becomes a clear choice for the customers. Our objective effectively captures this goal when the quality level is defined to be the sum total of ratings received by the product.
\end{enumerate}
In the above applications, it is important that algorithms addressing our objectives can perform well for small values of $T$ given the limited onboarding resources available (or, alternatively, from the perspective of minimizing the utilization of such resources). It is also in this small $T$ regime that the cumulative performance measures that we consider for the arms are appropriate. For example, maximizing the total ratings for a service provider or a product is appropriate as an objective only when the number of ratings is small since, while customers have greater confidence in the average rating if the number of ratings is higher, the rating volume does not matter much beyond a point. Similarly, for the application to training, the total rating can be assumed to be a proxy for the skill level of a worker only in the early phase of a worker's lifetime on the platform since skill increments are expected to diminish over successive job allocations. 

We next discuss our technical results in detail.

{\bf The max objective.} We first discuss our results for the max objective $(m=1)$. A key assumption we make in the paper is that the rewards for all arms are non-negative; this is motivated by the applications discussed above where rewards represent ratings or skill increments. Under this assumption, in the {\it full-information} setting where the reward distributions of the arms are known, we first show that the optimal policy for the max objective is identical to the one for the sum objective: one always pulls the arm with the highest mean reward (Proposition~\ref{prop:main}). This additionally implies that the optimal rewards under the two objectives are identical.

A standard approach in MAB problems is to design a policy that minimizes {\it regret}, i.e., the quantity of loss relative to the optimal full-information policy for a given objective over time. 
In the classical setting with the sum objective, it is well known that any policy must incur an instance-dependent asymptotic regret of $\Omega(\sum_{i\neq i^*} (\Delta_i\log T)/d_i)$ as $T\rightarrow \infty$ \citep{lai1985asymptotically}. Here, $\Delta_i = \mu^*-\mu_i$, i.e., it is the difference between the highest mean reward $\mu^*$ belonging to the arm $i^*$ and the mean reward $\mu_i$ of arm $i$; and $d_i$ is a quantity that captures an appropriate notion of divergence between the reward distribution of arm $i$ and the ``closest'' distribution within the space of possible distributions having a mean that is at least $\mu^*$. Additionally, it is also well-known that any policy must incur an instance-independent regret of $\Omega(\sqrt{KT})$ in the worst-case over the set of possible bandit instances \citep{auer2002finite}. 

Since the optimal full-information reward is the same under the sum and the max objectives, and since the maximum of a set of non-negative numbers is always at most the sum of the numbers, any lower bound on the regret for the sum objective implies the same lower bound on the max objective (conversely, any upper bound achieved by a policy on the regret under the max objective also holds under the sum objective). However, a key feature of the max objective is that the rewards earned from arms that do not eventually turn out to be the ones yielding the highest cumulative reward are effectively a waste. Owing to this feature, we show that any policy must incur a {\it higher} instance-dependent regret of $\Omega(\sum_{i\neq i^*} (\mu^*\log T)/{d_i})$ in this case (Theorem~\ref{thm:asymplwb}). Moreover, we show that an instance-independent regret of $\Omega(K^{1/3}T^{2/3})$ is inevitable in the worst-case (Theorem~\ref{thm:lb}). Both these results rely on novel arguments that are a significant departure from those involved in proving the corresponding lower bounds for the sum objective.

Attaining these lower bounds simultaneously requires algorithmic innovation. For the sum objective, well-performing policies are typically based on the principle of optimism in the face of uncertainty. A popular policy class is the Upper Confidence Bound (UCB) class of policies \citep{agrawal1995sample,auer2002finite,auer2010ucb}, in which a confidence interval is maintained for the mean reward of each arm, and at each time, the arm with the highest upper confidence bound is chosen. For a standard tuning of these intervals, this policy -- termed UCB1 in literature due to \cite{auer2002finite} -- guarantees the optimal instance-dependent asymptotic regret of $\textup{O}(\sum_{i\neq i^*} (\Delta_i\log T)/d_i)$ up to a constant factor and a regret of $\textup{O}(\sqrt{KT\log T})$ in the worst case. With a more refined tuning, $\textup{O}(\sqrt{KT})$ can be achieved \citep{audibertminimax, lattimore2018refining}. 

It is easy to show that as long as the mean rewards for the arms are distinct, several conventional policies designed for the sum objective, including UCB1, attain the optimal instance-dependent asymptotic regret bound of $\textup{O}(\sum_{i\neq i^*} (\mu^*\log T)/{d_i})$ for the max objective up to a constant factor; see Section~\ref{apxsec:asym} in the Appendix for a general result. Essentially, for a fixed instance, in the long run, these algorithms only expend the inevitable $\textup{O}(\log T/d_i)$ number of pulls on any suboptimal arm $i$ in expectation, thus focusing mostly on the optimal arm as is required to maximize the max objective. However, the worst-case performance of any of these policies can be disastrous. A direct way to see this is to consider an extreme example where all $K$ arms yield a deterministic reward of 1. Then, UCB1 will pull each of the arms in a round-robin fashion until a total of $T$ pulls, resulting in the highest terminal cumulative reward of $\textup{O}(T/K)$; whereas a reward of $T$ is feasible by simply committing to an arbitrary arm from the start, implying a $\Omega(T)$ regret. This basic observation extends to instances with arms with random rewards and distinct means: in Proposition~\ref{prop:ucbbad}, we show that UCB1 necessarily incurs a $\Omega(T)$ regret for the max objective in a two-armed bandit problem with Bernoulli rewards and means $0.5$ and $0.5+1/\sqrt{T}$. In Theorem~\ref{thm:sumbad} we show a general result: {\it any} policy that guarantees a $\tilde{\textup{O}}(\sqrt{T})$ regret for the sum objective must necessarily incur a $\Omega(T^\alpha)$ regret in the worst case for the max objective for any $\alpha\in (0,1)$. Essentially, these policies waste too many pulls distinguishing between arms with similar rewards, which, while not wasteful for the sum objective, can be starkly detrimental for the max objective.



This observation suggests that any worst-case optimal policy must, at some point, stop exploring and permanently commit to a single arm. A natural candidate is the basic explore-then-commit (ETC) policy, which uniformly explores all arms until some time that is fixed in advance, and then commits to the empirically best arm \citep{lattimore2018bandit, slivkins2019introduction}. When each arm is chosen $(T/K)^{2/3}$ times in the exploration phase, this strategy can be shown to achieve a regret of $\textup{O}(K^{1/3}T^{2/3}\sqrt{\log K})$ relative to the sum objective \citep{slivkins2019introduction}. It is easy to argue that it achieves the same regret relative to the max objective. However, this policy is excessively optimized for the worst case where the means of all the arms are within $(K/T)^{1/3}$ of each other. When the arms are easier to distinguish, this policy's performance is quite poor due to excessive exploration. For example, consider a two-armed bandit problem with Bernoulli rewards and means $(0.5, 0.5-\Delta)$, where $\Delta>0$. For this fixed instance, ETC will pull both arms $\Omega(T^{2/3})$ times and hence incur a regret of $\Omega(T^{2/3})$ for the max objective regardless of $\Delta$. However, UCB1 will incur the optimal instance-dependent regret of $\textup{O}(\log T/\Delta^2)$ for this instance, which could be much smaller if $\Delta$ is large. Thus, although the worst-case regret of UCB1 is $\Omega(T)$, its performance can be significantly better than ETC for easy bandit instances. 

These observations motivate us to seek a practical policy for our objective with a graceful dependence of performance on the difficulty of the bandit instance, which will achieve both: the worst-case bound of $\tilde{\textup{O}}(K^{1/3}T^{2/3})$ and the instance-dependent asymptotic bound of $\textup{O}(\sum_{i\neq i^*} (\mu^*\log T)/{d_i})$. We emphasize that since we are interested in optimizing performance in the finite $T$ regime for the applications we consider, it is not sufficient to satisfy ourselves with the notion of asymptotic optimality. We must contend with the possibility of encountering instances where the reward gaps across arms are small in relation to the range of $T$ values we may be interested in.

We propose a new policy with an explore-then-commit structure, in which appropriately defined confidence bounds on the means of the arms are utilized to guide exploration, as well as to decide when to stop exploring. We call this policy Adaptive Explore-then-Commit (ADA-ETC). 
We show that ADA-ETC adapts to the problem difficulty by exploring less, if appropriate, while attaining the same regret guarantee of $\textup{O}(K^{1/3}T^{2/3}\sqrt{\log K})$ attained by vanilla ETC in the worst case (Theorem~\ref{thm_stopping}). 
In particular, ADA-ETC guarantees an instance-dependent asymptotic regret of $\textup{O}(\log T)$ as $T\rightarrow\infty$, matching our instance-dependent lower bound up to a constant factor. Finally, our numerical experiments demonstrate that ADA-ETC results in significant improvements over the performance of vanilla ETC in easier settings, while never performing worse in difficult ones, thus corroborating our theoretical results. Our numerical results also demonstrate that naive ways of introducing adaptive exploration based on upper confidence bounds, e.g., simply using the upper confidence bounds of UCB1, may lead to no improvement over vanilla ETC for practical values of $T$ and $K$.

{\bf The case of $m>1$.} We next consider an extension to settings where one is interested in maximizing the expected average cumulative reward across the top $m$ arms with the highest cumulative rewards. When the decision-maker can pull one arm per time period, this objective, however, is equivalent to the max objective: the best way to maximize the average cumulative reward across the top $m$ arms is to invest all pulls in the best arm. In practice, though, such a solution is far from being appropriate. For example, online labor platforms typically want to provide robust service guarantees to the clients, and hence, training a handful of ``stars'' while most other workers serving the clients are inadequately trained is not a desirable outcome. Moreover, while we expect $T/m$ to be small, $T$ itself could be large. In a training application, if one invests all the $T$ jobs into training a single worker, it may not be reasonable to assume that the skill increments of this worker are i.i.d. over time. In application to improving ratings of workers, the total rating may not be an appropriate metric to capture customers' preferences in the perverse extreme where all $T$ jobs are spent on improving the total ratings of a single worker.  



To account for these concerns, we consider a modification of our problem. While the objective remains the same, we assume that there are $T/m$ periods, and in each period, the decision-maker pulls $m$ distinct arms. This is equivalent to the constraint that the decisions of $T$ pulls are sequentially taken over $T/m$ batches of size $m$, with the additional requirement that the pulls in each batch are distinct. Such batching has the additional benefit that it may significantly reduce the onboarding period for new market participants. With such a constraint, it is not feasible, let alone optimal, for the decision-maker to invest all $T$ pulls in a single arm.  (In the concluding Section~\ref{sec:conclusion}, we discuss another formulation to achieve this goal.) Extending Proposition~\ref{prop:main}, we can show that the optimal policy that maximizes the average cumulative reward across the top $m$ arms is the one that always pulls the $m$ arms with the highest mean in each time period (i.e., in each batch). We then design an adaptive explore-then-commit policy inspired by the max objective (m-ADA-ETC) that achieves a $\widetilde{\textup{O}}(K^{1/3}T^{2/3}/m)$ upper bound on the regret. We also prove a $\Omega(K^{1/3}T^{2/3}/m^{4/3})$ lower bound on the regret in this case for when $2m\leq K<T$.
Our extensive numerical tests show that this policy significantly outperforms other natural policies, including the policy of implementing the optimal algorithm for the max objective independently on $m$ randomly selected sets of arms, each of size $\approx K/m$. 


{\bf Organization.} The paper is organized as follows. We discuss relevant literature in Section~\ref{sec:lit}. Our model and the max objective are introduced in Section~\ref{sec:maxobj}. In this section, we also present the analysis of the max objective, where we first prove lower bounds on the regret and then present the ADA-ETC policy and the corresponding upper bounds that it achieves. In Section~\ref{sec:generalm}, we present the results for the extension of our objective for $m>1$. Our numerical experiments are presented in Section~\ref{sec:numerics}, in which we also describe the implementation of our algorithms in a dynamic market simulation. We conclude the paper by discussing further applications and open questions in Section~\ref{sec:conclusion}.

\subsection{Related literature}\label{sec:lit}
We discuss the connections of our model and results to five distinct streams of literature.


\noindent {\bf Pure exploration in bandits.} Our max objective endogenizes the goal of quickly identifying the arm with approximately the highest mean reward so that a substantial amount of time can be spent earning rewards from that arm (e.g., ``training'' a worker). This goal is related to the {\it pure exploration} (or {\it best-arm identification}) problem in multi-armed bandits. Several variants of this problem have been studied, where the goal of the decision-maker is to either minimize the probability of misidentification of the optimal arm given a fixed budget of pulls \citep{audibert2010best, kaufmann2016complexity, carpentier2016tight}; or minimize the expected number of pulls to attain a fixed probability of misidentification, possibly within an approximation error \citep{even2002pac, mannor2004sample, even2006action, karnin2013almost, vaidhiyan2017learning, jamieson2014lil, kaufmann2016complexity}; or to minimize the expected suboptimality (called ``simple regret'') of a recommended arm after a fixed budget of pulls \citep{bubeck2009pure, bubeck2011pure, carpentier2015simple}.  \cite{jun2016top} additionally has studied the pure-exploration problem under batching constraints similar to our $m>1$ setting. Extensions to settings where multiple good arms are needed to be identified have also been considered \citep{bubeck2013multiple, kalyanakrishnan2012pac, zhou2014optimal, kaufmann2013information}. 

The critical difference from these problems is that in our scenario, the budget of $T$ pulls must not only be spent on identifying an approximately optimal arm but also on earning rewards on that arm. For example, consider the best-arm identification problem with two arms with means separated by $\Delta$ and a fixed budget of $T$ pulls. It is known that the optimal policy that minimizes the probability of misidentification is to allocate $T/2$ pulls to each arm, resulting in an exponentially small (in $T$, for a fixed $\Delta$) probability of misidentification \citep{audibert2010best}. But this policy necessarily incurs a regret of $\Theta(T)$ for our max objective, which requires quickly identifying and focusing on a ``good enough" arm for most (i.e., $T-\textup{o}(T)$) of the pulls. Moreover, any choice of apportionment of the budget of $T$ pulls to the identification problem, or a choice for a target for the approximation error or probability of misidentification (to qualify what is a ``good enough" arm), is a priori unclear and must arise {\it endogenously} from our primary objective. 

\noindent {\bf Bandits with switching costs and batched bandits.} The fact that focusing on one arm, in the long run, is prudent for our objective thematically relates this work to the literature on bandits with switching costs, where there is a cost incurred for switching from one arm to another \citep{cesa2013online,dekel2014bandits}. Another related line of work is on {\it batched} bandits, which imposes a constraint that the policy must split the arm pulls into a small number of batches \citep{perchet2016batched, jun2016top, gao2019batched, esfandiari}. However, we note that our objective does not simply amount to keeping the number of switches or batches low, or designing algorithms around controlling the size of batches along with identifying the best arm(s) sooner \citep{jin2019efficient}; it also matters how ``spread apart'' these switches are. For example, $\textup{o}(T)$ switches at the beginning of the time horizon may only result in a regret of $\textup{o}(T)$ (as is the case for our policy ADA-ETC), while a single switch at time (for example) $T/2$ necessarily results in a regret of $\Theta(T)$ for our objective. This implies that our objective requires a different algorithmic approach. To enforce this point, we note that the algorithm of \cite{cesa2013online}, which restricts the number of switches/batches to $\textup{O}(\log \log T)$ while attaining $\tilde{\textup{O}}(\sqrt{T})$ regret for the sum objective, necessarily achieves a suboptimal worst-case regret guarantee for our max objective as implied by Theorem~\ref{thm:sumbad}. 

\noindent {\bf Non-standard objectives in online learning.} While several non-standard objectives have been considered before in the literature on online learning and decision-making, the objectives we study appear novel. One related objective has been considered in what has been referred to as the max K-armed bandit model \citep{cicirello2005max, streeter2006asymptotically, streeter2006simple} or the extreme bandits model \citep{carpentier2014extreme, bhatt2022extreme, baudry2022efficient} in the literature. These works consider a multi-armed bandit problem where the objective is to maximize the expected value of the maximal reward across all $T$ pulls. Any policy is benchmarked against the policy of choosing an arm that yields the highest expected value of the maximal reward across $T$ pulls. The max ($m=1$) objective we consider sits in between the extremes of the classical sum objective on the one hand and the objective of extreme bandits on the other, in which we want to maximize the maximal total reward across arms (sum for each arm, then maximize the maximal sum across arms). 
In many ways, our max objective enables a cleaner analysis than extreme bandits, at least partly because it is just the mean rewards of the arms that matter in defining the benchmark optimal policy, as is the case for the sum objective. Defining the optimal benchmark policy itself can be non-trivial in extreme bandits: to obtain meaningful asymptotics, one typically has to assume that there exists a {\it unique} dominating arm with the highest expected value of maximal reward across $T$ independent pulls for any large enough $T$. 

\cite{even2009online} and \cite{even2010learning} introduce an online learning problem under a set of new cost-minimization objectives. In this problem, arms accrue costs over time, and the goal is to minimize the $L_d$ norm of the vector of expected accrued costs for each arm for $d>1$, where the expectation is over the randomization in each pull (but not over any randomness in the sequence of costs). An extreme example of this objective is the $d=\infty$ case, amounting to minimizing the maximum accrued cost across all arms. They consider both cases where the losses are adversarially generated \citep{even2009online} and the stochastic setting where the losses are generated i.i.d. from a fixed but unknown distribution over $[0,1]^K$ \citep{even2010learning}, which is more relevant to our work. 

There are two key distinctions between these works and our work. First, they assume full information feedback, i.e., the cost incurred by each arm is revealed at the end of each stage, while we assume bandit feedback, where only the reward of the arm that is pulled is observed. Thus there is no exploration vs. exploitation tradeoff in their model in the stochastic setting, which, on the other hand, is fundamental to our model. The hardness of their problem instead mainly arises from competing with a static benchmark policy that observes the entire sequence of costs in advance (as opposed to the static benchmark policy that optimizes the {\it expected} cost or reward typically considered in bandit problems). Second, at a high-level, even assuming bandit feedback under a weaker expected cost minimization benchmark (where the benchmark is, e.g., the static policy that minimizes the expected $L_d$ norm of the costs across arms, where the expectation is over both, the randomness in the cost distribution as well as in the policy), the tradeoff between exploration and exploitation appears to be more benign in these cost minimization objectives. This is because, under any non-trivial cost distribution across arms, the benchmark policy optimizing any $L_d$ norm objective for $d>1$ would pull all arms with a positive probability. Thus, myopic cost minimization efforts aren't misaligned with learning the distribution and organically result in exploration. This is unlike our case (or the traditional sum objective), where myopically optimizing the expected reward in interim information states results in pulling only a subset of arms (e.g., a single arm in the $m=1$ case), which may not allow learning the possibility that some other arm(s) may be optimal instead.

\noindent {\bf Explore-then-commit algorithms.} Explore-then-commit algorithms, both with adaptive and non-adaptive stopping rules, have been extensively studied for a range of multi-armed bandit problems under the sum objective \citep{perchet2013multi, perchet2016batched, garivier2016explore, jin2021double}. It can be shown that with a non-adaptive stopping rule, the best regret one can achieve is $\tilde\Theta(T^{2/3}K^{1/3})$ \citep{lattimore2018bandit}. As we have argued in Section~\ref{sec:intro}, this algorithm achieves a suboptimal instance-dependent performance for the max objective. \cite{garivier2016explore} has shown that with an adaptive stopping rule defined using upper and lower confidence bounds, one can achieve both an instance-independent regret of $\textup{O}(\sqrt{T})$ and instance-dependent regret of $\textup{O}(\log T/\Delta)$ (for the sum objective) in a two-armed bandit problem with Gaussian rewards. Such algorithms are promising for our max objective since they eventually commit to a single arm. However, because the commitment rule is not correctly optimized for the max objective, a suboptimal $\omega(T^{2/3})$ regret is inevitable in the worst case under this policy, as we show in Theorem~\ref{thm:sumbad}.



\noindent {\bf Learning in online platforms.} The operational concerns of learning with the goal of efficient matchmaking in online platforms and marketplaces have received significant attention in recent literature \citep{johari2021matching, shah2020adaptive, massoulie2018capacity, hsu2021integrated, suncongestion, kamble2022exploration}. The goal of these works is to design effective online learning policies that can be implemented by the platform in the face of capacity constraints induced by limited demand. Similar to the settings in these works, we consider a learning problem in a market setting with capacities induced by demand constraints. There are, however, two key differences from this literature. First, this literature typically focuses on the traditional objective of maximizing the total utility generated in the market, while our distinction is the focus on a new objective motivated by the problem of supply grooming in online platforms. Second, unlike these settings, where the capacity constraint results from limited demand, there are two types of capacity constraints that we account for: (a) the limited capacity of onboarding jobs, which are distinct from the regular jobs (this constraint determines $T$ in our model), and (b) the limited capacity of regular jobs due to which not all service workers' quality needs to be learned or not all workers need to be trained (this constraint determines $m$ in our model). Effectively, our focus is on a learning problem in the onboarding phase of arriving cohorts of workers or service workers given a limited supply of jobs for this onboarding process.

%% file: setup.tex
\section{Model and the max objective}\label{sec:maxobj}
\label{setup}
Consider the stochastic multi-armed bandit (MAB) problem parameterized by the number of arms, which we denote by $K$; the length of the decision-making horizon (the number of discrete times/stages), which we denote by $T$; and the probability distributions for arms $1, \dots, K$, denoted by $\nu_1, \dots, \nu_K$, respectively. We assume that the rewards are non-negative and their distributions have a bounded support, assumed to be $[0,1]$ (although, this latter assumption can be easily relaxed to allow, for instance, $\sigma$-Sub-Gaussian distributions with bounded $\sigma$). We define $\mathcal{V}$ to be the set of all $K$-tuples of distributions for the $K$ arms having support in $[0,1]$. Let $\mu_1,\dots, \mu_K$ be the means of the distributions. Without loss of generality, unless specified otherwise, we assume that $\mu_1\geq \mu_2\geq\dots\geq \mu_K$ for the remainder of the discussion. The distributions of the rewards from the arms are unknown to the decision-maker. We denote $\bnu = (\nu_1, \dots, \nu_K)$ and $\bmu = (\mu_1,\dots, \mu_K)$. We also define $\Delta_i = \mu_1 - \mu_i$ for $i \in \{1,\dots,K\}$.

At each time, the decision-maker chooses an arm to play and observes a reward. 
Let the arm played at time $t$ be denoted as $I_t$ and the reward be denoted as $X_t$, where $X_t$ is drawn from the distribution $\nu_{I_t}$, independent from the previous actions and observations. The history of actions and observations at any time $t\geq 2$ is denoted as $\mathcal{H}_t = (I_1, X_1, I_2, X_2,\dots, I_{t-1},X_{t-1})$, and $\mathcal{H}_1$ is defined to be the empty set $\phi$. A {\it policy} $\pi$ of the decision-maker is a sequence of mappings $(\pi_1,\pi_2,\dots,\pi_T)$, where $\pi_t$ maps every possible history $\mathcal{H}_t$ to an arm $I_t$ to be played at time $t$. Let $\Pi_T$ denote the set of all such policies. 

For an arm $i$, we denote $n^i_t$ to be the number of times this arm is played until and including time $t$, i.e., $n^i_t = \sum_{s=1}^t \mathbbm{1}_{\{I_s = i\}}$. We also denote $U^i_n$ to be the reward observed from the $n^{\textup{th}}$ pull of arm $i$. $(U^i_n)_{n\in\mathbb{N}}$ is thus a sequence of i.i.d. random variables, each distributed as $\nu_i$. Note that the definition of $U^i_n$ implies that we have $X_t = U^{I_t}_{n^{I_t}_t}$. We further define $\overline{U}^i_{t} \triangleqq \sum_{n=1}^{n^i_t}U^i_n$
to be the cumulative reward obtained from arm $i$ until time $t$. 

Once a policy $\pi$ is fixed, then for all $t = 1,\dots, T$, $I_t$, $X_t$, and $n^i_t$ for all $i \in \{1,\dots,K\}$, become well-defined random variables. We consider the following notion of reward for a policy $\pi$:
\begin{align}
\mathcal{R}_T(\pi,\bnu) = \E_{\bnu}\big(\max\big(\overline{U}^1_{T}, \overline{U}^2_{T}, \dots, \overline{U}^K_{T}\big)\big).
\end{align}
In words, the objective value attained by the policy is the expected value of the largest cumulative reward across all arms at the end of the decision making horizon. 

When the reward distributions $\nu_1, \dots, \nu_K$ are known to the decision-maker, then for a large $T$, the best reward that the decision-maker can achieve is \[\sup_{\pi\in\Pi_T} \mathcal{R}_T(\pi, \bnu).\]

A natural candidate for a ``good'' policy when the reward distributions are known is the one where the decision-maker exclusively plays arm $1$ (the arm with the with the highest mean), attaining an expected reward of $\mu_1T$. Let us denote $\cR^*_T(\bnu) \triangleqq \mu_1T$. One can show that, in fact, this is the best reward that one can achieve in our problem.

\begin{prop}
	\label{prop:main} For any bandit instance $\bnu\in\mathcal{V}$, $\sup_{\pi\in\Pi_T} \mathcal{R}_T(\pi,\bnu) = \cR^*_T(\bnu)$.
\end{prop}

The proof is presented in Section~\ref{apx:prop1} in the Appendix. This shows that the simple policy of always picking the arm with the highest mean is optimal for our problem.
Next, we denote the {\it regret} of any policy $\pi$ to be \[\mathcal{\textup{Reg}}_T(\pi, \bnu) = \sup_{\pi\in\Pi_T} \mathcal{R}_T(\pi, \bnu)- \mathcal{R}_T(\pi,\bnu).\] 

In the rest of this section, we focus on two objectives. The first is to design a policy $\pi_T\in\Pi_T$, which attains an asymptotically optimal instance-dependent (i.e., $\bnu$ dependent) bound on $\mathcal{\textup{Reg}}_T(\pi_T, \bnu)$, simultaneously for (almost) all instances $\bnu\in\mathcal{V}$ as $T\rightarrow \infty$.  The second objective is to design a policy $\pi_T\in\Pi_T$, which achieves the smallest regret in the worst-case over all distributions $\bnu\in\mathcal{V}$, i.e., the one that solves the optimization problem:
\[\PSR^*_T \triangleqq \inf_{\pi\in\Pi_T}\sup_{\bnu\in\mathcal{V}}\PSR_T(\pi, \bnu),\]
where $\PSR^*_T$ denotes the \textit{minmax} (or the \textit{best worst-case}) \textit{regret}. In the remainder of this section, we design a {\it single} policy that attains the first objective to within a constant factor and the second objective to within a logarithmic factor. 

%% file: lower-bound.tex
\subsection{Lower Bounds}\label{lb1}
We first provide an instance-dependent $\Omega(\log T)$ asymptotic lower bound on the regret. We let $\mathcal{M}$ be the set of distributions with support in $[0,1]$. For $\nu \in \mathcal{M}$, and $\mu\in[0,1]$, define $d_{\textup{inf}}\left(\nu , \mu, \mathcal{M}\right) = \inf\limits_{\nu^\prime \in \mathcal{M}} \left\{ \textup{D}(\nu, \nu^\prime) : \mu(\nu^\prime) > \mu \right\}$, where $\mu(\nu)$ denotes the mean of distribution $\nu$, and $\textup{D}(\nu, \nu^\prime)$ is the Kullback-Leibler (KL) divergence between the distributions $\nu$ and $\nu^\prime$. $d_{\textup{inf}}\left(\nu , \mu , \mathcal{M}\right)$ is thus the smallest KL divergence between the distribution $\nu$ and any other distribution in $\mathcal{M}$ whose mean is at least $\mu$. 

We say that a sequence of policies $(\pi_T)_{T\in\mathbb{N}}$, where $\pi_T\in \Pi_T$ for all $T\in\mathbb{N}$, is {\it consistent} for a class $\mathcal{V} = \mathcal{M}^K$ of stochastic bandits, if for all $\bnu \in \mathcal{V}$ such that there is a unique arm with the highest mean reward, and for any $p >0$, we have that $\lim\limits_{T \to \infty} \PSR_T(\pi_T, \bnu)/{T^p} = 0$. We then have the following result.

\begin{theorem}\label{thm:asymplwb}
	Consider a class $\mathcal{V}=\mathcal{M}^K$ of $K$-armed stochastic bandits and 
	let $(\pi_T)_{T\in\mathbb{N}}$ be a consistent sequence of policies for $\mathcal{V}$. 
	Then, for all $\bnu \in \mathcal{V}$ such that the optimal arm is unique,
	\[\liminf\limits_{T \to \infty} \frac{\PSR_T(\pi_T, \bnu)}{\log(T)} \geq \sum_{i\neq k^*} \frac{\mu^*}{d_{\textup{inf}}\left(\nu_i, \mu^*, \mathcal{M}\right)}, \]
	where $k^*$ is the optimal arm with the highest mean $\mu^*$.
\end{theorem}

The proof of Theorem~\ref{thm:asymplwb} is presented in Section~\ref{apx:asymplwb} in the Appendix. The result has an intuitive explanation. For convenience, we denote $d_i = d_{\textup{inf}}\left(\nu_i, \mu^*, \mathcal{M}\right)$. Similar to the proof of the lower bound for the sum objective \citep{lai1985asymptotically}, we can show that for any consistent sequence of policies, each suboptimal arm $i$ must be pulled $\Omega(\log T/d_i)$ number of times in expectation. However, unlike the sum objective where each such pull yields a mean reward of $\mu_i$ and results in an expected regret of $\Delta_i$, for the max objective, each such pull is wasteful and results in an expected regret of $\mu^*$. 

Despite this intuitive explanation of the result, the proof is not straightforward. In particular, showing that each suboptimal arm $i$ must be pulled $\log T/d_i$ times in expectation doesn't directly allow us to account for a regret contribution of $\mu^*\log T/d_i$ from arm $i$. This is because, in the full-information setting, with a (relatively high) probability of $\log T/(Td_i)$, one can choose to pull a suboptimal arm $i$ for all the $T$ time periods (and pull the optimal arm for $T$ periods with the remaining probability), thus ensuring that it gets pulled $\log T/d_i$ times in expectation and at the same time resulting in an expected reward contribution of $\mu_i\log T/d_i$, and hence a regret contribution of $(\mu^*-\mu_i)\log T/d_i = \Delta_i\log T/d_i$. To show that this regret is not achievable, we prove a stronger result: we show that  for each $\alpha \in (0,1]$, a suboptimal arm $i$ must be pulled $\alpha\log T/d_i$ times in expectation until time $T^\alpha$ (Proposition~\ref{prop:auxiliary_stronger} in the Appendix). We then argue that the probability of a suboptimal arm being the one with the highest cumulative reward cannot be too high for any consistent sequence of policies, and thus the best way to satisfy the stronger set of lower bounds on the number of pulls for the suboptimal arms in terms of minimizing regret is to chalk these pulls as wasted. This allows us to conclude the higher lower bound on the regret.

We next show that for our objective, a regret of $ \Omega(K^{1/3} T^{2/3})$ is inevitable in the worst case. 
\begin{theorem}\label{thm:lb}
	Suppose that $T>K$. Then, $\PSR^*_T \geq  \Omega((K-1)^{1/3} T^{2/3}).$
\end{theorem}
The proof is presented in Section~\ref{apx:thm1} in the Appendix. Informally, the argument for the case of $K=2$ arms is as follows. Consider two bandits with Bernoulli rewards, one with mean rewards $(1/2 +1/T^{1/3}, 1/2)$, and the other with mean rewards $(1/2 +1/T^{1/3}, 1/2+2/T^{1/3})$. Then until time $\approx T^{2/3}$, no algorithm can reliably distinguish between the two bandits. Hence, until this time, either $\Omega(T^{2/3})$ pulls are spent on arm 1 irrespective of the underlying bandit, or $\Omega(T^{2/3})$ pulls are spent on arm 2 irrespective of the underlying bandit. In both cases, the algorithm incurs a regret of $\Omega(T^{2/3})$, essentially because of wasting $\Omega(T^{2/3})$ pulls on a suboptimal arm that could have been spent on earning a reward on the optimal arm. This latter argument is not entirely complete, however, since it ignores the possibility of always picking a suboptimal arm until time $T$, in which case spending time on the suboptimal arm in the first $\approx T^{2/3}$ periods was not wasteful. However, even in this case, we can argue that one incurs a regret of $\approx T\times(1/T^{1/3}) = \Omega(T^{2/3})$. Thus a regret of $\Omega(T^{2/3})$ is unavoidable. Our formal proof builds on this basic argument to determine the optimal dependence on $K$. 

Finally, we show that policies optimized for the sum objective do not suffice for attaining the optimal instance-independent regret performance for the max objective. In particular, we show that policies achieving the near-optimal instance-independent regret guarantee of $\tilde{\textup{O}}(\sqrt{T})$ for the sum objective necessarily incur a suboptimal instance-independent regret for the max objective.

\begin{theorem}\label{thm:sumbad}
Consider a policy for the two-armed bandit problem that achieves an instance-independent regret bound of $C(\log T)^g\sqrt{T}$ for the sum objective for some $C,\,g>0$ and $T$ large enough. Then the worst-case regret for this policy for the max objective is $\Omega(T^{\alpha})$ for any $\alpha <1 $.   
\end{theorem}
The proof is presented in Section~\ref{apx:sumbad} in the Appendix. The high-level idea of the proof is to show that to achieve a $\tilde{\textup{O}}(\sqrt{T})$ regret bound, any sum-optimal policy must distinguish between arms whose means are separated by $1/T^{\alpha/2}$ for any $\alpha\in [0,1)$. But this requires pulling both arms at least $\Omega(T^{\alpha})$ times. This necessarily results in $\Omega(T^{\alpha})$ regret for the max objective. 

%% file: algo.tex
\subsection{Adaptive Explore-then-Commit (ADA-ETC)}
We now define an algorithm that we call Adaptive Explore-then-Commit (ADA-ETC) specifically designed for our problem. It is formally defined in Algorithm~\ref{alg:main}. The algorithm can be simply described as follows. After choosing each arm once, choose the arm with the highest upper confidence bound, until there is an arm such that (a) it has been played at least $\tau= \lceil T^{2/3}/K^{2/3}\rceil$ times, and (b) its empirical mean is higher than the upper confidence bounds on the means of all other arms. Once such an arm is found, commit to this arm until the end of the decision horizon. 
\begin{algorithm}[h]
	\caption{Adaptive Explore-then-Commit (ADA-ETC)}\label{alg:main}
	\vspace{0.2cm}
	{\bf Input:} $K$ arms with horizon $T$. \\
	{\bf Define:} $\tau = \lceil\frac{T^{2/3}}{K^{2/3}}\rceil$. For $n\geq 1$, let $\bar{\mu}_n^i$ be the empirical average reward from arm $i$ after $n$ pulls and it remains fixed after $\tau$ pulls, i.e., $\bar{\mu}_n^i = \frac{1}{\min\{n, \tau\}} \sum_{s=1}^{\min\{n, \tau\}} U_s^i$.  
	Also, for $n\geq 1$, define,
	\begin{align}
	\ucb^i_n &= \bar{\mu}_n^i + \sqrt{\frac{4}{n}\log\left(\frac{T}{Kn^{3/2}}\right)}\mathbbm{1}_{\left\{ n< \tau \right\}}.\label{ucb}\\
	\lcb^i_n &=  \bar{\mu}_n^i - \bar{\mu}_n^i\mathbbm{1}_{\{n< \tau\}}.\label{lcb}
	\end{align}
	Also, for $t\geq 1$, let $n_t^i$ be the number of times arm $i$ is pulled until and including time $t$. \\
	\vspace{0.1in}
	{\bf Procedure:}
	\begin{itemize}[wide, labelwidth=!, labelindent=0pt]
		\item {\bf Explore Phase:} From time $t=1$ until $t=K$, pull each arm once. For $K<t\leq T$: 
		\begin{enumerate}
			\item Identify $L_t \in \arg\max_{i \in [K]} \lcb^i_{n^i_{t-1}}$, breaking ties arbitrarily. If 
			\begin{align}
			\lcb_{n_{t-1}^{L_t }}^{L_t } > \max_{i \in [K]: i \neq L_t } \ucb^i_{n^i_{t-1}},\label{cond:stopping}
			\end{align}
			then define $i^* \triangleqq L_t$, break, and enter the Commit phase. Else, continue to Step 2.
			\item Identify $E_t \in \arg\max_{i \in [K]} \ucb^i_{n^i_{t-1}}$, breaking ties arbitrarily. Pull arm $E_t$. 
		\end{enumerate}
		\item {\bf Commit Phase:} Pull arm $i^*$ until time $t=T$.
	\end{itemize}
\vspace{0.3cm}
\end{algorithm}

The upper confidence bound is defined in Equation~\ref{ucb}. In contrast to its definition in UCB1, it is tuned to eliminate wasteful exploration and to allow stopping early if appropriate. We enforce the requirement that an arm is played at least $\tau$ times before committing to it by defining a trivial ``lower confidence bound" (Equation~\ref{lcb}),  which takes value $0$ until the arm is played less than $\tau$ times, after which both the upper and lower confidence bounds are defined to be the empirical mean of the arm. The stopping criterion can then be simply stated in terms of these upper and lower confidence bounds (Equation~\ref{cond:stopping}): stop and commit to an arm when its lower confidence bound is strictly higher than the upper confidence bounds of all other arms (this can never happen before $\tau$ pulls since the rewards are non-negative). 

Note that the collapse of the upper and lower confidence bounds to the empirical mean after $\tau$ pulls ensures that each arm is not pulled more than $\tau$ times during the Explore phase. This is because choosing this arm to explore after $\tau$ pulls would imply that its upper confidence bound = lower confidence bound is higher than the upper confidence bounds for all other arms, which means that the stopping criterion has been met  and the algorithm has committed to the arm.

\vspace{0.2cm}
\begin{remark}\label{rem:ub}
{\it A heuristic rationale behind the choice of the upper confidence bound is as follows. Consider a suboptimal arm whose mean is smaller than the highest mean by $\Delta$. Let $P_e$ be the probability that this arm is misidentified and committed to in the Commit phase. Then the expected regret resulting from this misidentification is approximately $P_e \Delta T$. Since we want to ensure that the regret is at most $\textup{O}(T^{2/3}K^{1/3})$ in the worst-case, we can tolerate a $P_e$ of at most $\approx K^{1/3}/(\Delta T^{1/3})$. Unfortunately, $\Delta$ is not known to the algorithm. However, a reasonable proxy for $\Delta$ is $1/\sqrt{n}$, where $n$ is the number of times the arm has been pulled. This is because it is right around $n\approx1/\Delta^2$, when the distinction between this arm and the optimal arm is expected to occur. Thus a good (moving) target for the probability of misidentification is $\delta_n \approx (K^{1/3}n^{1/2})/ T^{1/3}$. This necessitates the $\sqrt{\log (1/\delta_n)} \approx \sqrt{\log(T/(Kn^{3/2}))}$ scaling of the confidence interval in Equation~\ref{ucb}. In contrast, we numerically find that utilizing the traditional scaling of  $\sqrt{\log T}$ as in UCB1 results in significant performance deterioration. Our tuning is reminiscent of similar tuning of confidence bounds under the ``sum'' objective to improve the performance of UCB1 \citep{audibertminimax, auer2010ucb, lattimore2018refining}.}
\end{remark}
\vspace{0.2cm}
\begin{remark}\label{rem:lb}
{\it Instead of defining the lower confidence bound to be $0$ until an arm is pulled $\tau$ times, one may define a non-trivial lower confidence bound to accelerate commitment, perhaps in a symmetric fashion as the upper confidence bound. However, this doesn't lead to an improvement in the regret bound. The reason is that if an arm looks promising during exploration, then eagerness to commit to it is imprudent, since if it is indeed optimal then it is expected to be chosen frequently during exploration anyway; whereas, if it is suboptimal then we preserve the option of eliminating it by choosing to not commit until after $\tau$ pulls. Thus, to summarize, ADA-ETC eliminates wasteful exploration primarily by reducing the number of times suboptimal arms are pulled during exploration through the choice of appropriately aggressive upper confidence bounds, rather than by being hasty in commitment.}
\end{remark}
\vspace{0.2cm}

Let $\textup{ADA-ETC}_{K,T}$ denote the implementation of ADA-ETC using $K$ and $T$ as the input for the number of arms and the time horizon, respectively. We characterize the regret guarantees achieved by $\textup{ADA-ETC}_{K,T}$ in the following result. 
\begin{theorem}[ADA-ETC performance]
	\label{thm_stopping}
	Let $K<T$. Consider a $\bnu \in\mathcal{V}$ such that the optimal arm is unique and relabel arms so that $\mu_1> \mu_2\geq\dots\geq \mu_K$. Then the expected regret of $\textup{ADA-ETC}_{K,T}$ is upper bounded as:\footnote{We define $\log^+(a) = \log(\max(a,1))$ for $a>0$.}
	\begin{align*}
&\PSR_T(\textup{ADA-ETC}_{K,T}, \bnu)\\
&\leq \underbrace{\mu_1\sum_{i=2}^K \min\left(\frac{11}{\Delta_i^2} + \frac{16}{\Delta_i^2} \log^+\left(\frac{T\Delta_i^3}{K}\right)+\frac{24}{\Delta_i^2}\sqrt{\log^+\left(\frac{T\Delta_i^3}{K}\right)},\tau\right) + \mu_1\tau \sum_{i=2}^K \min(2,\frac{648K}{T\Delta_i^3})
}_{\textup{{\tiny Regret contribution from wasted pulls in the Explore phase}}}\\
&~~+\underbrace{\sum_{i=2}^K\exp(-\frac{\tau \Delta_i^2}{2})T\Delta_i+ \sum_{i=2}^K \min(1,\frac{320K}{T\Delta_i^3})T\left( \Delta_i-\Delta_{i-1} \right)}_{\textup{{\tiny Regret contribution from misidentification in the Commit phase}}},
	\end{align*}
where $\tau = \lceil\frac{T^{2/3}}{K^{2/3}}\rceil$.	
	In the worst case, we have
	\[
	\sup_{\bnu\in\mathcal{V}}\PSR_T(\textup{ADA-ETC}_{K,T}, \bnu) \leq \textup{O}(K^{1/3} T^{2/3} \sqrt{\log K}).
	\]
\end{theorem}

The proof of Theorem~\ref{thm_stopping} is presented in Section~\ref{apx:thm2} in the Appendix. Theorem~\ref{thm_stopping} features an instance-dependent regret bound and a worst-case bound of $\textup{O}(K^{1/3} T^{2/3} \sqrt{\log K})$. The first two terms in the instance-dependent bound arise from the wasted pulls during the Explore phase. Under vanilla Explore-then-Commit, to obtain near-optimality in the worst case, every arm must be pulled $\tau$ times in the Explore phase \citep{slivkins2019introduction}. Hence, the expected regret from the Explore phase is $\Omega(K\tau) = \Omega(T^{2/3}K^{1/3})$ irrespective of the instance. On the other hand, our bound on this regret depends on the instance and can be significantly smaller than $K\tau$ if the arms are easier to distinguish. In particular, for a fixed $K$ and $\nu$ (with $\Delta_2>0$), the regret from exploration (and the overall regret) is $\textup{O}(\sum_{i\geq 2}\mu_1\log T/\Delta_i^2)$ under ADA-ETC as opposed to $\Omega(T^{2/3}K^{1/3})$ under ETC as $T\rightarrow \infty$. This shows that ADA-ETC attains the instance-dependent lower bound on regret of Theorem~\ref{thm:asymplwb} up to a constant factor.

The next two terms in our instance-dependent bound arise from the regret incurred due to committing to a suboptimal arm, which can be shown to be $\textup{O}(K^{1/3} T^{2/3} \sqrt{\log K})$ in the worst case, thus matching the guarantee of ETC. The first of these terms is not problematic since it is the same as the regret arising under ETC. The second term arises due to the inevitably increased misidentifications occurring due to stopping early in adaptive versions of ETC. If the confidence bounds are aggressively small, then this term increases. In ADA-ETC, the upper confidence bounds used in exploration are tuned to be as small as possible while ensuring that this term is no larger than $\textup{O}(K^{1/3} T^{2/3})$ in the worst case (see Remark~\ref{rem:ub}). Thus, our tuning of the Explore phase ensures that the performance gains during exploration do not come at the cost of higher worst-case regret (in the leading order) due to misidentification.
\vspace{0.2cm}
\begin{remark}\label{rem:ucb}
{\it
It is possible to show that using the confidence bounds of UCB1 under ADA-ETC results in the same asymptotic instance-dependent regret bound of $\textup{O}(\sum_{i\geq 2}\mu_1\log T/\Delta_i^2)$ and an instance-independent regret bound of $\textup{O}(K^{1/3} T^{2/3} \sqrt{\log K})$ in the worst case. However, for fixed $T$ and $K$, the bounds derived for ADA-ETC, as defined, have an improved dependence on the instance owing to the reasons mentioned in Remark~\ref{rem:ub}. As we shall see in Section~\ref{sec:numerics}, this results in significant performance gains for practical values of $T$ and $K$. Optimizing finite $T$ performance is particularly important for our applications as discussed in Section~\ref{sec:intro}.} 
\end{remark}
\vspace{0.2cm}

%% file: gen_setup.tex
\section{The Case of $\boldsymbol{m}>1$} \label{sec:generalm}
Building on our observations in $m=1$ case, we extend our problem to settings where $m>1$. 
We let $K$ be the number of all available arms and suppose that the objective is to maximize the expected average cumulative reward across the top $m$ arms. 
With $T$ pulls and no additional constraints, the optimal policy in this problem is to always pull the arm with the highest mean -- essentially, the optimal policy is the same as that under the $m=1$ objective. However, as we discussed in Section~\ref{sec:intro}, such a policy is not practical for the applications we consider. We thus modify the problem by assuming that there are $T/m$ decision points, i.e., the time horizon is $T/m$, and at each time, $m$ distinct arms must be chosen (amounting to a total of $T$ pulls). For simplicity of notation, we assume that $T/m$ is an integer (this assumption comes at the cost of only a $\textup{O}(1)$ increase in regret). 

\begin{remark} {\it It is worth noting here that, while the optimal policy under full information without batching for any $m$ is the same as that under the max objective (m=1), the optimal regret under incomplete information differs depending on $m$. To see this, note that for $m=K$, the objective of maximizing the average total reward across all arms is equivalent to the classical sum objective, and thus the optimal instance-dependent, as well as the worst-case regret, is lower compared to the max objective, corresponding to the case of $m=1$. Thus, varying $m$ offers an interesting interpolation between the sum and the max objectives, and characterizing the optimal regret on this spectrum is an interesting direction for future work.} 
\end{remark}


We reuse the notation for the distribution of the arms and their means: $\nu_i$ denotes the probability distribution for arm $i \in [K]$ and $\mu_i$ denotes its mean. We label the arms so that $\mu_1\geq \mu_2\geq\dots\geq \mu_K$ (breaking ties arbitrarily), and we let $\bnu$ and $\bmu$ denote the vector of probability distributions and their means, respectively.  We refer to the arms in the set $[m]$ as the optimal arms and the rest of them as the suboptimal arms.
Different from the $m=1$ case, here we define two measures to capture the difference between the means of optimal arms and the suboptimal arms: $\bar{\Delta}_i = \mu_i -\mu_{m+1}$ for $i \in [m]$ and $\Delta_j = \mu_{m} - \mu_j$ for $j \geq m+1$. That is, $\bar{\Delta}_i$ for each optimal arm $i$ is the difference between the mean of $i$ and the that of the best suboptimal arm.   $\Delta_j$, on the other hand, for each sub-optimal arm $j$, is the difference between the mean of $j$ and that of the worst optimal arm.
These two measures will be crucial in the analysis of this problem.

At each time, the decision-maker chooses exactly $m$ arms to play and observes a reward from each played arm. With some abusive reuse of notation, we let the vector $\boldi_t$ denote the set of $m$ arms played at time $t$ and let the vector $\bx_t$ denote the rewards from those arms. With more abuse of notation, we then have $\bx_t(i) \sim \nu_{\boldi_t(i)}$ for $i \in \boldi_t$, and all these rewards are assumed to be independent from the previous actions and observations.

We again let the history of actions and observations at any time $t\geq 2$ be denoted as $\mathcal{H}_t = (\boldi_1, \bx_1, \boldi_2, \bx_2,\dots, \boldi_{t-1}, \bx_{t-1})$, and define  $\mathcal{H}_1$ to be the empty set $\phi$. Hence, a policy $\pi$ of the decision-maker is a sequence of mappings $(\pi_1,\pi_2,\dots,\pi_{\Tminline})$, in which $\pi_t$ maps every possible history $\mathcal{H}_t$ to a set of $m$ arms to be played at time $t$. We let $\Pi$ denote the set of all such policies.

Recall that $n^i_t$ denotes the number of times arm $i$ is played until and including time $t$, i.e., $n^i_t = \sum_{s=1}^t \mathbbm{1}_{\{ i \in \boldi_s \}}$, and $U^i_n$ denotes the reward observed from the $n^{\textup{th}}$ pull of arm $i$. Note that $(U^i_n)_{n\in\mathbb{N}}$ is a sequence of i.i.d. random variables, with each $U^i_n$ distributed as $\nu_i$. Finally, let $\overline{U}^i_{t}$ be the cumulative reward obtained from arm $i$ until time $t$. 


We now consider the following notion of reward for a policy $\pi$:
\begin{align}
\rew_T(\pi,\bnu) = \E_{\bnu}\big(\obj\big(\overline{U}^1_{\Tminline}, \overline{U}^2_{\Tminline}, \dots, \overline{U}^K_{\Tminline}\big)\big),
\end{align}
where $\obj(\mathbf{x})$ denotes the average of the largest $m$ elements in $\mathbf{x}\in\mathbb{R}^K$, for an integer $m$ with $1\leq m\leq K$: \[  \obj(\mathbf{x}) \triangleqq \frac{1}{m} \sum_{i = 1}^m \mathbf{y}(i), \] where the vector $\mathbf{y}$ is vector $\mathbf{x}$ sorted in non-increasing order (breaking ties arbitrarily).

In other words, the objective value attained by the policy is the expected value of the average of the $m$ largest cumulative rewards across all arms at the end of the decision-making horizon. 
When the reward distributions $\nu_1, \dots, \nu_K$ are known to the decision-maker, then for a large $T$, the best reward that the decision-maker can achieve is
\[\sup_{\pi\in\Pi} \rew_T(\pi, \bnu).\]

In a similar spirit to the $m=1$ case, a natural candidate for a good policy when the reward distributions are known is the one where the decision-maker focuses on the top $m$ arms with the highest means, attaining an expected reward of $\bestm \frac{T}{m}$, where $\bestm \triangleq \frac{1}{m} \sum_{i = 1}^m \mu_i$, the average mean of the $m$ highest mean arms. Let us denote $\rewR^*_T(\bnu) \triangleqq \bestm \frac{T}{m}$. One can show a result similar to Proposition~\ref{prop:main} here too: $\rewR^*_T(\bnu)$ is the best reward that one can achieve in our problem.

\begin{prop}
	\label{prop_general:main} For any bandit instance $\bnu\in\mathcal{V}$, $\sup_{\pi\in\Pi} \rew_T(\pi,\bnu) = \rewR^*_T(\bnu)$.
\end{prop}

The proof is presented in Section~\ref{apx_general:prop1} in the Appendix. This shows that the policy that picks the $m$ arms with the highest means in all periods is optimal. Next, we denote the regret of any policy $\pi$ to be \[\PPSR_T(\pi, \bnu) = \sup_{\pi\in\Pi} \rew_T(\pi, \bnu)- \rew_T(\pi,\bnu) .\]
We once again focus on finding a policy $\pi$ that achieves the smallest $\bnu$-dependent asymptotic regret (as $T\rightarrow \infty$) simultaneously for all $\bnu$, and also the smallest the worst-case regret over all distributions $\bnu\in\mathcal{V}$ for a fixed $T$. 
To the latter end, let $\PPSR^*_T$ denote the \textit{minmax} (or the \textit{best worst-case}) regret: 
\[\PPSR^*_T \triangleqq \inf_{\pi\in\Pi}\sup_{\bnu\in\mathcal{V}}\PPSR_T(\pi, \bnu).\]

In the remainder of this section, we will show that a regret of $\Omega(\frac{(K-m)^{1/3} T^{2/3}}{m^{4/3}})$ is inevitable in the worst case. We then will design a policy that attains this regret with a mildly weaker dependence on $m$.

\subsection{Lower Bound} \label{sec:general_lb}
It is clear from the results of the $m=1$ case that an $\Omega(\log T)$ instance-dependent regret and $\Omega(T^{2/3})$ instance-independent regret is inevitable for any policy.  In the following result, we try to capture the dependence on $m$ in the lower bound on the optimal instance-independent regret. 

\begin{theorem}\label{thm_general:lb}
	Suppose that $2m \leq K < T$. Then, $\PPSR^*_T \geq  \Omega(\frac{(K-m)^{1/3} T^{2/3}}{m^{4/3}}).$
\end{theorem}
The proof is presented in Section~\ref{apx_general:thm_lb} in the Appendix and it extends the proof for the $m=1$ case while tackling new challenges to capture the dependence on $m$.
However, we conjecture that the dependence on $m$ in this bound is sub-optimal. In the next section, we will present an algorithm that attains a regret upper bound of $\widetilde{\textup{O}}(\frac{(K-m)^{1/3} T^{2/3}}{m})$, which we believe is the best achievable.\footnote{Intuitive reasoning for the improved lower bound is as follows. Assume $m$ divides $K$ and $K/m \geq 2$. Then, an adversary can construct $m$ independent bandit sub-problems with $K/m$ arms in each problem such that the best arm needs to be identified and exploited in each of the $m$ problems to optimize our original objective. If the decision-maker additionally has the constraint that $T/m$ pulls can be expended in each sub-problem, then the lower bound from the $m=1$ case would imply that a regret of $\Omega\big((K/m-1)^{1/3} (T/m)^{2/3}\big) = \Omega(\frac{(K-m)^{1/3} T^{2/3}}{m})$ is inevitable in each sub-problem and thus inevitable in the original problem. However, this argument assumes the constraint of $T/m$ pulls per sub-problem, which may induce an avoidable loss.} We leave the closure of this gap as an interesting open question for future work.

%% file: gen_algo.tex
\subsection{Adaptive Explore-then-Commit for General $\boldsymbol{m}$ ($\boldsymbol{m}$-ADA-ETC)}\label{sec:algo_general} 
We now present the algorithm we design for the $m>1$ case. This is an extension of the ADA-ETC policy that we call  $m\textup{-ADA-ETC}$. It is formally defined in Algorithm~\ref{alg:main_general}. 

This algorithm shares a similar design logic with its $m=1$ counterpart: after choosing each arm at least once, pull $m$ arms with the highest upper confidence bounds, until there are $m$ arms such that (a) they all have been played at least $\tau= \lceil T^{2/3}/(K-m)^{2/3}\rceil$ times, and (b) the smallest empirical mean among them is higher than the upper confidence bounds on the means of all other arms. Then, commit to these $m$ arms until the end of the decision-making horizon. 

The upper confidence bound defined in Equation~\ref{gen_ucb} is similar to that we define in Equation~\ref{ucb} for $m=1$ case, with a modified dependence on the problem parameters, and aims to eliminate wasteful exploration by stopping early. Additionally, the design of the lower confidence bound in Equation~\ref{gen_lcb} and the stopping criterion in Equation~\ref{cond:stopping_gen} again enforce the requirement that all arms must be played at least $\tau$ times before being committed to by the algorithm.
However, different from the $m=1$ case, some arms may be pulled more than $\tau$ times while the algorithm is still in the Explore phase. Although this doesn't mean that those pulls are wasteful: we can show that the arms that get pulled more than $\tau$ pulls during exploration are the ones that will be included in the set of exploited arms in the Commit phase (see Lemma~\ref{lma:exploit} in the Appendix). This fact can be interpreted as meaning that a subset of arms may enter the Commit phase earlier than the others.
In accordance, we introduce arm-specific exploration stopping times in our proof of performance guarantees of Algorithm~\ref{alg:main_general}. 


\begin{algorithm}[h]
	\caption{Adaptive Explore-then-Commit for General $m$ ($m\textup{-ADA-ETC}$)}\label{alg:main_general}
	\vspace{0.2cm}
	{\bf Input:} $K$ arms, $m$ to be ultimately selected, and horizon $T$. \\
	{\bf Define:} $\tau = \lceil\frac{T^{2/3}}{\left(K-m\right)^{2/3}}\rceil$. For $n\geq 1$, let $\bar{\mu}_n^i$ be the empirical average reward from arm $i$ after $n$ pulls and it remains fixed after $\tau$ pulls, i.e., $\bar{\mu}_n^i = \frac{1}{\min\{n, \tau\}} \sum_{s=1}^{\min\{n, \tau\}} U_s^i$. 
	Also, for $n\geq 1$, define,
	\begin{align}
		\ucb^i_n &= \bar{\mu}_n^i + \sqrt{\frac{4}{n}\log\left(\frac{T}{(K-m)n^{3/2}}\right)}\mathbbm{1}_{\left\{ n< \tau \right\}}.\label{gen_ucb}\\
		\lcb^i_n &=  \bar{\mu}_n^i - \bar{\mu}_n^i\mathbbm{1}_{\{n< \tau\}}.\label{gen_lcb}
	\end{align}
	Also, for $t\geq 1$, let $n_t^i$ be the number of times arm $i$ is pulled until and including time $t$. \\
	\vspace{0.1in}
	{\bf Procedure:}
	\begin{itemize}[wide, labelwidth=!, labelindent=0pt]
		\item {\bf Explore Phase:} From time $t=1$ until $t=\lceil\Kminline\rceil$, pull each arm once. For $\lceil\Kminline\rceil<t\leq \lfloor\Tminline\rfloor$: 
		\begin{enumerate}
			\item Identify $U_t$, the $(m+1)^\textup{st}$ largest element of $\left( \ucb^1_{n^1_{t-1}}, \dots, \ucb^K_{n^K_{t-1}} \right)$, breaking ties arbitrarily. Define $\bolde_t = \{i \in [K] ~|~  \ucb^i_{n^i_{t-1}} > \ucb_{n_{t-1}^{U_t}}^{U_t}\}$. If 
			\begin{align}
				\min_{i \in \bolde_t} \lcb_{n_{t-1}^i}^i > \max_{j \in [K]\setminus \bolde_t} \ucb^j_{n^j_{t-1}},   \label{cond:stopping_gen}
			\end{align}
			then let $\boldi^* = \bolde_t$; break, and enter the Commit phase. Else, continue to Step 2. \\
			\item Pull all arms in $\bolde_t$ once. 
		\end{enumerate}
		\item {\bf Commit Phase:} Pull all arms in $\boldi^*$ until time $t=\lfloor\Tminline\rfloor$.
	\end{itemize}
	\vspace{0.3cm}
\end{algorithm}

\begin{remark}\label{rem:ub_gen}
{\it The parallel between the designs for the upper confidence bounds, Equation~\ref{gen_ucb} to that of in Equation~\ref{ucb} for the $m=1$ case, follows from a similar heuristic rationale. Consider the following example: Let $m \leq K/2$. The $m$ optimal arms all have a mean of $1$, and the remaining $K-m$ arms have a mean of $1-\Delta$. Let $P_e$ denote the probability of labeling some fixed sub-optimal arm $j$ as optimal (i.e., one of the top $m$ arms). Then, the expected regret contributed due to this error is approximately $P_e \frac{\Delta}{m} \frac{T}{m}$, since our objective considers the average reward from top $m$ arms with the highest cumulative rewards. But the event of incorrectly labeling $j$ as optimal can occur by displacing any one of the top $m$ arms. We can thus approximately bound the probability of this event by the probability of the event that the error in the mean estimate of any one of the top $m$ arms is of order $\Delta$ during exploration. If we let $P_e(i)$ be the probability of such an event for an optimal arm $i$, then by a union bound, $P_e \lessapprox \sum_{i = 1}^m P_e(i)$. This implies that the expected regret due to misidentifying $1$ arm is approximately upper bounded as $\sum_{i = 1}^m P_e(i) \Delta \frac{T}{m^2}$. Now, to ensure that the regret due to this misidentification is at most $\textup{O}(\frac{(K-m)^{1/3} T^{2/3}}{m})$, we can have $P_e(i)$ at most $\approx (K-m)^{1/3}/(\Delta T^{1/3})$, $i \in [m]$. Since $\Delta$ is not known to the algorithm, we again use $1/\sqrt{n}$ as a proxy for $\Delta$. Then, the target for this probability of misidentification is $\delta_n \approx ((K-m)^{1/3}n^{1/2})/ T^{1/3}$. Hence, we get the relevant scaling of the confidence bound in Equation~\ref{gen_ucb}, i.e., $\sqrt{\log (1/\delta_n)} \approx \sqrt{\log(T/((K-m)n^{3/2}))}$. }
\end{remark}


Let $m\textup{-ADA-ETC}_{K,T}$ denote the implementation of $m\textup{-ADA-ETC}$ using $m$, $K$ and $T$ as the input for the number of arms to be selected, the total number of available arms, and the total number of assignments, respectively. We characterize the regret guarantees it achieves in the following result. 

\begin{theorem}[$\boldsymbol{m}\textup{-ADA-ETC}$ performance]
	\label{thm_stopping_generalm}
	Let $K<T$. Consider a $\bnu \in\mathcal{V}$ such that there is a unique set of $m$ optimal arms and relabel arms so that $\mu_1\geq \mu_2\geq\dots\geq \mu_m > \mu_{m+1} \geq \dots \geq \mu_K$. Then the expected regret of $m\textup{-ADA-ETC}$ is upper bounded as:
	\begin{align*}
		&\PPSR_T(m\textup{-ADA-ETC}_{K,T}, \bnu) \leq\\
		&~\underbrace{\frac{1}{m}\sum_{i=1}^m\mu_i \min\left(\kappa, ~ \frac{320(K-m)^2\tau}{mT \bar{\Delta}_i^3}, ~\frac{2(K-m)\tau}{m} \right) 
		+ \frac{K-m}{m}\tau\min\big(\sum_{j=m+1}^K \mu_j \min (1,  \frac{328 (K-m)}{T \Delta_j^3}), ~\mu_m  \big)}_{\textup{{\scriptsize Regret contribution from wasted pulls in the Explore phase}}}\\
		&~~+\underbrace{\frac{T}{m^2} \sum_{i = 1}^m \sum_{j = m+1}^K (\mu_i-\mu_j) \min\left( \min\left(\frac{320 (K-m)}{T \bar{\Delta}_i^3}, \frac{320 (K-m)}{T \Delta_j^3}\right) + \frac{3}{2}\exp(-\frac{\tau \bar{\Delta}_i^2}{2}) + \frac{3}{2}\exp(-\frac{\tau \Delta_j^2}{2}), 1 \right)}_{\textup{{\scriptsize Regret contribution from misidentification in the Commit phase}}},
	\end{align*}
	where $\kappa = \sum_{j = m+1}^K \frac{11}{\Delta_j^2} + \frac{16}{\Delta_j^2} \log^+\big(\frac{T\Delta_j^3}{K-m}\big) + \frac{24}{\Delta_j^2}\sqrt{\log^+\big(\frac{T\Delta_j^3}{K-m}\big)}$, and 
	$\tau = \lceil\frac{T^{2/3}}{\left(K-m\right)^{2/3}}\rceil$.	
	In the worst case, we have
	\[
	\sup_{\bnu\in\mathcal{V}}\PPSR_T(m\textup{-ADA-ETC}_{K,T}, \bnu) \leq \textup{O}\Big(\frac{(K-m)^{1/3} T^{2/3} \sqrt{\log (K-m)}}{m}\Big).
	\]
\end{theorem}

The proof of Theorem~\ref{thm_stopping_generalm} is presented in Section~\ref{apx_general:thm_ub_gen} in the Appendix. Theorem~\ref{thm_stopping_generalm} features an instance-dependent $\textup{O}(\log T)$ regret bound and a worst-case bound of $\textup{O}(\frac{(K-m)^{1/3} T^{2/3} \sqrt{\log (K-m)}}{m})$. The first two terms in the instance-dependent bound arise from the wasted pulls during the Explore phase. The first term is due to exploration of the suboptimal arms and it specifies the pulls lost from the optimal arms while exploring. The second term is similar but it quantifies the loss in the case where the algorithm commits to a suboptimal set of arms. Both of these terms are at most $\frac{K-m}{m}\tau \mu_m \leq \frac{K-m}{m}\tau = \textup{O}(\frac{(K-m)^{1/3} T^{2/3}}{m})$ in the worst case. The next term arises from the regret incurred to committing to some number of suboptimal arms. We can show that this regret is $\textup{O}\Big(\frac{(K-m)^{1/3} T^{2/3} \sqrt{\log (K-m)}}{m}\Big)$ in the worst case. 

%% file: numerics.tex
\section{Numerical Experiments} \label{sec:numerics}
In this section, we benchmark our proposed algorithms against candidate algorithms in the literature. First, we highlight the differences in the instance-independent regret under the sum and the max objectives, validating the claim from Theorem~\ref{thm:sumbad} that policies achieving optimal worst-case regret under the sum objective are necessarily suboptimal for the max objective. We then show that, even for fixed instances, although existing algorithms achieve the order-optimal $O(\log T)$ instance-dependent regret bound for the max objective, our algorithms achieve significantly improved finite $T$ performance owing to the refined tuning of the exploration phase and early commitment in harder instances. Finally, we show how our algorithms can be implemented in the dynamic setting of online labor platforms and discuss certain tradeoffs that may arise.


\subsection{A study of instance-independent regret}
As we have shown in our technical results, the optimal instance-independent regret for the max objective is $\tilde\Theta(T^{2/3}K^{1/3})$, different from the $\tilde\Theta(\sqrt{KT})$ achievable for the sum objective. The goal of our first set of numerical experiments is to validate the claim of Theorem~\ref{thm:sumbad} that policies that achieve (up to $\log$ factors) the optimal regret for the sum objective have a suboptimal worst-case regret performance for the max objective. 

\begin{figure}[h]
	\begin{subfigure}{0.47\textwidth}
		\centering
		\includegraphics[width=\textwidth]{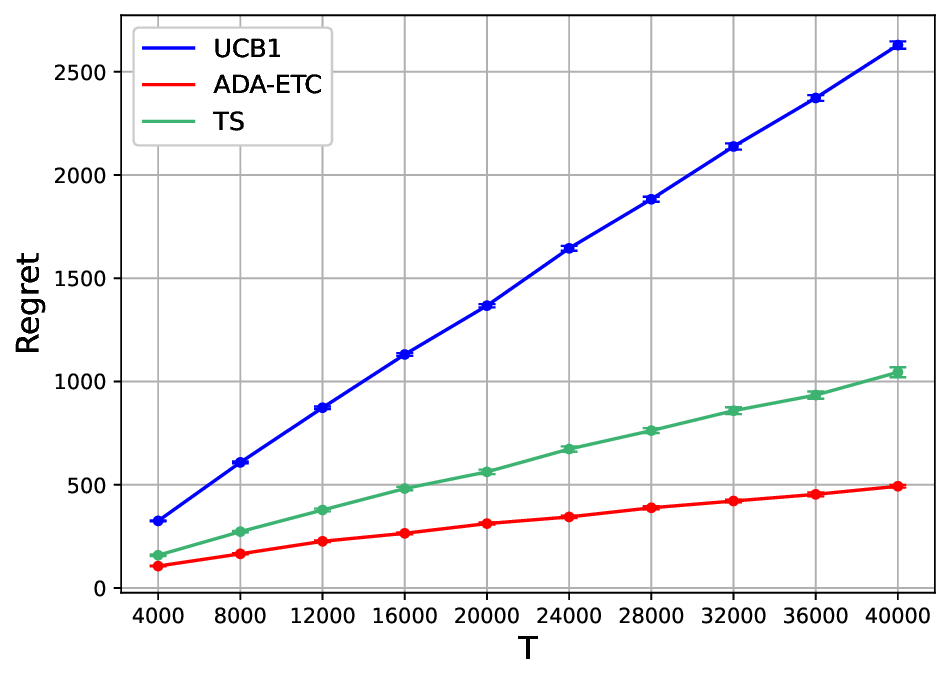}
		\caption{Max objective}
	\end{subfigure}%
	\hfill
	\begin{subfigure}{0.47\textwidth}
		\centering
		\includegraphics[width=\textwidth]{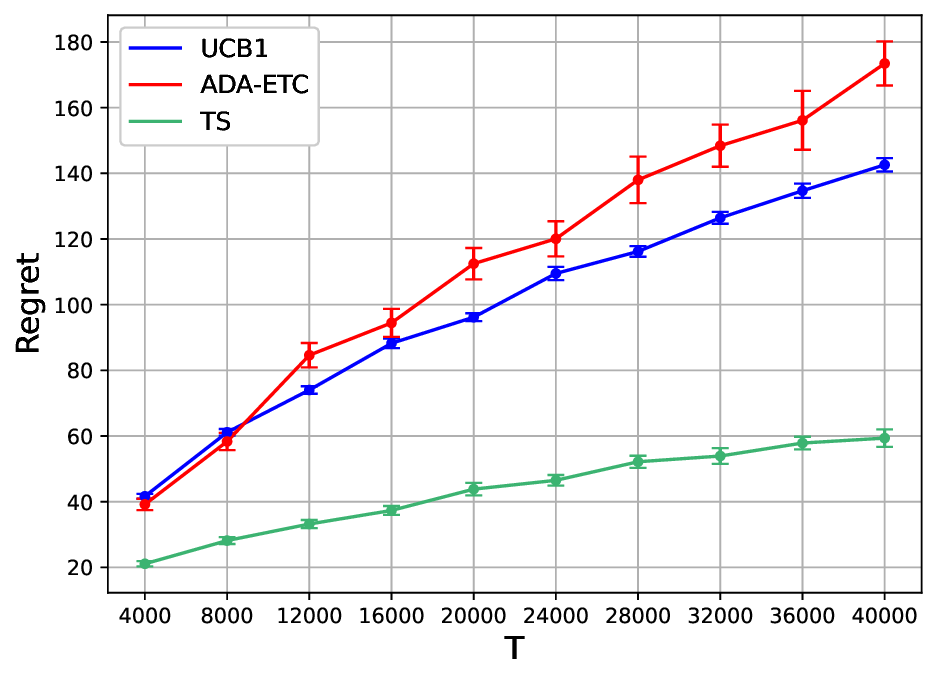}
		\caption{Sum objective}
	\end{subfigure} 
	\caption{Average sum and max regret for 2-armed bandit instances with a mean-reward gap of $1/T^{2/5}$.} \label{fig:trendReversal}
\end{figure}

This distinction can be numerically observed through the following set of instances: consider a sequence of two-armed bandit problems with Bernoulli rewards and means $1/2$ and $1/2 +1/T^{2/5}$, where $T$ is the number of available pulls. Figure~\ref{fig:trendReversal} (left) shows the regret performance of ADA-ETC under the max objective, benchmarked against the two popular policies designed for the sum objective: 
UCB1 
and Thompson Sampling \citep{thompson1933likelihood, russo2018tutorial}. Figure~\ref{fig:trendReversal} (right) shows the regret performance of these policies under the sum objective. 

First, observe that the performance of ADA-ETC is worse than these policies for the sum objective. This is natural since ADA-ETC explores only for $\Theta(T^{2/3})$ pulls before committing to an arm, which is not sufficient to distinguish arms whose means are separated by $1/T^{2/5}$ (for which at least $\Omega(T^{4/5})$ pulls are necessary). Thus ADA-ETC makes an identification error with a constant probability, incurring a regret of $1/T^{2/5}\times \Omega(T) = \Omega(T^{3/5})$, which is higher than the near-optimal worst-case performance of $\tilde\Theta(\sqrt{T})$ achieved by the other algorithms. 

\begin{figure}[h]
	\begin{subfigure}{0.47\textwidth}
		\centering
		\includegraphics[width=\textwidth]{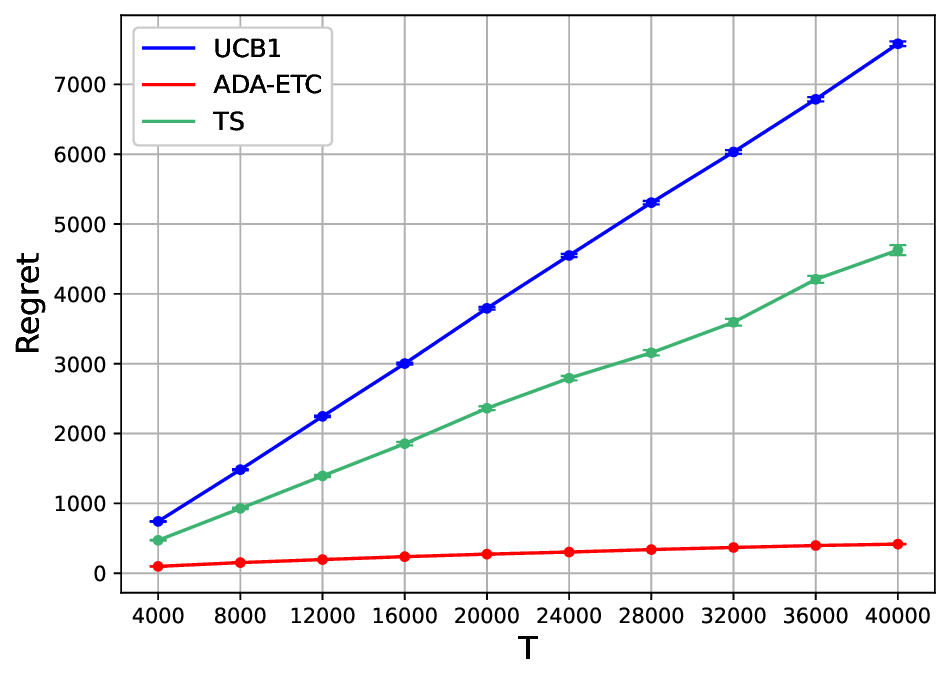}
		\caption{Max objective}
	\end{subfigure}%
	\hfill
	\begin{subfigure}{0.47\textwidth}
		\centering
		\includegraphics[width=\textwidth]{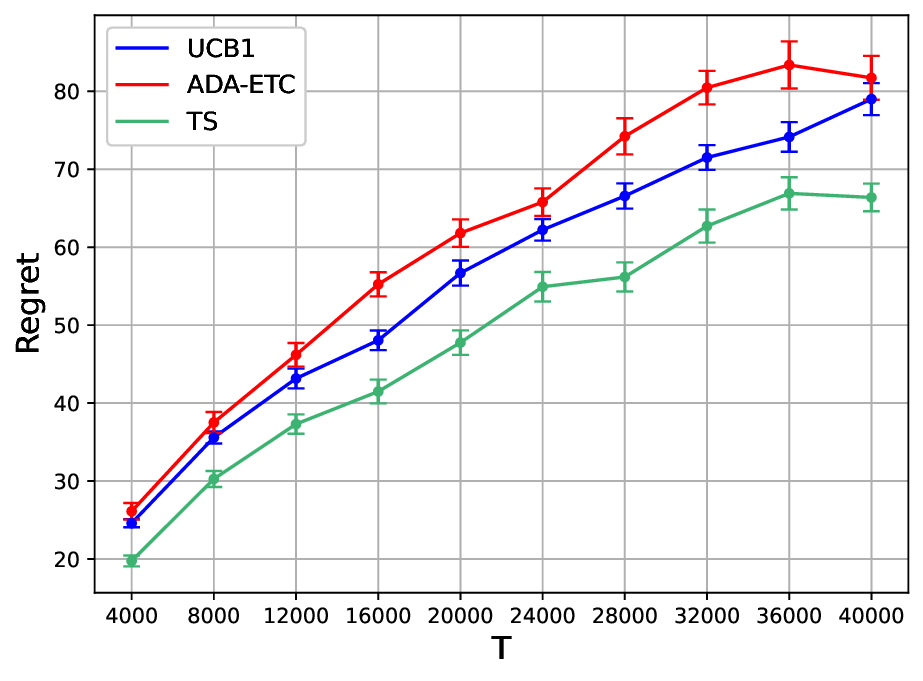}
		\caption{Sum objective}
	\end{subfigure} 
	\caption{Average sum and max regret for 2-armed bandit instances with a mean-reward gap of $1/\sqrt{T}$.} \label{fig:trendReversal2}
\end{figure}

For the max objective, however, the performance of ADA-ETC is significantly better compared to all policies. This is because to keep the sum regret bounded by $\tilde{\textup{O}}(\sqrt{T})$, all the other algorithms must distinguish between the two arms (since the gap is too large), and hence they explore both arms for time $\tilde\Omega(T^{4/5})$, incurring a regret of $\Omega(T^{4/5})$ for the max objective. The sub-optimal performance of the benchmark algorithms for the max objective is much more starkly demonstrated through a sequence of two-armed bandit problems whose means are separated by $1/\sqrt{T}$. Figure~\ref{fig:trendReversal2} shows this comparison, which suggests that the regret of these benchmark algorithms is linear in $T$. We, in fact, prove that this is the case for UCB1.

\begin{prop}\label{prop:ucbbad}
Consider a sequence (indexed by $T$) of two-armed bandit instances with Bernoulli rewards and means $1/2$ and $1/2+1/\sqrt{T}$, respectively. Consider a UCB policy where the upper confidence bound for an arm $i$ after $n$ pulls is defined to be its empirical mean plus the quantity $c\sqrt{\frac{\log T}{n}}$ for any $c>1/\sqrt{2}$. Then the regret of this policy for this sequence of instances is $\Theta(T)$.
\end{prop}
Since the exploration constant $c$ is at least $1$ for the order-optimal sum-regret bound of UCB1 to hold, this implies that UCB1 incurs a worst-case regret of $\Theta(T)$ for the max objective. The proof can be found in Section~\ref{apx:ucbbad} in the Appendix.

\subsection{A study of regret for fixed instances: $m=1$ }\label{sec:m1exp}
We compare the performance of ADA-ETC with four algorithms described in Table~\ref{algorithms}. 
UCB1 never stops exploring and pulls the arm with the highest upper confidence bound at each time step. 
TS (Thompson Sampling, \cite{thompson1933likelihood}) also never commits to an arm and pulls an arm based on the environment sampled from the posterior. ETC pulls arms in a round-robin fashion and commits to the arm with the highest empirical mean after each arm has been pulled $\tau$ times. Naive ADA-ETC (NADA-ETC) has the same algorithmic structure as ADA-ETC: it explores based on upper confidence bounds and commits if the lower confidence bound of an arm rises above upper confidence bounds for all other arms. It differs from ADA-ETC in how the upper confidence bounds are defined -- it uses the same upper confidence bound as UCB1. These definitions are presented in Table~\ref{algorithms}.  

\begin{table}[h]
	\caption{Benchmark Algorithms}
	\label{algorithms}
	\centering\scriptsize{
		\begin{tabular}{lll}
			\toprule
			\vspace{-0.25cm}\\
			\multirow{2}{*}{ADA-ETC}&$\ucb^i_n=$ $\bar{\mu}_n^i + \sqrt{\frac{4}{n}\log\left(\frac{T}{Kn^{3/2}}\right)}\mathbbm{1}_{\left\{ n< \tau \right\}}$\\\vspace{0.15cm}&$\lcb^i_n=$ $\bar{\mu}_n^i - \bar{\mu}_n^i\mathbbm{1}_{\{n< \tau\}}$\\
			\multirow{2}{*}{ETC}&$\ucb^i_n=$ $*$\\\vspace{0.15cm}&$\lcb^i_n=$ $*$\\
			\multirow{2}{*}{TS}&$\ucb^i_n=$ $*$\\\vspace{0.15cm}&$\lcb^i_n=$ $*$\\
			\multirow{2}{*}{UCB1}&$\ucb^i_n=$ $\bar{\mu}_n^i + \sqrt{\frac{1}{n}\log\left(T\right)}$\\\vspace{0.15cm}&$\lcb^i_n=$ $*$\\
			\multirow{2}{*}{NADA-ETC}&$\ucb^i_n=$ $\bar{\mu}_n^i + \sqrt{\frac{1}{n}\log\left(T\right)}\mathbbm{1}_{\left\{ n< \tau \right\}}$\\\vspace{0.15cm}&$\lcb^i_n=$ $\bar{\mu}_n^i - \bar{\mu}_n^i\mathbbm{1}_{\{n< \tau\}}$\\
			\vspace{-0.25cm}\\
			\bottomrule
		\end{tabular}
	}
\end{table}


\begin{figure}[h]
	\centering
	\begin{subfigure}{0.48\textwidth}
		\centering
		\includegraphics[width=\textwidth]{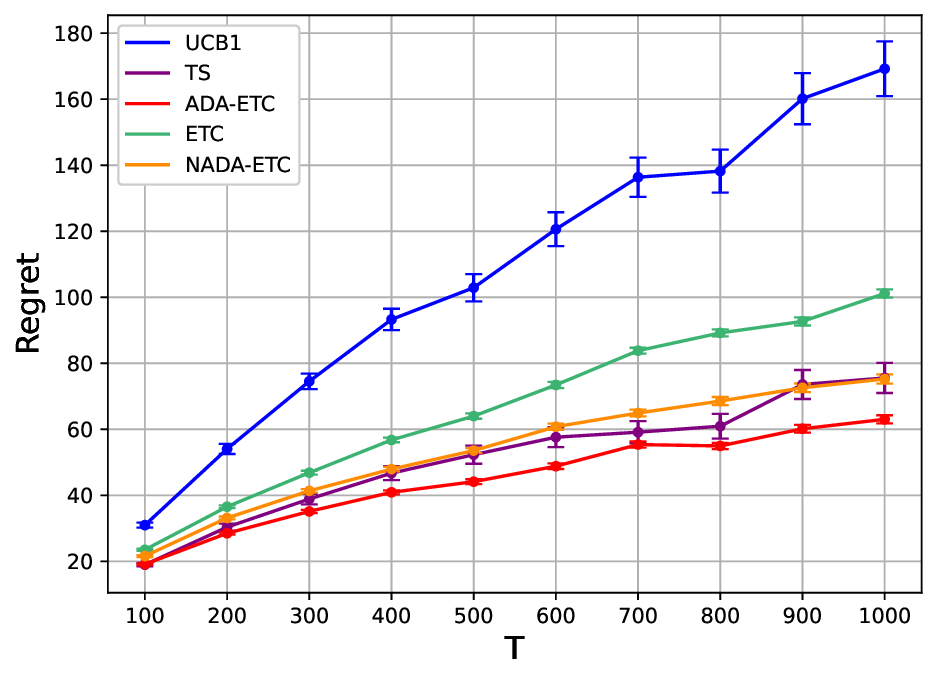}
		\caption{$K=4$, $\alpha = 0$}
	\end{subfigure}%
	\hfill
	\begin{subfigure}{0.48\textwidth}
		\centering
		\includegraphics[width=\textwidth]{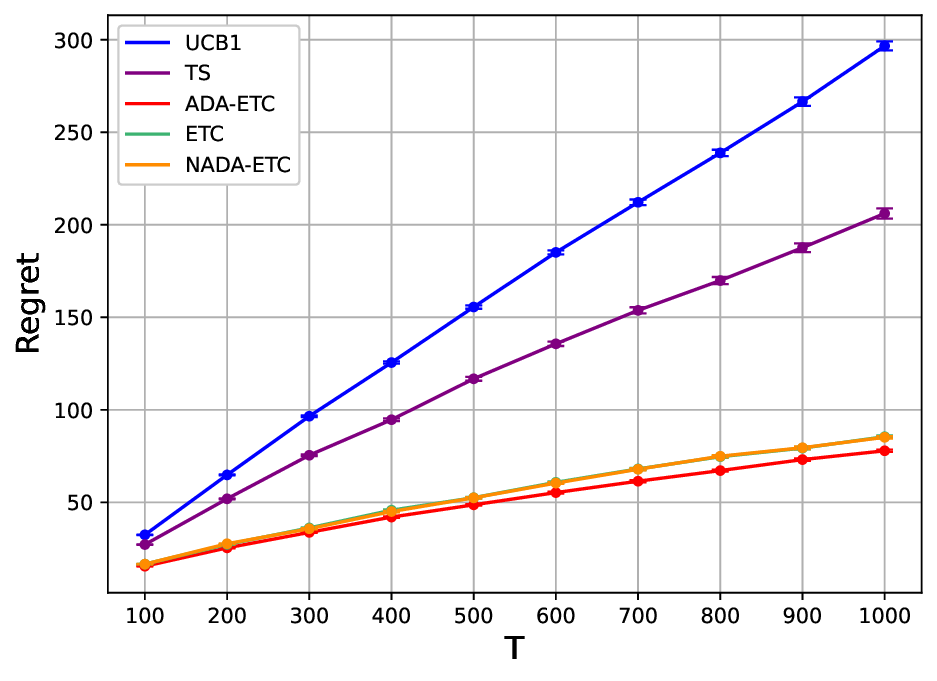}
		\caption{$K=4$, $\alpha = 0.4$}
	\end{subfigure} 
	\vspace{0.1cm}
	
	\begin{subfigure}{0.48\textwidth}
		\centering
		\includegraphics[width=\textwidth]{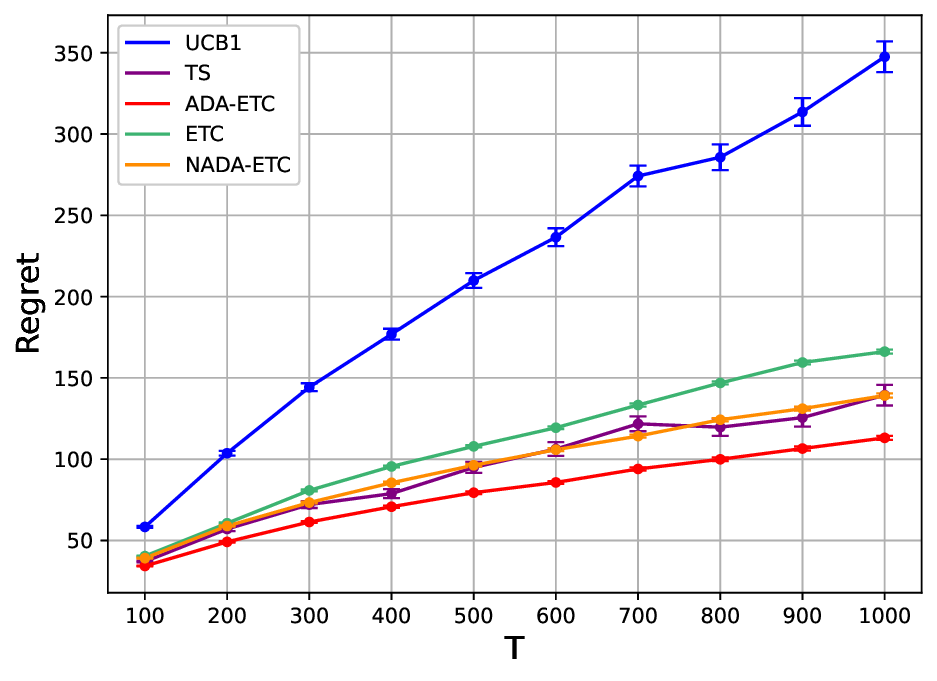}
		\caption{$K=8$, $\alpha = 0$}
	\end{subfigure}%
	\hfill
	\begin{subfigure}{0.48\textwidth}
		\centering
		\includegraphics[width=\textwidth]{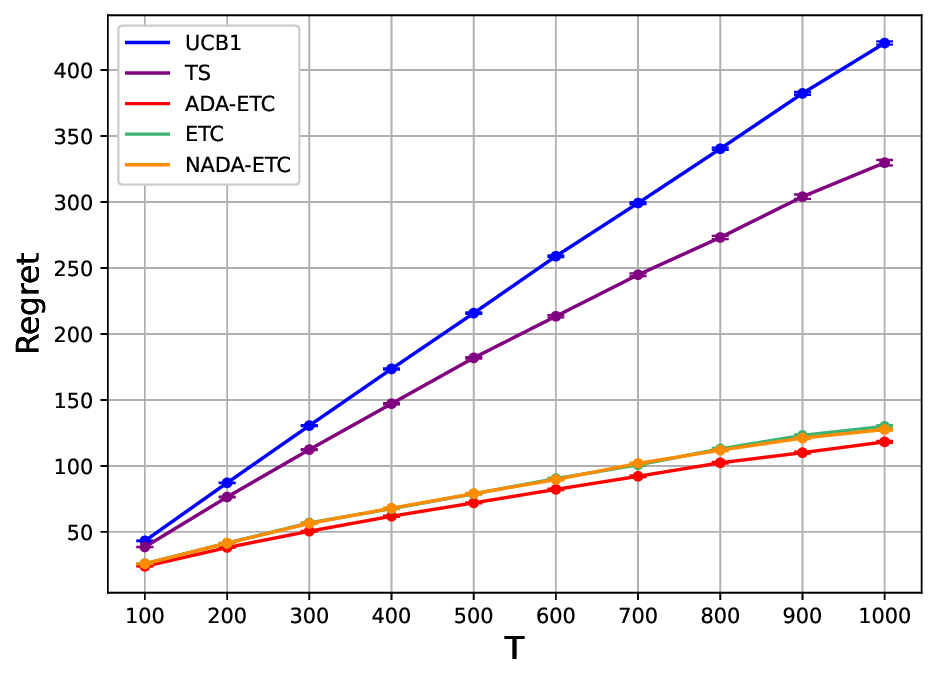}
		\caption{$K=8$, $\alpha = 0.4$}
	\end{subfigure} 

\caption{Performance comparison of ADA-ETC for varying values of $T$.} \label{fig:experiments}
\end{figure}

\noindent{\bf Instances.} We let $\nu_i \sim \textup{Bernoulli}(\mu_i)$, where $\mu_i$ is uniformly sampled from $[\alpha, 1-\alpha]$ for each arm in each instance. We sample two sets of instances, each of size $200$, with $\alpha \in\{0, 0.4\}$. The regret for an algorithm for each instance is averaged over $50$ runs to estimate the expected regret.
We vary $K \in \{4, 8\}$ and $T \in \{100, \dots, 1000\}$. The average regret over the $200$ instances under different algorithms and settings is presented in Figure~\ref{fig:experiments}.

\noindent{\bf Discussion.} ADA-ETC shows the best performance uniformly across all settings, although there are settings where its performance is similar to ETC and NADA-ETC. These are settings where either (a) $\alpha=0.4$, in which case, the arms are expected to be close to each other and hence adaptivity in exploring has little benefits, or (b) 
$T/K$ is relatively small, due to which $\tau$ is small. In these latter situations, the exploration budget of $\tau$ is expected to be exhausted for almost all arms under ADA-ETC, yielding a performance more similar to ETC, e.g., if $K=8$ and $T = 100$, then $\tau = \lceil 12.5^{2/3}\rceil = 6$, i.e., a maximum of six pulls can be used per arm for exploring. When $\alpha$ is smaller, i.e., when arms are easier to distinguish, or when $\tau$ is large, the performance of ADA-ETC is significantly better than those of ETC and NADA-ETC. 
This illustrates the gains from the adaptivity of exploration under ADA-ETC.

It is interesting to note that while NADA-ETC improves over the performance of ETC, the gains are much more under ADA-ETC. For example, the performance of ETC and NADA-ETC are virtually the same when $\alpha=0.4$, whereas ADA-ETC performs strictly better. This observation suggests that naively adding adaptivity to exploration, e.g., based on UCB1's upper confidence bounds, may not lead to significant improvements over the performance of ETC in finite parameter settings, and appropriate refinement of the confidence bounds is crucial to the gains of ADA-ETC. 
We note that UCB1's performance is consistently poor despite its asymptotic optimality, suggesting inferior instance-dependent performance for these instances in finite $T$ settings. In comparison, TS performs well when $\alpha=0$, but its performance deteriorates for $\alpha=0.4$ when arms are more difficult to distinguish since it becomes more likely that more than one arm is often pulled in the long run. These observations demonstrate the importance of introducing an appropriate stopping criterion for exploration to achieve robust performance.

\begin{figure}[h]
	\centering
	\begin{subfigure}{0.32\textwidth}
		\centering
		\includegraphics[width=\textwidth]{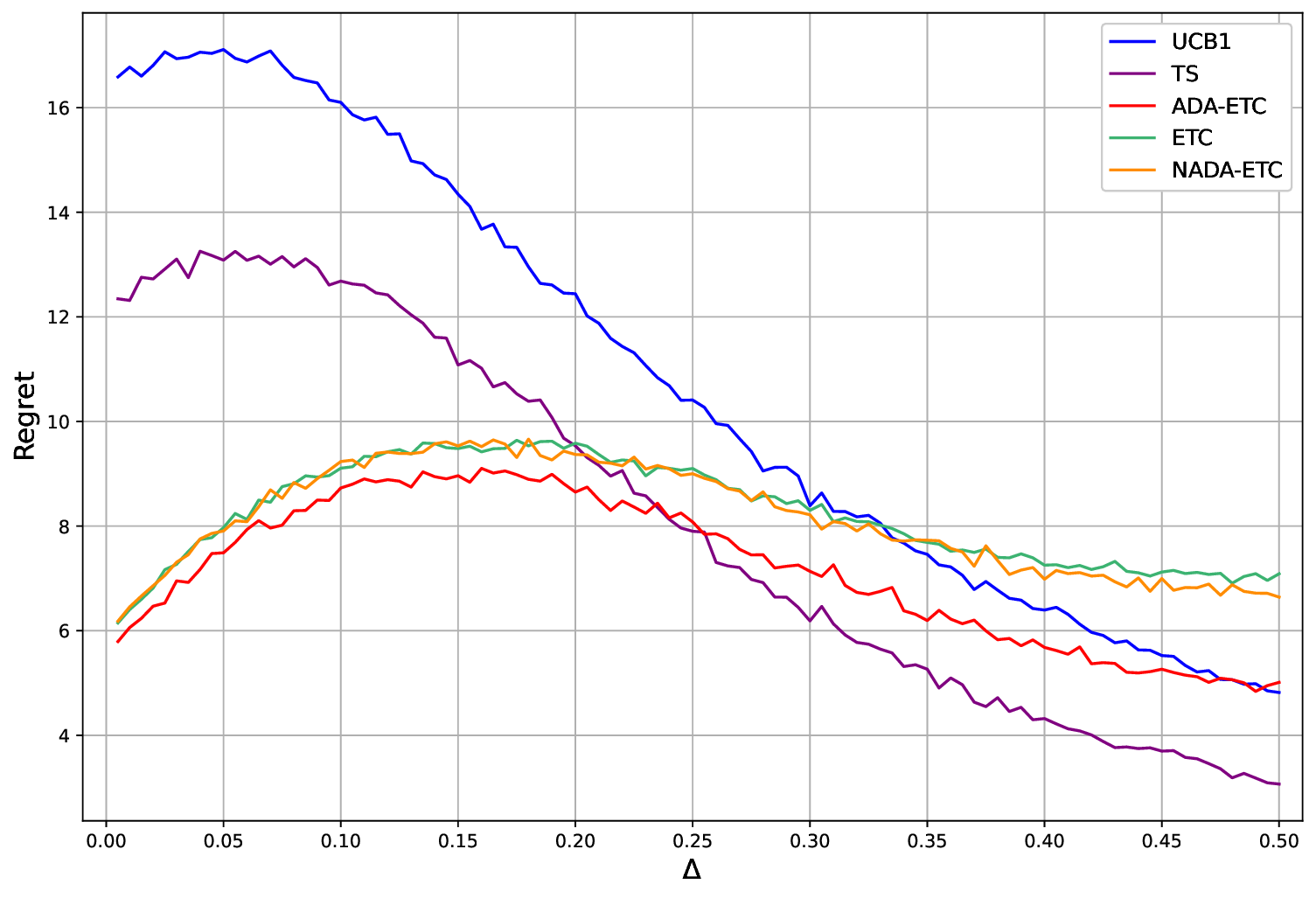}
		\caption{$K=2$, 1 opt.}
	\end{subfigure}%
	\hfill
	\begin{subfigure}{0.32\textwidth}
		\centering
		\includegraphics[width=\textwidth]{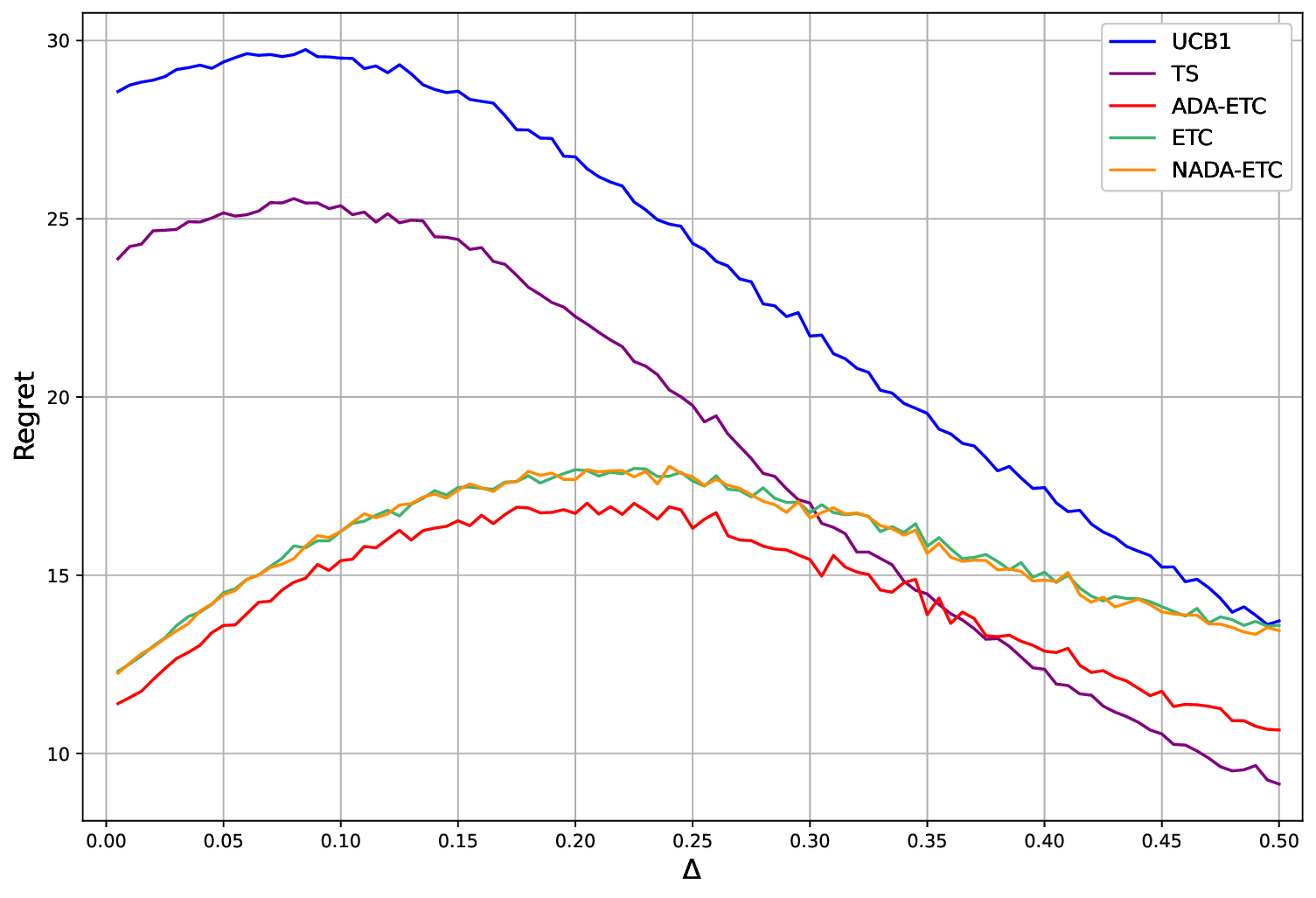}
		\caption{$K=4$, 1 opt.}
	\end{subfigure} 
	\hfill
	\begin{subfigure}{0.32\textwidth}
		\centering
		\includegraphics[width=\textwidth]{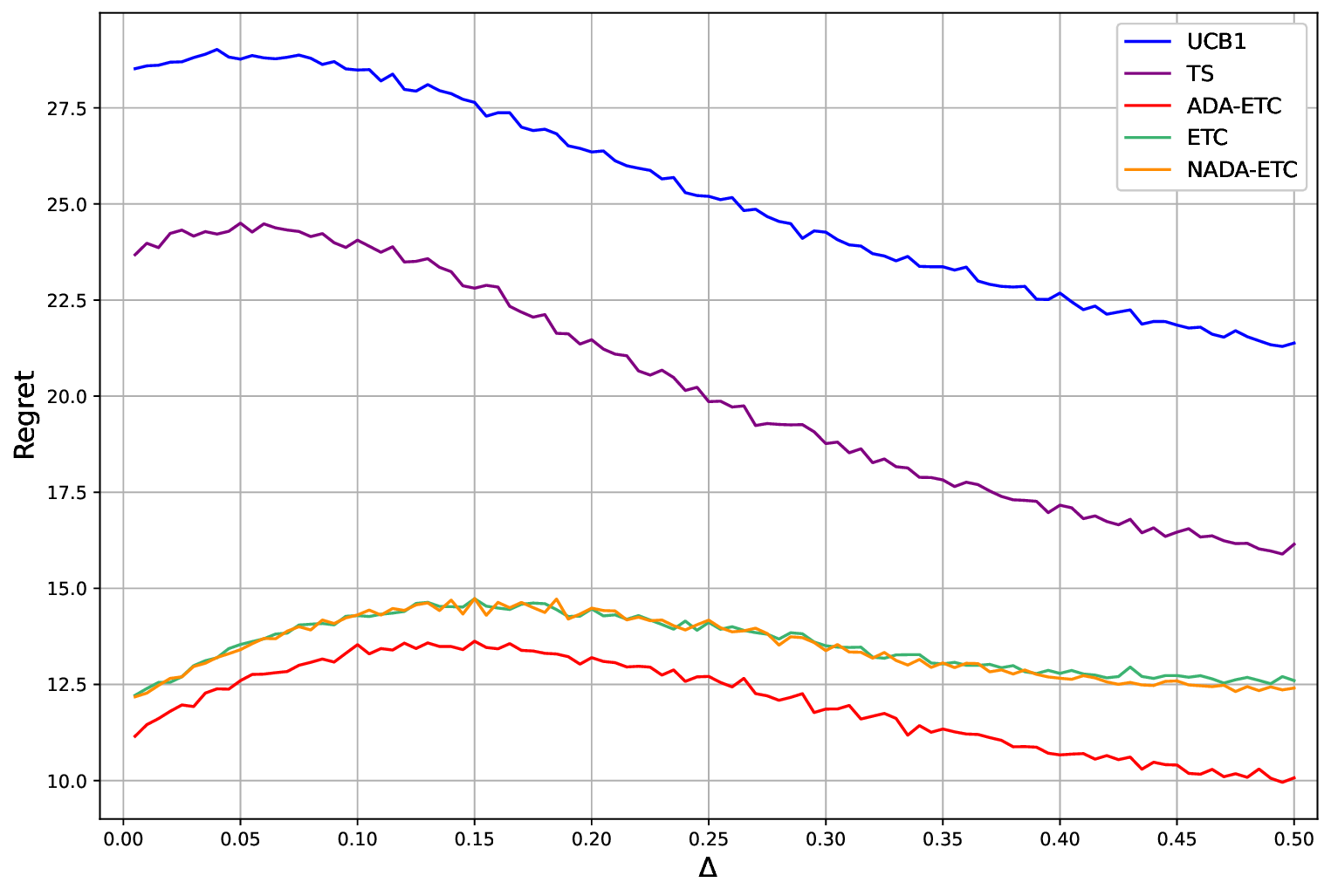}
		\caption{$K=4$, 2 opt.}
	\end{subfigure}%
	
	\caption{Performance comparison of ADA-ETC for varying values of $\Delta$.} \label{fig:experiments2}
\end{figure}

Next, we consider a set of bandit instances where the mean rewards of arms are set as $0.5$ or  $0.5-\Delta$, for some $\Delta>0$. Similar to the above experiments, we let $\bnu \sim \textup{Bernoulli}(\bmu)$. We vary $K \in \{2,4\}$ and fix $T=100$. The setting of Figure~\ref{fig:experiments2}(a) has two arms with $\bmu = \left( 0.5, 0.5-\Delta \right)$. Settings of Figures~\ref{fig:experiments2}(b) and (c) have four arms with $\bmu = \left( 0.5, 0.5-\Delta, 0.5-\Delta, 0.5-\Delta \right)$ and $\bmu = \left( 0.5, 0.5, 0.5-\Delta, 0.5-\Delta \right)$, respectively. Note the setting in Figure~\ref{fig:experiments2}(c) has two optimal arms. 
The average regret for each instance for different values of $\Delta$ is estimated over $1,000$ runs.

In all of the settings, we observe that the regret under UCB1 and TS follows a similar trend, with TS achieving a lower regret than UCB1. 
In Figures~\ref{fig:experiments2}(a) and (b), we note that TS starts to outperform ADA-ETC as $\Delta$ grows, suggesting good instance-dependent performance for large values of $\Delta$. ADA-ETC, however, consistently outperforms UCB1 across the spectrum.  As expected, the performances of TS and UCB1 deteriorate when two optimal arms are introduced in Figure~\ref{fig:experiments2}(c). 
In moving from Figure~\ref{fig:experiments2}(b) to~\ref{fig:experiments2}(c), we observe a lower peak in terms of average regret for ADA-ETC, ETC and NADA-ETC. This is a natural consequence of having two optimal arms: all else being equal, a second optimal arm reduces the probability of misidentifying an optimal arm.

%% file: gen_numerics.tex
\subsection{A study of regret for fixed instances: general $m$ } \label{sec:numerics_general}
We compare the performance of $m\textup{-ADA-ETC}$ with five algorithms described in Table~\ref{algorithms_m}. 
\begin{table}[h]
	\caption{Benchmark Algorithms}
	\label{algorithms_m}
	\centering\scriptsize{
		\begin{tabular}{lll}
			\toprule
			\vspace{-0.25cm}\\
			\multirow{2}{*}{$m\textup{-ADA-ETC}$}&$\ucb^i_n=$ $\bar{\mu}_n^i + \sqrt{\frac{4}{n}\log\left(\frac{T}{(K-m)n^{3/2}}\right)}\mathbbm{1}_{\left\{ n< \tau \right\}}$\\\vspace{0.15cm}&$\lcb^i_n=$ $\bar{\mu}_n^i - \bar{\mu}_n^i\mathbbm{1}_{\{n< \tau\}}$\\
			\multirow{2}{*}{$m\textup{-ETC}$}&$\ucb^i_n=$ $*$\\\vspace{0.15cm}&$\lcb^i_n=$ $*$\\
			\multirow{2}{*}{$m\textup{-UCB1}$}&$\ucb^i_n=$ $\bar{\mu}_n^i + \sqrt{\frac{1}{n}\log\left(T\right)}$\\\vspace{0.15cm}&$\lcb^i_n=$ $*$\\
			\multirow{2}{*}{$m\textup{-NADA-ETC}$}&$\ucb^i_n=$ $\bar{\mu}_n^i + \sqrt{\frac{1}{n}\log\left(T\right)}\mathbbm{1}_{\left\{ n< \tau \right\}}$\\\vspace{0.15cm}&$\lcb^i_n=$ $\bar{\mu}_n^i - \bar{\mu}_n^i\mathbbm{1}_{\{n< \tau\}}$\\
			\multirow{2}{*}{RADA-ETC}&$\ucb^i_n=$ $\bar{\mu}_n^i + \sqrt{\frac{4}{n}\log\left(\frac{T}{Kn^{3/2}}\right)}\mathbbm{1}_{\left\{ n< \tau \right\}}$\\\vspace{0.15cm}&$\lcb^i_n=$ $\bar{\mu}_n^i - \bar{\mu}_n^i\mathbbm{1}_{\{n< \tau\}}$\\
			\vspace{-0.25cm}\\
			\bottomrule
		\end{tabular}
	}
\end{table}
$m\textup{-ETC}$ pulls arms in a round-robin fashion and commits to the $m$ arms with the highest empirical means after each arm has been pulled $\tau$ times. As before,  $m\textup{-NADA-ETC}$ has the same algorithmic structure as $m\textup{-ADA-ETC}$: it explores based on upper confidence bounds and commits if the lower confidence bound of all arms rise above the upper confidence bounds of all other arms. It differs from $m\textup{-ADA-ETC}$ in how the upper confidence bounds are defined.

Additionally, we introduce a natural benchmark that utilizes ADA-ETC, which we call randomized ADA-ETC (RADA-ETC).
RADA-ETC randomly groups $K$ arms into $m$ subsets of size $K/m$ and runs ADA-ETC (until the end of Explore phase) for each subset with a budget of $T/m$ and returns one arm from each subset. Those arms are then pulled until the total number of remaining pulls allows. 

\begin{figure}[h]
	\centering
	\hspace{0.8cm}
	\begin{subfigure}{0.42\textwidth}
		\centering
		\includegraphics[width=\textwidth]{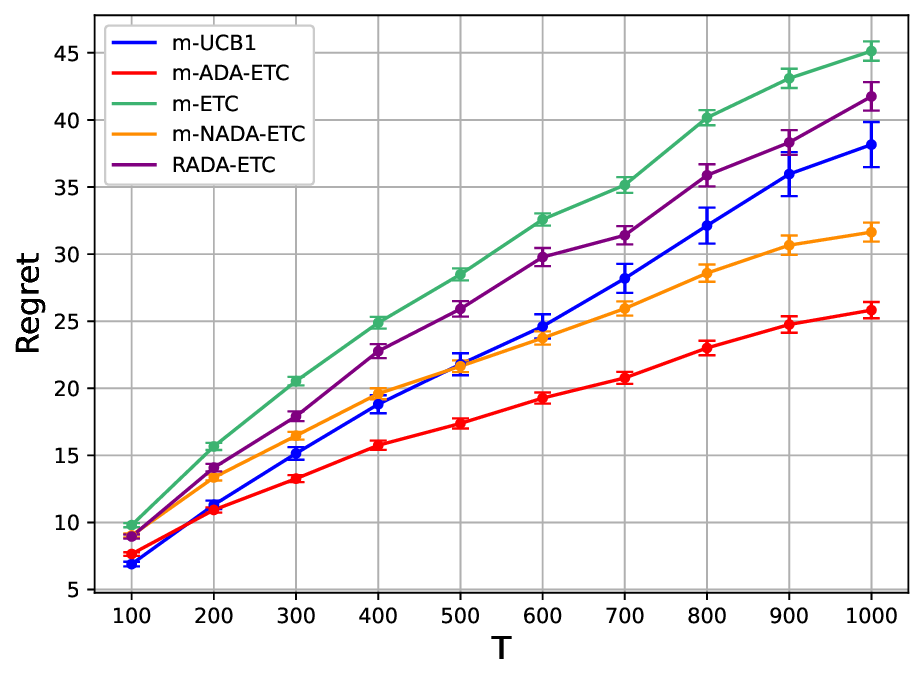}
		\caption{$m=2$, $K=4$, $\alpha = 0$}
	\end{subfigure}%
	\hfill
	\begin{subfigure}{0.42\textwidth}
		\centering
		\includegraphics[width=\textwidth]{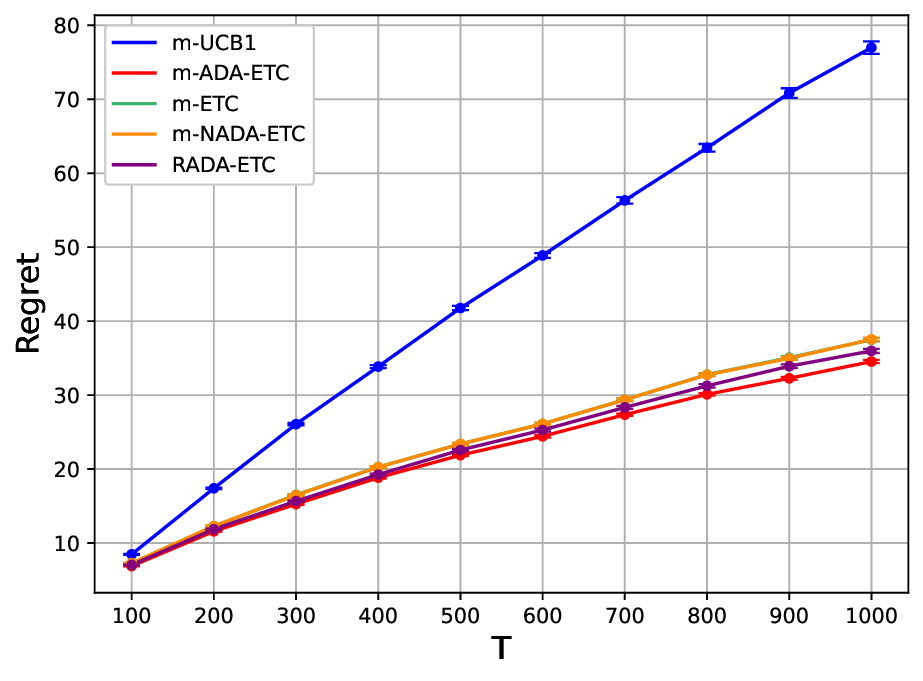}
		\caption{$m=2$, $K=4$, $\alpha = 0.4$}
	\end{subfigure} 
	\hspace{0.8cm}
	\vspace{0.1cm}
	
	\hspace{0.8cm}
	\begin{subfigure}{0.42\textwidth}
		\centering
		\includegraphics[width=\textwidth]{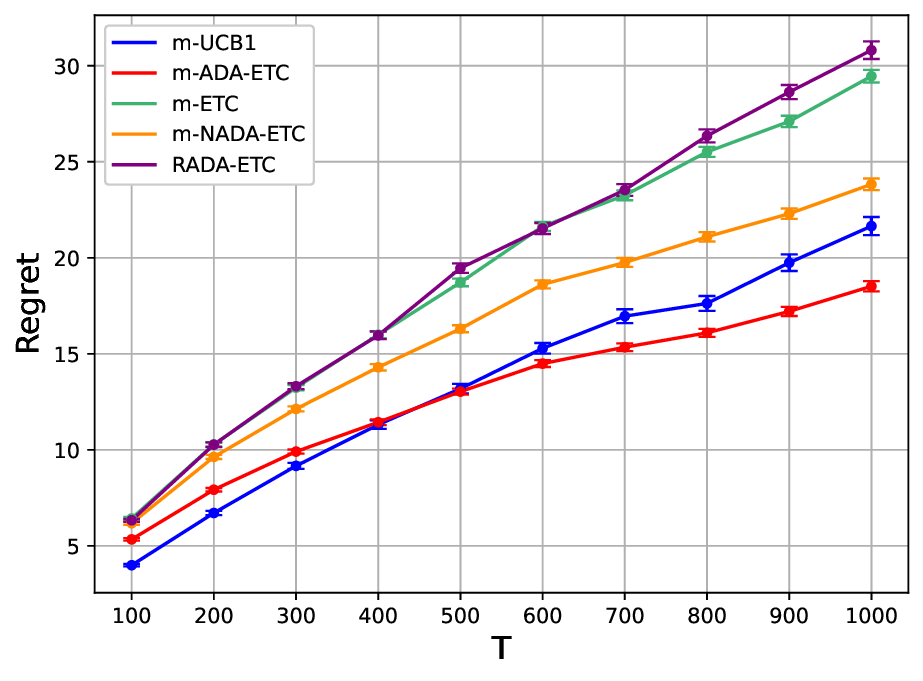}
		\caption{$m=4$, $K=8$, $\alpha = 0$}
	\end{subfigure}%
	\hfill
	\begin{subfigure}{0.42\textwidth}
		\centering
		\includegraphics[width=\textwidth]{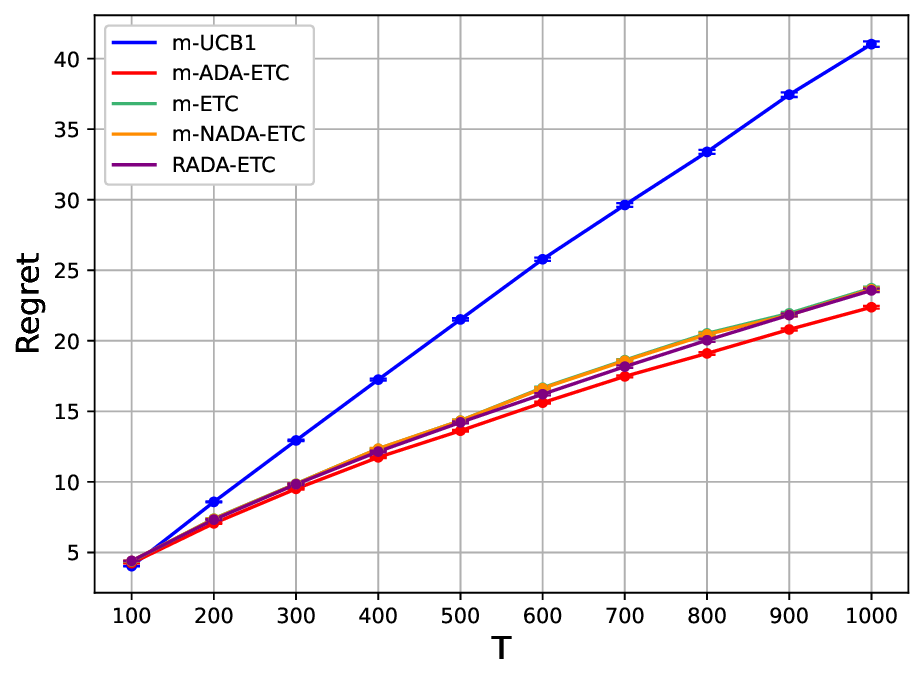}
		\caption{$m=4$, $K=8$, $\alpha = 0.4$}
	\end{subfigure} 
	\hspace{0.8cm}
	\vspace{0.1cm}

	\hspace{0.8cm}
	\begin{subfigure}{0.42\textwidth}
		\centering
		\includegraphics[width=\textwidth]{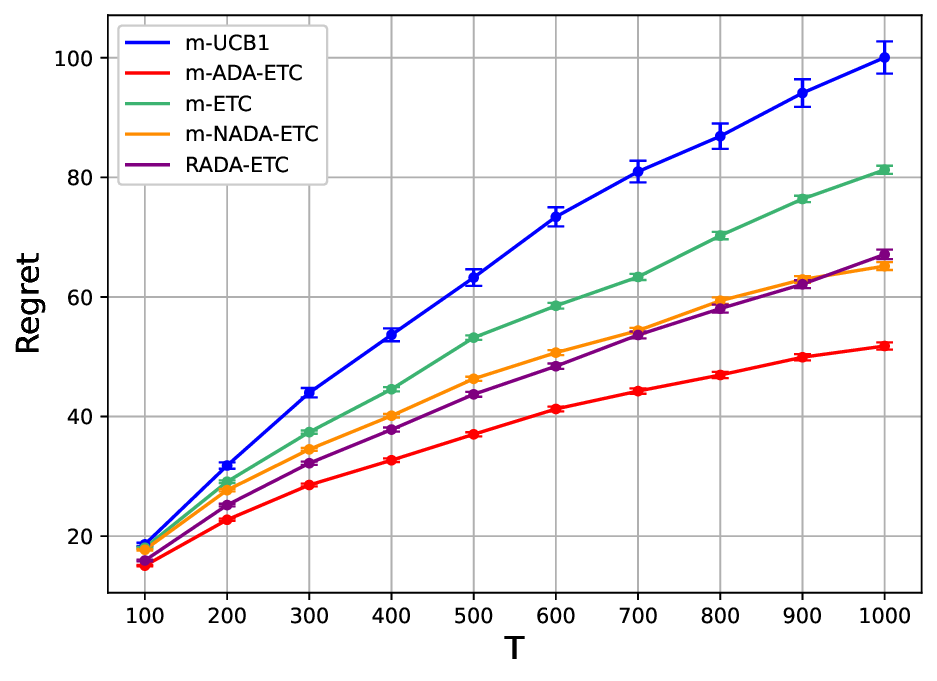}
		\caption{$m=2$, $K=8$, $\alpha = 0$}
	\end{subfigure}%
	\hfill
	\begin{subfigure}{0.42\textwidth}
		\centering
		\includegraphics[width=\textwidth]{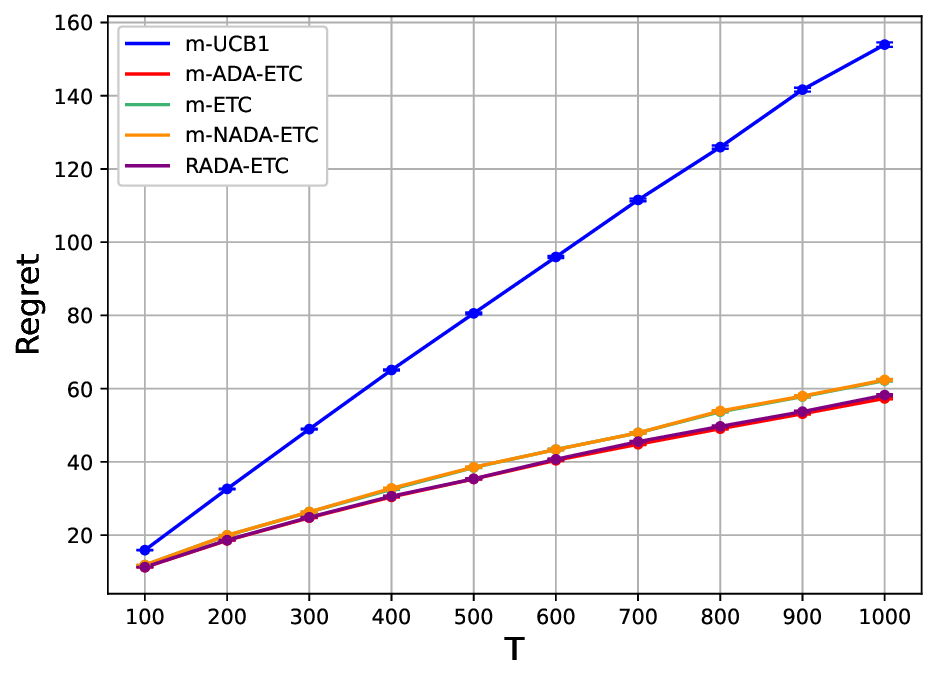}
		\caption{$m=2$, $K=8$, $\alpha = 0.4$}
	\end{subfigure} 
	\hspace{0.8cm}
	\caption{Performance comparison of $\boldsymbol{m}\textup{-ADA-ETC}$ for varying values of $T$.} \label{fig:experiments_m1}
\end{figure}

\noindent{\bf Instances.} We let $\nu_i \sim \textup{Bernoulli}(\mu_i)$, where $\mu_i$ is uniformly sampled from $[\alpha, 1-\alpha]$ for each arm in each instance. We sample two sets of instances, each of size $200$, with $\alpha \in\{0, 0.4\}$. The regret for an algorithm for each instance is averaged over $50$ runs to estimate the expected regret. 
We vary $m \in \{2, 4 \}$, $K \in \{4, 8\}$ and $T \in \{100, \dots, 1000\}$. The average regret over the $200$ instances under different algorithms and settings is presented in Figure~\ref{fig:experiments_m1}.

\noindent{\bf Discussion.} 
Similar to the $m = 1$ case, $m\textup{-ETC}$ and $m\textup{-NADA-ETC}$ perform as well as $m\textup{-ADA-ETC}$ when we either have (a) $\alpha = 0.4$, that is, the arms are expected to be close to each other and so that adaptivity in exploring is not as beneficial, or (b) $T/(K-m)$ is small, so that $\tau$ is small and is likely to be exhausted under $m\textup{-ADA-ETC}$. A similar observation holds for our new benchmark RADA-ETC. 

It is interesting to note that RADA-ETC's performance is comparable to ETC, even outperforming ETC in certain cases. This suggests the gains from adaptivity can more than compensate for the performance loss due to random grouping. Its performance, however, is mostly worse than  $m\textup{-NADA-ETC}$ across the spectrum. A mild exception is Figure~\ref{fig:experiments_m1}(e), where, since the size of the random subsets $(K/m)$ is relatively large compared to $m$ ($= 2$), the probability of the top $2$ arms being in the same random subset is significantly lower. Thus the performance loss due to random grouping is expected to be small, and the refined adaptivity may help improve performance over $m\textup{-NADA-ETC}$.


It is also interesting to note that $m\textup{-UCB1}$ outperforms $m\textup{-ADA-ETC}$ in Figures~\ref{fig:experiments_m1}(a) and (c) for smaller $T$ values when $\alpha =0$. 
Looking at the evolution of the performance of $m\textup{-UCB1}$ through Figures~\ref{fig:experiments_m1}(a), (c), and (e), we observe that it seems to flourish in well-separated instances when $m$ is large relative to $K$. 


And finally, as in the $m=1$ case, $m\textup{-ADA-ETC}$ consistently outperforms $m\textup{-NADA-ETC}$, thus illustrating the gains from the tuning of the upper confidence bounds under $m\textup{-ADA-ETC}$.

%% file: numerics_amazon.tex
\subsection{Application to product grooming on electronic commerce platforms} \label{sec:numerics_amazon}
In this section, we numerically evaluate ADA-ETC's performance for the problem of grooming a single product amongst competing ones on Amazon.com, using rating distributions of four product types. 
These product types are: (1) automobile dash cams that are attached to the rearview mirror, (2) mid-size snow shovels (non-electric), (3) cordless leaf blowers (of a similar power rating), and (4) humidifiers for single-room use.

\begin{figure}[h]
	\centering
	\begin{subfigure}{0.48\textwidth}
		\centering
		\includegraphics[width=\textwidth]{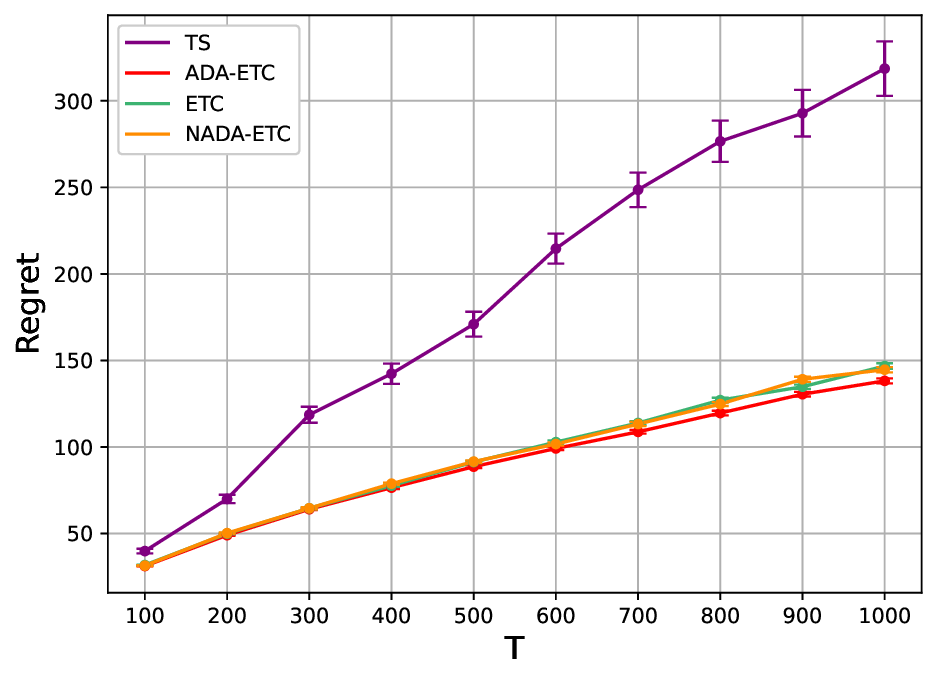}
		\caption{Dash cams}
	\end{subfigure}%
	\hfill
	\begin{subfigure}{0.48\textwidth}
		\centering
		\includegraphics[width=\textwidth]{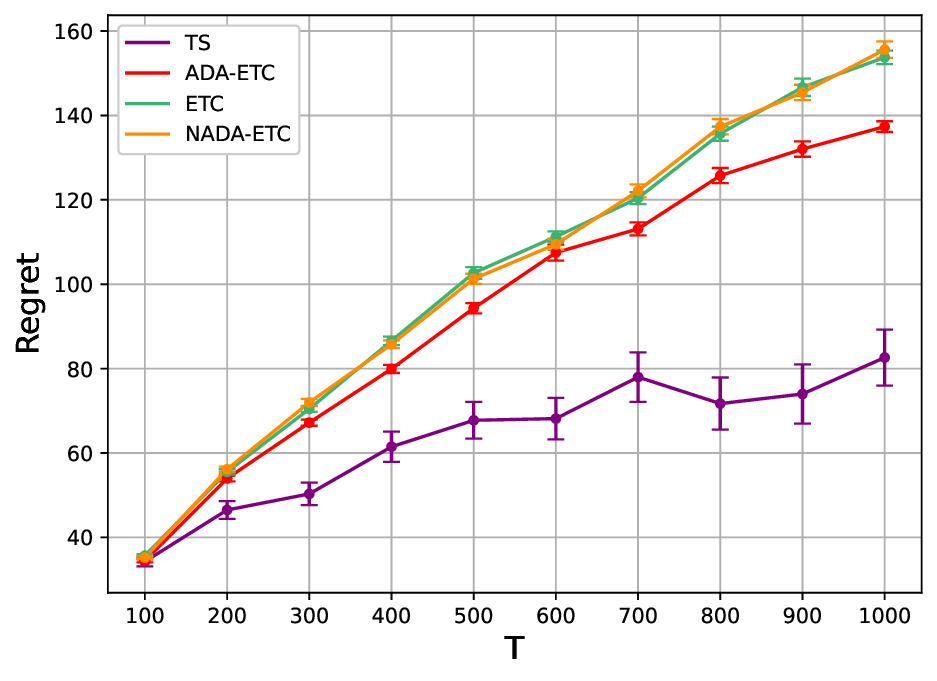}
		\caption{Snow shovels}
	\end{subfigure} 
	\vspace{0.1cm}
	
	\begin{subfigure}{0.48\textwidth}
		\centering
		\includegraphics[width=\textwidth]{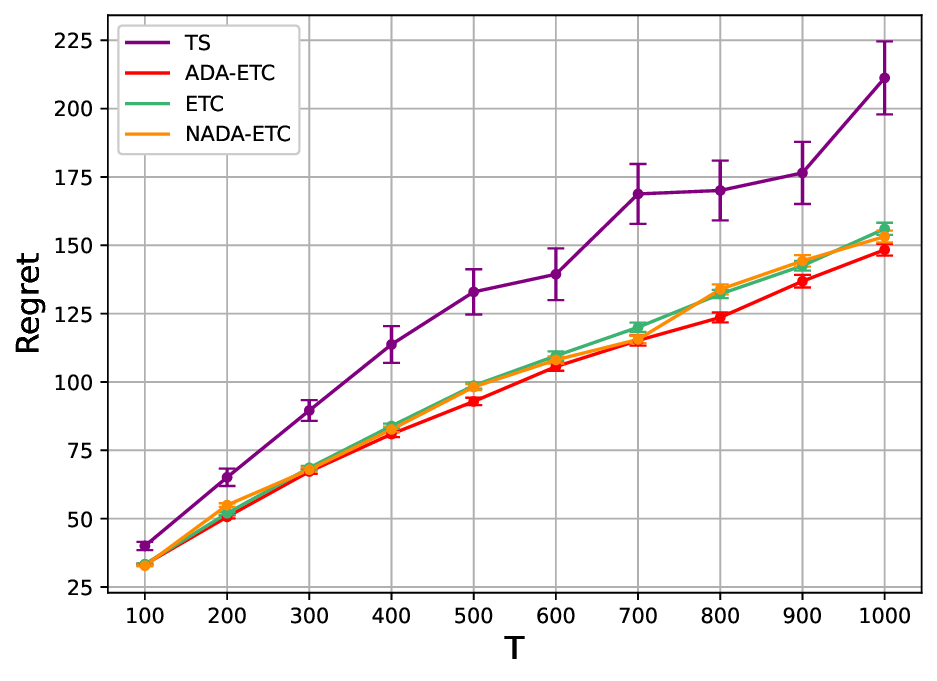}
		\caption{Leaf blowers}
	\end{subfigure}%
	\hfill
	\begin{subfigure}{0.48\textwidth}
		\centering
		\includegraphics[width=\textwidth]{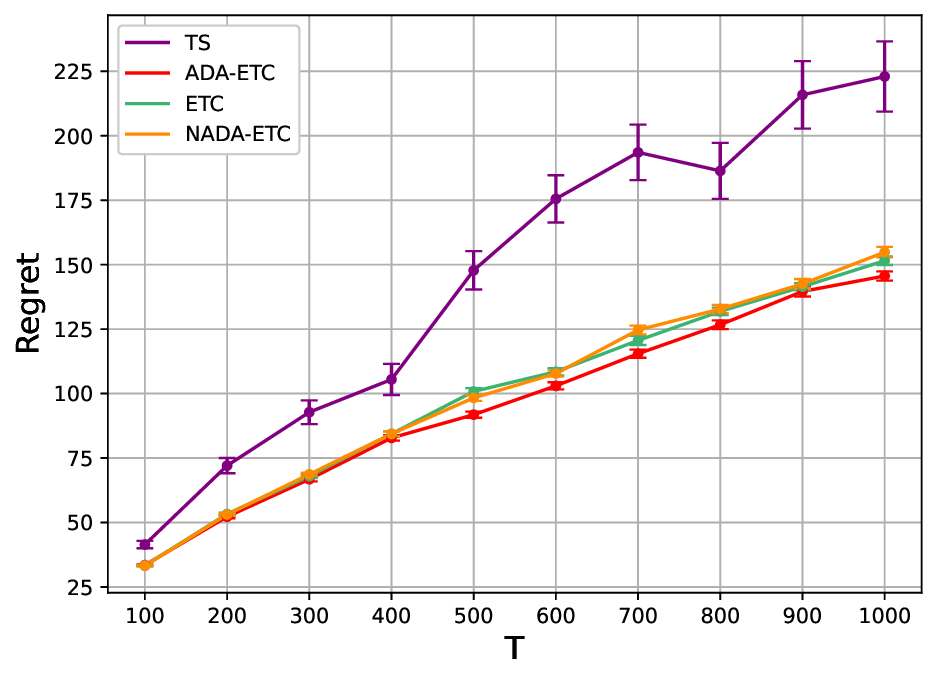}
		\caption{Humidifiers}
	\end{subfigure} 
	
	\caption{Performance comparison of ADA-ETC for varying values of $T$.} \label{fig:experiments_amazon}
\end{figure}

Different from the earlier experiments, we assume that the rewards (ratings that the products receive from customers following a purchase) are in $\{0.2, 0.4, 0.6, 0.8, 1 \}$, where the probability of each rating is set directly using the rating distribution of the corresponding product obtained from Amazon.com (a scale of 1 to 5 is mapped to the aforementioned discrete range). The regret for an algorithm for each product is averaged over $100$ runs to estimate the expected regret.
Fixing $K = 6$, i.e., considering $6$ competing products for each product type, we vary $T \in \{100, \dots, 1000\}$. The average regret under different algorithms and settings is presented in Figure~\ref{fig:experiments_amazon}. Since UCB1 incurs very high regret compared to the other benchmarks in all instances, we exclude its performance to differentiate the performance of other policies better. We include a reference figure with UCB1 in Appendix~\ref{apxsec:moreFigs}, Figure~\ref{fig:experiments_amazon_w_UCB1}.

In all settings, we observe a trend similar to our observations from Section~\ref{sec:m1exp}. We observe that ADA-ETC consistently outperforms ETC and NADA-ETC. TS performs poorly in all settings, except for the case of Snow shovels, where it significantly outperforms all other algorithms. This is because, in this case, there is a unique product with the highest true quality that is considerably better than other competing products (rating distribution of each product can be found in Tables~\ref{table:1}-\ref{table:4} in Appendix~\ref{apxsec:moreFigs}). As we observed in Section~\ref{sec:m1exp}, TS performs well in such settings. In this case, the performance gain of ADA-ETC over ETC and NADA-ETC is also more significant compared to other settings where the products have more-or-less similar ratings.



%% file: numerics_market.tex
\subsection{Application to supply grooming in online service platforms} \label{sec:numerics_market}
In this section, we describe how we can use the algorithms developed for the static multi-armed bandit problem for the problem of onboarding novice workers or service providers in online platforms. The key distinction in these platforms relative to the static model is that the workers arrive and depart over time, and there is a fixed capacity of training jobs arriving per unit of time. Despite this distinction, we argue that our algorithms can be directly implemented.

Consider a dynamic market model where workers and jobs (that can be used for onboarding) arrive according to some stochastic arrival process. The new workers undergo an onboarding phase over a certain duration, during which they are successively matched to jobs. Suppose that the demand for well-rated workers to serve the demand for high-value jobs is such that $m$ out of every $K$ incoming workers, on average, need to be groomed to have good ratings at the end of the onboarding period.\footnote{We suppose that $m$ is the smallest integer such that grooming $m$ out of some integer $K$ workers is a practically good enough approximation to serve the demand for jobs. For example, suppose that a 5 /17 fraction of the incoming workers must be groomed (to be exact). This fraction amounts to 1 out of every 3.4, which can be approximated to 1 in every 3. This may be too excessive an approximation. However, 5/17 also equals 3 out of every 10.2, which can be approximated to 3 out of 10, which may be an acceptable approximation. In this case, we assume that $m=3$ and $K=10$.} To align with our earlier notation, suppose that $T$ jobs arrive on average in the duration over which $K$ successive workers arrive, where suppose for now that $T$ is an integer multiple of $m$ (as described below, we will lower the consumption of jobs to some $T'<T$, which can be chosen to satisfy this assumption). The main idea behind how our algorithms can be used in the dynamic setting with arrivals and departures is that of ``cohorting." 



{\bf Cohorting.} We can consider $K$ successively arriving workers as a cohort, out of which $m$ must be trained. Each cohort gets matched to $m$ jobs in a batched assignment, amounting to batched pulls of $m$ distinct arms. Suppose that the training duration lasts for $T/m$ assignments. Then the number of jobs demanded by each cohort for their onboarding is $T/m\times m$, which is precisely the capacity of jobs available per cohort.\footnote{We note that batching of $m$ pulls per time period is not necessary, and we make this assumption to align with our static bandit model and objective for the general $m$ case. In particular, we could assume that each cohort consumes a single job in each assignment, with an onboarding duration of $T$ assignments. However, without the batching assumption, optimizing the $m>1$ objective would result in a single ``well-groomed" worker in each cohort, which is an undesirable outcome, as we have discussed earlier.} Effectively, this capacity $T$ determines the duration of the training period: the larger the $T$, the better rated the outflow of workers will be from training. The job assignments for each cohort can be made using our $m$-ADA-ETC algorithm. 

Extending this basic cohorting approach, we can consider creating larger cohorts. For instance, for any integer $H$, $K_H = KH$ consecutive workers can be considered a cohort, out of which $m_H= mH$ workers must be trained to satisfy the average demand. For each cohort, we make batched assignments of $mH$ jobs in every assignment. And the onboarding duration for each cohort is $TH/mH=T/m$ batches of assignments. Thus the total number of job assignments needed per cohort is $T/m\times mH=TH \triangleq T_H$, which is precisely the capacity of incoming jobs available for cohorts of size $KH$. Note that the job duration is immaterial to the above discussion concerning the balancing of the mean supply and demand.

Naturally, larger cohorts are better since training the best $m$ workers out of every $H$ successive cohorts of size $K$ is worse than training the best $mH$ out of a single larger cohort of $KH$ successively arriving workers. However, creating larger cohorts delay the assignments of jobs by a factor of $H$ on average. This results in an interesting tradeoff reminiscent of the market thickness vs. delay tradeoff studied extensively in the dynamic matching literature \citep{akbarpour2020thickness, baccara2020optimal, loertscher2022optimal}. We study this tradeoff in our market simulation, where we see diminishing marginal gains from increasing $H$.

\begin{remark} {\it One concern is that, since the arrival processes are stochastic and the job durations are random, if the demand for onboarding jobs exactly matches the supply in such a stochastic system, then a growing backlog of cohorts may result (the issue is akin to ``null recurrence" in Markovian systems; see \cite{suncongestion} for a discussion), resulting in a reduced supply than required. This concern can be addressed by reducing the number of batched assignments for each cohort from $TH/mH$ to $T'_H/mH$ for some $T'_H<TH$ (in our simulation, we define $T'_H = TH-\Theta(\sqrt{TH})$ as justified in \cite{hsu2021integrated} and \cite{suncongestion}). Ensuring adequate supply thus comes at the cost of (slightly) reduced outgoing quality levels.}
\end{remark}




\subsection{Simulation}

\begin{figure}[h]
	\centering	
	\hspace{0.2cm}
	\begin{subfigure}{0.47\textwidth}
		\centering
		\includegraphics[width=\textwidth]{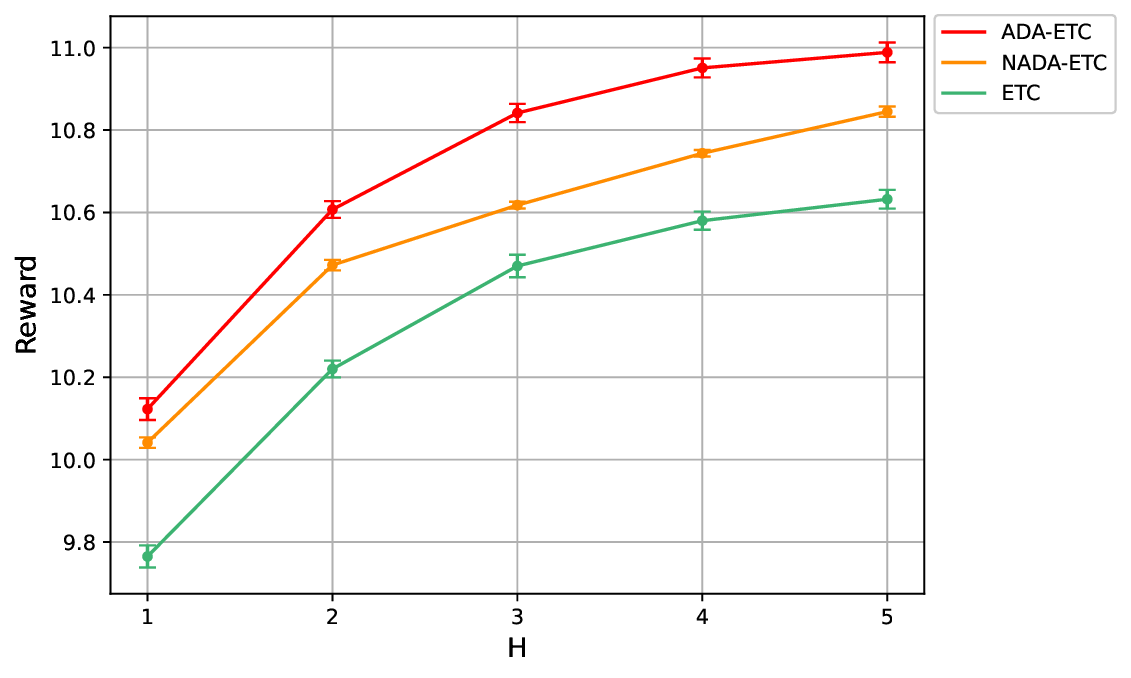}
		\caption{$m_H=H$, $K_H=2H$, $T_H=20H$, $10000$ periods}
\end{subfigure}
\hfill
\begin{subfigure}{0.47\textwidth}
	\centering
	\includegraphics[width=\textwidth]{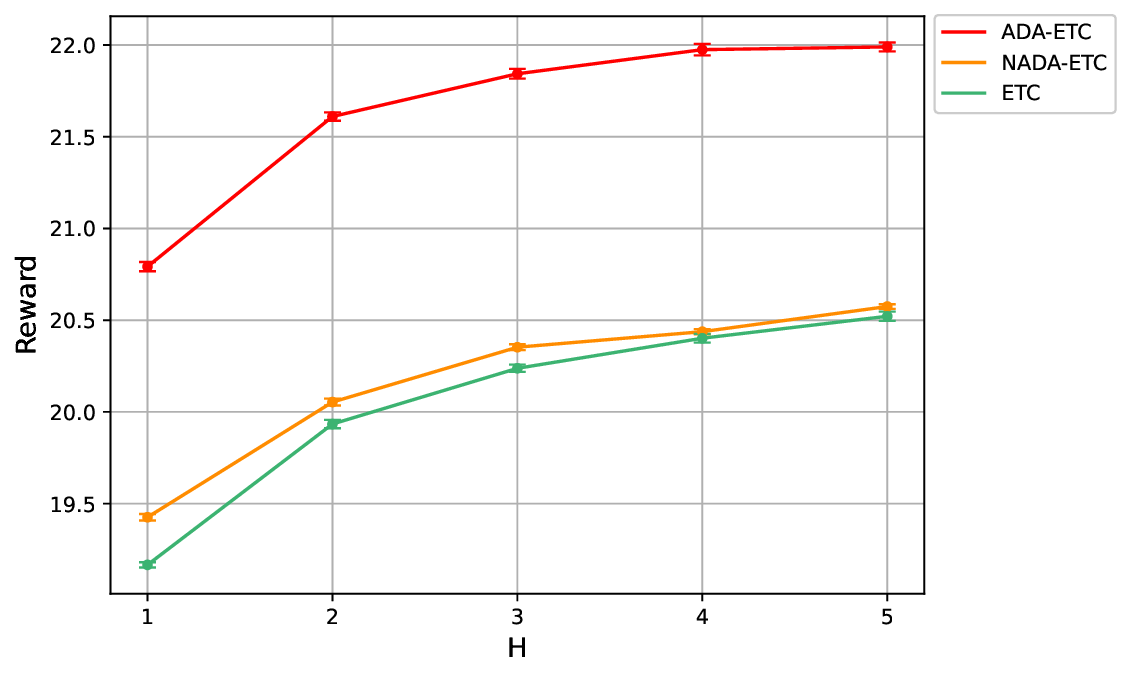}
	\caption{$m_H=H$, $K_H=4H$, $T_H=40H$, $20000$ periods}
\end{subfigure}	
\hspace{0.2cm}
\caption{Average skill level of trained workers right at the end of the set periods for various $m$ values}
\label{fig:market}
\end{figure}

\begin{figure}[h]
	\centering	
	\hspace{0.2cm}
	\begin{subfigure}{0.47\textwidth}
		\centering
		\includegraphics[width=\textwidth]{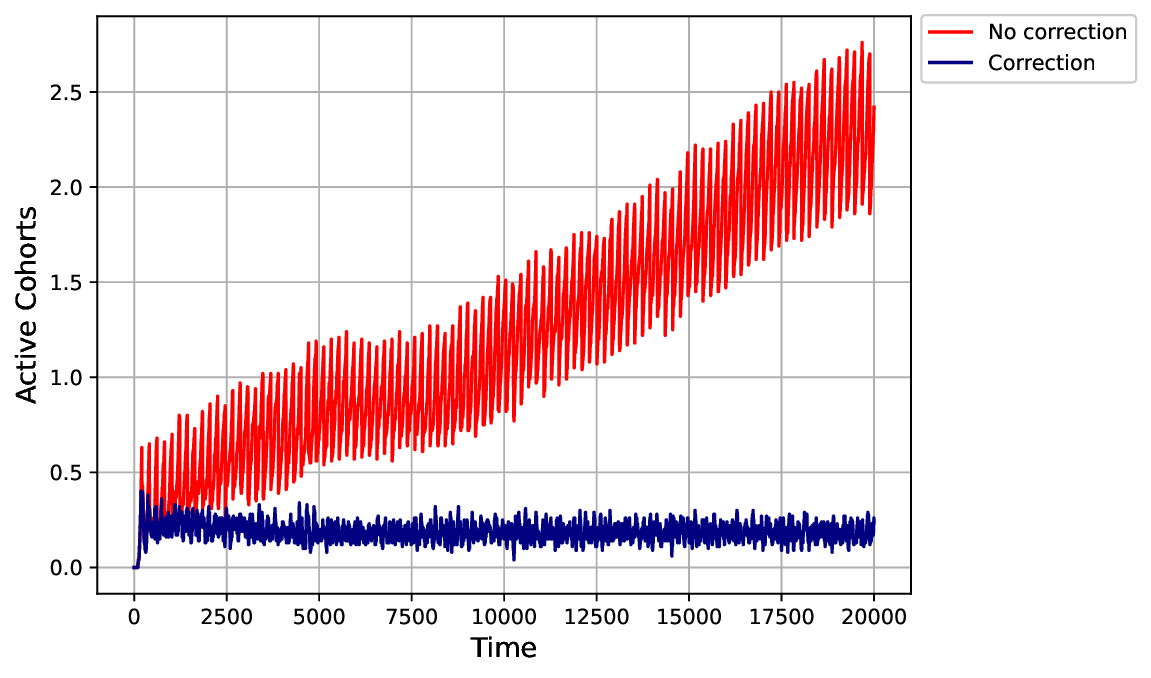}
		\caption{Active cohorts for $m_H=5$, $K_H=20$, $T_H=200$}
	\end{subfigure}
	\hfill
	\begin{subfigure}{0.47\textwidth}
		\centering
		\includegraphics[width=\textwidth]{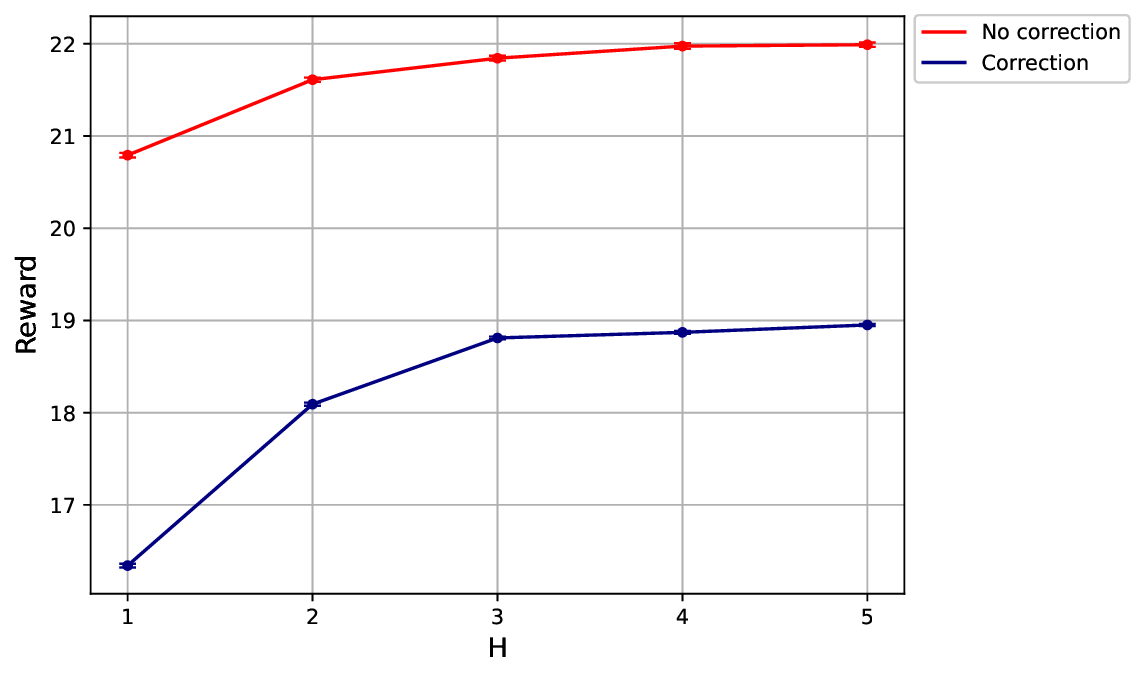}
		\caption{$m_H= H$, $K_H=4H$, $T_H=40H$}
	\end{subfigure}	
\hspace{0.2cm}
\caption{Effect of decreasing the number of assignments for each cohort for ADA-ETC. Snapshots right after $20,000$ periods.}\label{fig:market2} 
\end{figure}

We present the simulated results in Figures~\ref{fig:market} and~\ref{fig:market2}. In Figure~\ref{fig:market}(a), we set the {\it base} case at $m=1$, $K=2$, $T=20$ and vary $H$ from $1$ to $5$. So we get $m_H=H$, $K_H=2H$, $T_H=20H$ for $H \in \{1,\dots,5\}$. Similarly, in Figure~\ref{fig:market}(b) we set $m_H=H$, $K_H=4H$, $T_H=40H$ for $H \in \{1,\dots,5\}$. Essentially, compared to the former case where one in two workers must be groomed, in the latter setting we must groom one in every four workers. Note that the ratio of $K_H/T_H$ remains the same across the two settings since we consider the same stochastic process of worker and job arrivals.

The discrete time simulation works as follows. At each time period, exactly one job arrives and remains in the system until it is assigned to a cohort, whereas a new worker arrives with probability $0.1$ and idly waits until being included in a cohort. The rating received by a worker for a job is modeled as a Bernoulli random variable (e.g., the worker receives a ``like" or a ``dislike"). The true but unknown mean rating for an arriving worker is uniformly sampled from $[0,1]$. After enough number of idle workers arrive to form the next cohort, a new cohort is formed, i.e., activated, and is ready to receive (batched) job assignments. At each time period, depending on the available number of jobs in the system, each cohort receives the set number of (batched) job assignments in the order of activation, i.e., the priority is given to the oldest cohort that can receive job assignments, until either there are no more enough jobs to assign to a cohort or there are no active cohorts that can receive job assignments. Each job takes a single period to perform. After an active cohort receives all of its allotted jobs ($T_H$), it leaves the system and its reward (average total ratings of the top $m_H$ workers) is noted.

In Figure~\ref{fig:market}, we report the average total ratings of all top workers (arms) who leave the system prior to time $10,000$ for Figure~\ref{fig:market}(a), and time $20,000$ for Figure~\ref{fig:market}(b). In both of these settings, a similar number of workers are trained across different $m_H$ values for varying $H$ (with small variations due to randomness of worker arrivals and different cohort sizes). In Figure~\ref{fig:market}(b), the onboarding periods are doubled (since fewer workers need to be groomed) and hence a higher reward per worker is obtained.

The two benchmark algorithms included in Figure~\ref{fig:market}, NADA-ETC and ETC (and their general $m$ counterparts, $m\textup{-NADA-ETC}$ and $m\textup{-ETC}$) are the same algorithms described in the previous sections. As before, our proposed algorithm, ADA-ETC (and its general $m$ counterpart, $m\textup{-ADA-ETC}$), outperforms the benchmarks. Another important observation here is that, as $H$ grows, we see diminishing marginal gains on the average ultimate skill level of all workers.

In Figure~\ref{fig:market2}, we use the numerical setting of Figure~\ref{fig:market}(b) and experiment with using $T'_H = TH-\sqrt{TH}$ instead of $TH$ as the number of jobs required throughout the lifetime of one cohort. For numerical convenience, we use the largest $T''_H$ smaller than $T'_H$ and divisible by the corresponding $m_H$ value. We call this the ``correction" procedure. ``No correction" refers to the original setting described at the beginning of this section.

In Figure~\ref{fig:market2}(a), we report the average number of active cohorts across $100$ simulations with and without this ``correction" procedure. We observe that in the original case, even though the supply matches the demand for jobs, the average number of active cohorts increases with time due to randomness in the system. With the correction, we observe that the system has consistent access to enough jobs to assign to all active cohorts and has these cohorts leave the system without delay.
Although, this decrease in the consumption of jobs per cohort comes at the cost of reduced average total ratings of the groomed workers as Figure~\ref{fig:market2}(b) shows. 

%% file: conclusion.tex
\section{Discussion and Conclusion}\label{sec:conclusion}
In this paper, we proposed and analyzed new maximal objectives under the multi-armed bandit framework. While these objectives are primarily motivated in the context of supply grooming in online platforms, they have other applications. 

{\bf Broader applications.} More broadly, our model and objectives capture exploration vs. exploitation tradeoffs in situations where the rewards represent incremental progress made by an ``arm" toward a goal (improving skills, obtaining more ratings, etc.). In these scenarios, it is natural that the global objective is some function of the terminal levels of progress across arms. And in many of these scenarios, only the highest progress levels matter. For example, consider a supercomputing facility allocating computing time across multiple research teams trying to solve a common hard problem (say in computational biology or astrophysics). In such cases, the progress made by teams is often non-transferable, and it is the maximal terminal level of progress across teams that ultimately matters. Such problems of resource allocation (e.g., monetary budget allocation) across competing teams also arise in large organizations. 

In the same spirit, our objectives are also relevant to developing advanced talent within a region for participation in external competitions like Science Olympiads, the Olympic games, etc., with limited training resources. Only the terminal skill levels of those finally chosen to represent the region in these settings matter. The resources spent on others, despite resulting in skill advancement, are wasteful (purely from the perspective of optimizing for success in the external competition). This feature is not captured by the sum objective, while it is better captured by our maximal objectives. 

While our algorithms may not be directly applicable to the above settings given the stylized nature of our assumptions (i.i.d. rewards in particular), our analysis provides insights into the change in the nature of the exploration vs. exploitation tradeoffs and the algorithmic desiderata resulting from these new objectives.

{\bf Application to the sum objective.} It is interesting to note that the max regret bounds reported for our algorithms also hold for the traditional sum regret. In particular, ADA-ETC achieves the order-optimal $\textup{O}(\log T)$ instance-dependent regret bound and a $\tilde{\textup{O}}(K^{1/3}T^{2/3})$ instance-independent regret bound for the sum objective. Thus, while sum-optimized algorithms are not a viable alternative for the max objective given their almost-linear worst-case regret performance implied by Theorem~\ref{thm:sumbad}, max-optimal algorithms such as ADA-ETC are a reasonable alternative for optimizing the sum-objective. For instance, in certain settings where the sum objective is appropriate, it may also be desirable that the bandit algorithm invests $T-\textup{o}(T)$ pulls in a single arm in the long run, even in the worst case. For example, a major application of multi-armed bandit theory is to the design of clinical trials of drugs in healthcare settings \citep{villar2015multi}. In many cases, the primary objective of such a trial is to minimize the cost of the trial (making the sum objective appropriate). However, it may also be desirable to achieve high accuracy in the performance estimate of the drug that is ultimately chosen. Thus, if two drugs have very similar performances, it is wasteful from the accuracy perspective to expend a $\Theta(T)$ treatment budget on both these arms, which is what a policy like UCB1 could do (Proposition~\ref{prop:ucbbad}). While ETC can address this concern, ADA-ETC is a strictly better alternative given its order-optimal instance-dependent performance, unlike ETC.

{\bf Future directions.} Several research directions result from our work. In general, the many technical pursuits that have been fruitfully explored for the sum objective open up similar lines of inquiry for the maximal objectives. For instance, significant effort has gone into designing algorithms that achieve exact asymptotic optimality (not just upto a constant factor) and worst-case optimality (not just up to $\log$ factors) for the sum objective \citep{audibertminimax2009, garivier2011kl, lattimore2018refining}. Such advancements remain open for the max objective.  As another example, Bayesian analysis of the classical multi-armed bandit problem under the sum objective results in the elegant theory of Gittins indices \citep{gittins2011multi}. It would be interesting to similarly analyze the problem under the maximal objective. We detail some more practically meaningful extensions below.



\begin{enumerate}
\item {\bf Other extensions of the $m=1$ case.} In the case of $m>1$, we impose the constraint that $m$ arms must be chosen simultaneously to avoid the practically undesirable solution of pulling only the best arm. However, other meaningful problem formulations address this issue. One natural objective is to maximize the expected minimum terminal reward across the top $m$ terminal rewards. This objective avoids the perverse optimal benchmark of always pulling the optimal arm and may be meaningful in many scenarios where a platform seeks to improve its service guarantees. This formulation, however, excessively focuses on the weakest of the top workers, which may result in lower average terminal reward across the top workers compared to our general $m$ approach, which devotes equal attention to all these workers in the long run. Nevertheless, this generalization is an interesting direction for future work.

\item {\bf Adaptation of Thompson sampling for the max objective.} In our numerical experiments, we observed that TS performs quite well for the max objective in the case of well-separated instances. However, its performance deteriorates as the arms get closer. This suggests that introducing an appropriate stopping criterion to TS could result in a robust policy that performs well across the spectrum of separation. Investigating the design of such a criterion, along with providing performance guarantees, is an interesting direction for future work.  

\item {\bf General global objectives.} Finally, our paper presents the possibility of studying a variety of new objectives under existing online learning models motivated by training or grooming applications, where both the sum or the maximal objectives may be inappropriate. For example, many students who get trained for the mathematics olympiad and who are eventually not chosen to represent the country go on to become impactful researchers owing to their early training. Thus the excessive focus on grooming the best students may not be appropriate and caution is necessary to apply our approaches in such societal settings. In such settings, other global objectives that achieve a more balanced allocation, e.g., maximizing the $L^p$ norm of the terminal rewards across arms, for some $p\in (0,1)$, may be more appropriate. Characterizing the optimal regret frontier (instance-dependent or worst-case) across this spectrum as $p$ varies is an exciting open question. 


\end{enumerate}
We are optimistic that the analysis in this paper will be useful in tackling many of these pursuits.



%% file: appendix.tex
\begin{APPENDICES} 

\newpage	
\section{Establishing regret benchmarks.}
	\subsection{Proof of Proposition \ref{prop:main}}\label{apx:prop1}
	For any policy $\pi$, we have that
	\begin{align}
		\mathcal{R}_T(\pi,\bnu) &= \textup{E}\big(\max_{i\in[K]} \overline{U}^i_{T}\big)\nonumber\\
		&= \textup{E}\big(\max_{i\in[K]}\big(\sum_{t=1}^{T}U^i_{n^i_{t-1}+1}\mathbbm{1}_{\{I_t =i\}}\big)\big)\nonumber\\
		&\overset{(a)}{\leq}  \textup{E}\left(\sum_{t=1}^{T}\max_{i\in[K]}\big(U^i_{n^i_{t-1}+1}\mathbbm{1}_{\{I_t =i\}}\big)\right)\nonumber\\
		&=\sum_{t=1}^{T}\textup{E}\left(\max_{i\in[K]}\big(U^i_{n^i_{t-1}+1}\mathbbm{1}_{\{I_t =i\}}\big)\right)\nonumber\\
		&\overset{(b)}{=}  \sum_{t=1}^{T}\textup{E}\left(U^{I_t}_{n^{I_t}_{t-1}+1}\max_{i\in[K]}\big(\mathbbm{1}_{\{I_t =i\}}\big)\right)\nonumber\\
		&=\sum_{t=1}^{T}\textup{E}\left(U^{I_t}_{n^{I_t}_{t-1}+1}\right)\nonumber\\
		&=\sum_{t=1}^{T}\textup{E}\left(\textup{E}\left(U^{I_t}_{n^{I_t}_{t-1}+1}\mid \mathcal{H}_t\right)\right)\nonumber\\
		&\overset{(c)}{=} \sum_{t=1}^{T}\textup{E}\left(\mu_{I_t}\right)\leq \mu_1T.\nonumber
	\end{align}
	Here, (a) is obtained due to pushing the max inside the sum; (b) is obtained because $U^i_{n^i_{t-1}+1}\geq 0$ for all $i$; and (c) holds because the reward for an arm in a period is independent of the past history of play and observations. Thus, the reward of $\mu_1T$ is the highest that one can obtain under any policy. And this reward can, in fact, be obtained by the policy of always picking arm $1$. This shows that
	\[\sup_{\pi\in\Pi} \mathcal{R}_T(\pi,\bnu) = \cR^*_T(\bnu).\]
	\hfill\Halmos
	
		\subsection{Proof of Proposition \ref{prop_general:main}}\label{apx_general:prop1}
	Let $\boldi_t$ denote the set of arms pulled in period $t$ (note that $|\boldi_t| = m$ for all $t \in [T/m]$). Then, for any policy $\pi$, we have that
	\begin{align}
		\rew_T(\pi,\bnu) &= \textup{E}\left(\obj\left( \overline{U}^1_{\Tminline}, \overline{U}^2_{\Tminline} \dots, \overline{U}^K_{\Tminline} \right)\right)\nonumber\\
		&= \textup{E}\left( \obj\left(\sum_{t=1}^{\Tminline} U^1_{n^1_{t-1}+1}\mathbbm{1}_{\{ 1 \in \boldi_t\}}, \sum_{t=1}^{\Tminline} U^2_{n^2_{t-1}+1}\mathbbm{1}_{\{ 2 \in \boldi_t\}}, \dots, \sum_{t=1}^{\Tminline} U^K_{n^K_{t-1}+1}\mathbbm{1}_{\{ K \in \boldi_t\}}\right)\right)\nonumber\\
		&\overset{(a)}{\leq} \textup{E}\left( \sum_{t=1}^{\Tminline} \obj\left(U^1_{n^1_{t-1}+1}\mathbbm{1}_{\{ 1 \in \boldi_t\}}, U^2_{n^2_{t-1}+1}\mathbbm{1}_{\{ 2 \in \boldi_t\}}, \dots, U^K_{n^K_{t-1}+1}\mathbbm{1}_{\{ K \in \boldi_t\}}\right)\right)\nonumber\\
		&=\sum_{t=1}^{\Tminline} \textup{E}\left( \obj\left(U^1_{n^1_{t-1}+1}\mathbbm{1}_{\{ 1 \in \boldi_t\}}, U^2_{n^2_{t-1}+1}\mathbbm{1}_{\{ 2 \in \boldi_t\}}, \dots, U^K_{n^K_{t-1}+1}\mathbbm{1}_{\{ K \in \boldi_t\}}\right)\right)\nonumber\\
		&\overset{(b)}{=} \sum_{t=1}^{\Tminline} \textup{E}\left( \frac{1}{m} \sum_{i \in \boldi_t} U^i_{n^i_{t-1}+1}\right)\nonumber\\
		&=\sum_{t=1}^{\Tminline} \frac{1}{m} \textup{E}\left(\textup{E}\left(\sum_{i \in \boldi_t} U^i_{n^i_{t-1}+1}\mid \mathcal{H}_t\right)\right)\nonumber\\
		&\overset{(c)}{=}\sum_{t=1}^{\Tminline} \frac{1}{m} \textup{E}\left(\sum_{i \in \boldi_t} \textup{E}\left(U^i_{n^i_{t-1}+1}\mid \mathcal{H}_t\right)\right)\nonumber\\
		&\overset{(d)}{=}\sum_{t=1}^{\Tminline} \frac{1}{m} \textup{E}\left(\sum_{i \in \boldi_t} \mu_i\right) \leq \sum_{t=1}^{\Tminline} \frac{1}{m} \sum_{i = 1}^m \mu_i = \bestm \frac{T}{m}.
	\end{align}
	Here (a) is obtained due to pushing the function $\obj$ inside the sum; (b) is obtained because $U^i_{n^i_{t-1}+1}\geq 0$ for all $i$ and exactly $m$ arms are pulled in each period; (c) is obtained because, conditioned on the history and for a given policy, the set of arms that will be pulled in a period is fixed; and (d) holds because the reward for an arm in a period is independent of the past history of play and observations. Thus, the reward of $\bestm \frac{T}{m}$ is the highest that one can obtain under any policy. And this reward can, in fact, be obtained by the policy of always picking the top $m$ arms. This shows that
	\[\sup_{\pi\in\Pi} \rew_T(\pi,\bnu) = \rewR^*_T(\bnu).\]
	\hfill\Halmos

	\newpage
	\section{Proofs of Lower Bounds}\label{apx:lwb}
	\subsection{Proof of Theorem~\ref{thm:asymplwb}}\label{apx:asymplwb}
The proof of Theorem \ref{thm:asymplwb} relies on the following key result.

\begin{prop}\label{prop:auxiliary_stronger}
	Consider a class $\mathcal{V}=\mathcal{M}^K$ of $K$-armed stochastic bandits and let $(\pi_T)_{T\in\mathbb{N}}$ be a consistent sequence of policies for $\mathcal{V}$. Then, for all $\alpha \in (0, 1]$ and $\bnu \in \mathcal{V}$ such that the optimal arm $k^*$ is unique,
	\[
	\liminf\limits_{T \to \infty} \frac{\E_{\bnu} \left[ n^i_{\lceil T^\alpha\rceil} \right]}{\log(T)} \geq \frac{\alpha}{d_{\textup{inf}}\left( \nu_i, \mu^*, \mathcal{M}\right)}
	\]
	holds for each suboptimal arm $i\neq k^*$ in $\bnu$, where $\mu^*$ is the highest mean.
\end{prop}

	\proof{Proof of Proposition~\ref{prop:auxiliary_stronger}.}
	In what follows, we denote $\p_{\bnu}$ to be the probability distribution induced by the policy $\pi$ on events until time $T$ under bandit $\bnu$, and we let $\E_{\bnu}$ denote the corresponding expectation.
	
	Let $\PSR_{\textup{SUM}, T}(\pi, \bnu)$ denote the expected regret of the sum objective after $T$ pulls of policy $\pi$ under the bandit instance $\bnu$, which can be defined as
	\begin{align}
		\PSR_{\textup{SUM}, T}(\pi, \bnu) &= \mu^*T  -\E_{\bnu}\big(\sum_{t=1}^T X_t \big)\\
		&= \mu^*T  -\textup{E}_{\bnu}\big(\sum_{i=1}^K\overline{U}^i_{T}\big),
	\end{align}
	where $X_t = U^{I_t}_{n^{I_t}_t}$, which is the reward due to the arm pulled at time $t$, and $\overline{U}^i_{t} = \sum_{n=1}^{n^i_t}U^i_n$, which
	is the cumulative reward obtained from arm $i$ until time $t$. We need the following two lemmas for our proof.
	
	\begin{lemma}\label{lma:asymlwb_stronger}
		Fix $\alpha \in (0, 1]$ and a policy $\pi$. Consider a K-armed bandit instance $\bnu$ with $\mu^* \triangleqq \mu_1 \geq \mu_2 \geq \dots \geq \mu_K$. Fix a suboptimal arm $i$ and let $A_i = \left\{n^i_{\lceil T^\alpha\rceil} > \frac{T^\alpha}{2}\right\}$. Then, 
		\[\PSR_{\textup{SUM}, T}(\pi, \bnu) > \p_{\bnu}(A_i) \frac{T^\alpha \Delta_i }{2}.\]
	\end{lemma}
	
	\begin{lemma}\label{lma:asymlwb2_stronger}
		Fix $\alpha \in (0, 1]$ and a policy $\pi$. Consider a K-armed bandit instance $\bnu$ with $\mu^* \triangleqq \mu_1 \geq \mu_2 \geq \dots \geq \mu_K$. Fix a suboptimal arm $i$ and construct another K-armed bandit instance $\bnu'$ satisfying $\mu'_i > \mu^* = \mu_1 \geq \mu_2 \geq \dots \geq \mu_{i-1} \geq \mu_{i+1}\geq \dots \geq \mu_K$. Let $A_i^c = \left\{n^i_{\lceil T^\alpha\rceil} \leq \frac{T^\alpha}{2}\right\}$. Then, 
		\[\PSR_{\textup{SUM}, T}(\pi, \bnu') \geq \p_{\bnu'}(A_i^c) \frac{T^\alpha \left( \mu'_i - \mu^* \right) }{2}.\]
	\end{lemma}
	The proof of Lemma~\ref{lma:asymlwb_stronger} is presented below at the end of this section. The proof of Lemma~\ref{lma:asymlwb2_stronger} is similar and hence is omitted.
	
	Fix $\alpha \in (0, 1]$. We proceed by constructing a second bandit $\bnu'$. Fix a suboptimal arm $i$, i.e., $\Delta_i >0$, and let $\nu'_j = \nu_j$ for $j \neq i$ and pick a $\nu'_i \in \mathcal{M}$ such that $\textup{D}(\nu_i, \nu'_i) \leq d_i + \epsilon$ and $\mu'_i > \mu^*$ for some arbitrary $\epsilon >0$. 
	
	Let $\mu_i$ ($\mu'$) be the mean of arm $i$ in $\bnu$ ($\bnu'$) and $d_i \triangleqq d_{\textup{inf}}\left( \nu_i, \mu^*, \mathcal{M}\right)$. 
	Recall that $d_{\textup{inf}}\left(\nu , \mu^* , \mathcal{M}\right) = \inf\limits_{\nu' \in \mathcal{M}} \left\{ \textup{D}(\nu, \nu') : \mu(\nu') > \mu^* \right\}$ where $\mu(\nu)$ denotes the mean of distribution $\nu$.
	
	
	Since any lower bound on the regret for the sum objective implies the same lower bound on the max objective, using Lemma~\ref{lma:asymlwb_stronger} and Lemma~\ref{lma:asymlwb2_stronger}, we have the following:
	\begin{align}
		\PSR_T(\pi, \bnu) + \PSR_T(\pi, \bnu') & \geq \PSR_{\textup{SUM}, T}(\pi, \bnu)+\PSR_{\textup{SUM}, T}(\pi, \bnu')\notag\\
		& > \frac{T^\alpha}{2} \left( \p_{\bnu}\left( A_i \right) \Delta_i + \p_{\bnu'}\left( A_i^c\right) (\mu'_i - \mu^*) \right)\notag\\
		&\geq \frac{T^\alpha}{2} \min\{\Delta_i, (\mu'_i - \mu^*)\} \left( \p_{\bnu}\left( A_i \right)+ \p_{\bnu'}\left( A_i^c\right) \right)\notag\\
		&= \frac{T^\alpha}{2} \min\{\Delta_i, (\mu'_i - \mu^*)\} \left( \overline{\p}_{\bnu}\left( A_i \right)+ \overline{\p}_{\bnu'}\left( A_i^c\right) \right)\notag\\
		&\geq \frac{T^\alpha}{4} \min\{\Delta_i, (\mu'_i - \mu^*)\} \exp\left( -\E_{\bnu}\left[ n^i_{\lceil T^\alpha\rceil} \right] (d_i +\epsilon) \right). \label{eq:lwb_reg_expr}
	\end{align}
	Here, $\overline{\p}_{\bnu}$ ($\overline{\p}_{\bnu'}$) is the probability distribution induced by the policy $\pi$ on events until time $\lceil T^\alpha\rceil$ under bandit $\bnu$ ($\bnu'$). The equality then results from the fact that the two events $\{n^i_{\lceil T^\alpha \rceil} > \frac{T^\alpha}{2}\}$ and $\{n^i_{\lceil T^\alpha\rceil} \leq \frac{T^\alpha}{2}\}$ depend only on the play until time $\lceil T^\alpha\rceil$. The last inequality follows from using the Bretagnolle-Huber inequality and divergence decomposition (see Theorem 14.2 and Lemma 15.1 in \cite{lattimore2018bandit}, respectively) combined with the fact that $\textup{D}(\nu_i, \nu'_i) \leq d_i + \epsilon$:
	\begin{align}
		\overline{\p}_{\bnu}\left( A_i\right) + \overline{\p}_{\bnu'}\left( A_i^c\right) \geq \frac{1}{2} \exp\left( - \textup{D}\left( \overline{\p}_{\bnu}, \overline{\p}_{\bnu'} \right) \right) \geq \frac{1}{2} \exp\left( - \E_{\bnu} \left[ n^i_{\lceil T^\alpha\rceil} \right] \left( d_i + \epsilon \right)\right), \label{eq:lwbasympthuber}
	\end{align}
	where the events $A_i$ and $A_i^c$ are defined as they have been in Lemmas \ref{lma:asymlwb_stronger} and~\ref{lma:asymlwb2_stronger} for the fixed arm $i$.
	
	Rearranging Equation~\ref{eq:lwb_reg_expr}, we obtain
	\begin{align}
		\frac{\E_{\bnu}\left[ n^i_{\lceil T^\alpha\rceil} \right]}{\log(T)} > \frac{1}{d_i +\epsilon} \frac{\log\left( \frac{T^\alpha  \min\{\Delta_i, ~\mu'_i - \mu^*\}}{4 (\PSR_T(\pi, \bnu) + \PSR_T(\pi, \bnu'))} \right)}{\log(T)}, 
	\end{align}
	and taking the limit inferior yields
	\begin{align}
		\liminf\limits_{T \to \infty} \frac{\E_{\bnu}\left[ n^i_{\lceil T^\alpha\rceil} \right]}{\log(T)} &> \frac{1}{d_i +\epsilon} \liminf\limits_{T \to \infty} \frac{\log\left(  \frac{T^\alpha  \min\{\Delta_i, ~\mu'_i - \mu^*\}}{4 (\PSR_T(\pi, \bnu) + \PSR_T(\pi, \bnu'))} \right)}{\log(T)},\notag\\
		&= \frac{1}{d_i +\epsilon} \liminf\limits_{T \to \infty} \frac{\alpha \log(T) + \log( \beta_i) - \log(4) - \log\left( \PSR_T(\pi, \bnu) + \PSR_T(\pi, \bnu') \right)}{\log(T)}\notag\\
		&= \frac{1}{d_i +\epsilon} \left(\alpha - \limsup\limits_{T \to \infty} \frac{\log\left( \PSR_T(\pi, \bnu) + \PSR_T(\pi, \bnu') \right)}{\log(T)} \right)\notag\\
		&\geq \frac{\alpha}{d_i +\epsilon}, \label{eq:consistent}
	\end{align}
	where $\beta_i = \min\{\Delta_i, \mu'_i - \mu^*\}$.
	
	Since $\pi$ is a consistent policy over the class $\mathcal{V}$, we can find a constant $c_p$ for any $p>0$ such that $\PSR_T(\pi, \bnu) + \PSR_T(\pi, \bnu') \leq c_p T^p$, which implies 
	\begin{align}
		\limsup\limits_{T \to \infty} \frac{\log\left( \PSR_T(\pi, \bnu) + \PSR_T(\pi, \bnu') \right)}{\log(T)} \leq \limsup\limits_{T \to \infty} \frac{p\log(T) + \log(c_p)}{\log(T)} = p. \label{eq:consistent_2}
	\end{align}
	Then, Equation~\ref{eq:consistent} follows from Equation \ref{eq:consistent_2} and the fact that $p>0$ is arbitrary. Since $\epsilon>0$ is arbitrary as well, we have
	\begin{align}
		\liminf\limits_{T \to \infty} \frac{\E_{\bnu}\left[ n^i_{\lceil T^\alpha\rceil} \right]}{\log(T)} \geq \frac{\alpha}{d_i} \label{eq:ub_on_pulls_asymp_2}
	\end{align}
	for each $i \neq k^*$, i.e., each suboptimal arm $i$ in $\bnu$.
	\hfill\Halmos\\
	\endproof
	
	We present the proof of Lemma~\ref{lma:asymlwb_stronger} before proceeding with the proof of Theorem~\ref{thm:asymplwb}.\\
	
	\proof{Proof of Lemma~\ref{lma:asymlwb_stronger}.}
	Recall that $I_t$ is the arm pulled at time $t$ and $X_t$ is the reward due to arm pulled at time $t$, i.e., $X_t \sim \nu_{I_t}$. Then, due to, e.g., Lemma 4.5 in \cite{lattimore2018bandit}, we can decompose the expected regret as
	\begin{align}
		\PSR_{\textup{SUM}, T}(\pi, \bnu)  = \sum_{\substack{j = 1\\ j\neq i}}^K \Delta_j \E_{\bnu}( n^j_T ) + \Delta_i \E_{\bnu}( n^i_T ).
	\end{align}
	Due to the non-negativity of expected number of pulls and the suboptimality gaps, we have
	\[
	\PSR_{\textup{SUM}, T}(\pi, \bnu) \geq \Delta_i \E_{\bnu}( n^i_T ).
	\]
	Now, we look at $\E_{\bnu}( n^i_T )$:
	\begin{align}
		\E_{\bnu}( n^i_T ) &= \E_{\bnu}\big( n^i_T \mid A_i\big) \p_{\bnu}(A_i) + \E_{\bnu}\big( n^i_T \mid A_i^c \big) \p_{\bnu}(A_i^c) \notag\\
		& \geq \E_{\bnu}\big( n^i_T \mid A_i\big) \p_{\bnu}(A_i) \notag\\
		& \overset{(a)}{>} \frac{T^\alpha}{2} \p_{\bnu}(A_i),
	\end{align}
	where (a) is due to event $A_i = \left\{n^i_{\lceil T^\alpha\rceil} > \frac{T^\alpha}{2}\right\}$. Finally, we have
	\begin{align}
		\PSR_{\textup{SUM}, T}(\pi, \bnu) >  \p_{\bnu}(A_i) \frac{T^\alpha \Delta_i }{2}.
	\end{align}	
	\hfill\Halmos
	\endproof
	
	\vspace{1cm}
\proof{Proof of Theorem~\ref{thm:asymplwb}.}
	
	Let $k^*$ denote the unique optimal arm in $\bnu$ and, without loss of generality, let $k^* = 1$, i.e., $\mu^* = \mu_1$.
	Let $I^*$ denote the arm with the highest cumulative reward after $T$ pulls and recall that $n^i_T$ denotes the number of pulls spent on arm $i$ until time $T$. Since all of the following expectations are over $\bnu$, we drop the subscript of $\bnu$ hereafter. We first look at the expected regret:
	\begin{align}
		\PSR_T(\pi, \bnu) &= \mu^*~T - \E\left[ \max\left( \overline{U}^1_{T}, \overline{U}^2_{T}, \dots, \overline{U}^K_{T} \right) \right]\\
		&\overset{(a)}{=} \E\left[ \sum_{t=1}^{T} U^1_t\right]  - \E\left[ \max\left( \overline{U}^1_{T}, \overline{U}^2_{T}, \dots, \overline{U}^K_{T} \right)  \mathbbm{1}_{\{I^* = 1\}} \right] - \E\left[ \max\left( \overline{U}^1_{T}, \overline{U}^2_{T}, \dots, \overline{U}^K_{T} \right)  \mathbbm{1}_{\{I^* \neq 1\}} \right]\\
		& \overset{(b)}{=} \E\left[ \sum_{t=1}^{T} U^1_t~\mathbbm{1}_{\{I^* = 1\}} \right]  +  \E\left[ \sum_{t=1}^{T} U^1_t~\mathbbm{1}_{\{I^* \neq 1\}} \right] - \E\left[ \sum_{t=1}^{n^1_T} U^1_t  ~\mathbbm{1}_{\{I^* = 1\}} \right] - \sum_{i\neq 1}\E\left[ \sum_{t=1}^{n^i_T}U^i_t ~\mathbbm{1}_{\{I^* = i\}} \right] \\
		& = \E\left[ \left( \sum_{t=1}^{T} U^1_t- \sum_{t=1}^{n^1_T} U^1_t \right) ~\mathbbm{1}_{\{I^* = 1\}} \right] +  \sum_{i\neq 1}\E\left[ \left( \sum_{t=1}^{T} U^1_t- \sum_{t=1}^{n^i_T}U^i_t \right) ~\mathbbm{1}_{\{I^* = i\}} \right] \\
		& \overset{(c)}{\geq}  \E\left[ \left( \sum_{t=1}^{T} U^1_t- \sum_{t=1}^{n^1_T} U^1_t \right) ~\mathbbm{1}_{\{I^* = 1\}} \right] +  \sum_{i\neq 1}\E\left[ \left( \sum_{t=1}^{T} U^1_t- \sum_{t=1}^{T}U^i_t \right) ~\mathbbm{1}_{\{I^* = i\}} \right]\\
		&=\E\left[ \left( \sum_{t=n^1_T+1}^{T} U^1_t\right) ~\mathbbm{1}_{\{I^* = 1\}} \right] +  \sum_{i\neq 1}\E\left[ \left( \sum_{t=1}^{T} U^1_t- \sum_{t=1}^{T}U^i_t \right) ~\mathbbm{1}_{\{I^* = i\}} \right]\\
		&\overset{(d)}{=}\mu^*\E\left[ (T-n^1_T)~\mathbbm{1}_{\{I^* = 1\}} \right] +  \sum_{i\neq 1}\E\left[ \left( \sum_{t=1}^{T} U^1_t- \sum_{t=1}^{T}U^i_t \right) ~\mathbbm{1}_{\{I^* = i\}} \right]\\
		&\overset{(e)}{=}\mu^*\sum_{i\neq 1}\E\left[ n^i_T~\mathbbm{1}_{\{I^* = 1\}} \right]+  \sum_{i\neq 1}\E\left[ \left( \sum_{t=1}^{T} U^1_t- \sum_{t=1}^{T}U^i_t \right) ~\mathbbm{1}_{\{I^* = i\}} \right]. \label{eqn:regretlwb}
	\end{align}
	
	Here, (a) is due to the fact that $\E\left[ U^1_t\right] = \mu^*$ for $t \in [T]$. (b) follows from the definition of $I^*$. (c) results from $n^i_T \leq T$ for $i \in [K]$. (d) is due to the fact that the future rewards from the first arm is independent of the past history of play and observations of policy $\pi$. Finally, (e) follows from the identity $T = \sum_{i = 1}^K n_T^i $. 
	
	We first focus on bounding the second term in the Expression~\ref{eqn:regretlwb}. In order to do that, for each suboptimal arm $i$, $i \neq 1$, define a ``good'' event
	\[ 
	G_i = \left\{ \overline{U}^1_{T} > \overline{U}^i_{T} + \frac{T\Delta_i}{2} \right\}.
	\]
	Notice that, for $i \neq 1$, $\Delta_i > 0$.
	
	We proceed by showing that event $G_i$ occurs with high probability. To that end, consider the complement event
	\begin{align}
		P\left(G_i^c\right) &= P\left( \overline{U}^1_{T} \leq \overline{U}^i_{T} + \frac{T\Delta_i}{2}  \right)\notag\\
		&= P\left( \frac{\overline{U}^1_{T} - \overline{U}^i_{T}}{T} - (\mu_1 - \mu_i) \leq \frac{\Delta_i}{2}  - (\mu_1 -\mu_i) \right).
	\end{align}
	By Hoeffding's inequality, 
	\begin{align}
		P\left(G_i^c\right) \leq \exp\left( -\frac{2T^2\left( \frac{\Delta_i}{2}  \right)^2 }{4T} \right) = \exp\left( -\frac{T\Delta_i^2}{8}\right),\notag
	\end{align}
	since $-1 \leq U^1_{t_1} - U^i_{t_2} \leq 1$ for any pair $t_1, t_2 \in [T]$.
	We thus also have that 
	\begin{align}
		P(I^* =i, G_i^c) \leq \exp\left( -\frac{T\Delta_i^2}{8}\right).\label{eqn:easybound} 
	\end{align}
	We then have
	\begin{align}
		&\E\left[ \left( \sum_{t=1}^{T} U^1_t- \sum_{t=1}^{T}U^i_t \right) ~\mathbbm{1}_{\{I^* = i\}} \right]\notag\\
		&~~=\E\left[ \left( \sum_{t=1}^{T} U^1_t- \sum_{t=1}^{T}U^i_t \right)\mid I^* = i, G_i\right] P(I^* = i, G_i) + \E\left[ \left( \sum_{t=1}^{T} U^1_t- \sum_{t=1}^{T}U^i_t \right)\mid I^* = i, G_i^c\right] P(I^* = i, G_i^c)\notag\\
		&~~\geq \frac{T\Delta_i}{2}P(I^* = i, G_i) -T P(I^* = i, G_i^c)\notag\\
		&~~\geq \frac{T\Delta_i}{2}P(I^* = i, G_i) -\textup{O}(1) \label{eqn:touselater}\\
		&~~\geq 0 -  \textup{O}(1)\label{eqn:regsecond}.
	\end{align}
	Thus the second term in \ref{eqn:regretlwb} is lower bounded by a (instance-dependent) constant. 
	
	Next, we bound the first term in \ref{eqn:regretlwb}. To do so, we first need an upper bound on $P(I^* = i)$ for any $i\neq 1$.
	By consistency of policy $\pi$, we have that $\PSR_T(\pi, \bnu)\leq \textup{o}(T^p)$ for every $p>0$. Thus from \ref{eqn:regretlwb} and \ref{eqn:touselater}, for any $i\neq 1$, we have that
	\begin{align}
		\textup{o}(T^p)&\geq \E\left[ \left( \sum_{t=1}^{T} U^1_t- \sum_{t=1}^{T}U^i_t \right) ~\mathbbm{1}_{\{I^* = i\}} \right]\geq \frac{T\Delta_i}{2}P(I^* = i, G_i) -\textup{O}(1).
	\end{align}
	
	This implies that for any $i\neq 1$,
	\begin{align}
		P(I^* = i, G_i) \leq \textup{o}(T^{p-1}), \label{eqn:diffbound}
	\end{align}
	for every $p>0$. Finally, \ref{eqn:easybound} and \ref{eqn:diffbound} together imply that, for any $i\neq 1$, $P(I^* = i ) = P(I^* = i, G_i)+ P(I^* = i, G_i^c) \leq \textup{o}(T^{p-1})$ for every $p>0$.
	
	Finally, we are ready to derive a lower bound on the first term in the expression \ref{eqn:regretlwb}. For any $\alpha \in (0,1)$, we have
	\begin{align}
		\E[ n^i_{\lceil T^\alpha\rceil}] &= \E\left[ n^i_{\lceil T^\alpha\rceil}~\mathbbm{1}_{\{I^* = 1\}}\right]  +\E\left[ n^i_{\lceil T^\alpha\rceil}~\mathbbm{1}_{\{I^* \neq 1\}} \right]\notag\\
		&\leq \E\left[ n^i_{\lceil T^\alpha\rceil}~\mathbbm{1}_{\{I^* = 1\}}\right]  + \lceil T^\alpha\rceil P(I^* \neq 1)\notag\\
		&\leq \E\left[ n^i_{T}~\mathbbm{1}_{\{I^* = 1\}}\right]  + \lceil T^\alpha\rceil P(I^* \neq 1)\notag\\
		&\leq \E\left[ n^i_{T}~\mathbbm{1}_{\{I^* = 1\}}\right]  + \textup{o}(T^{\alpha +p-1}),
	\end{align}
	for every $p>0$. But then from Proposition~\ref{prop:auxiliary_stronger}, we have 
	\begin{align}
		\frac{\alpha}{d_i}& \leq  \liminf\limits_{T\rightarrow \infty} \frac{\E[ n^i_{\lceil T^\alpha\rceil}]}{\log T}\\ 
		&\leq \liminf\limits_{T\rightarrow \infty}\frac{\E\left[ n^i_{T}~\mathbbm{1}_{\{I^* = 1\}}\right]}{\log T}  + \liminf\limits_{T\rightarrow \infty}\frac{\textup{o}(T^{\alpha +p-1})}{\log T}.
	\end{align}
	By choosing a $p$ such that $0<p<1-\alpha$, we have that $\liminf_{T\rightarrow \infty}\frac{\textup{o}(T^{\alpha +p-1})}{\log T} = 0$. And thus, for every $\alpha \in (0,1)$, we have
	\begin{align}
		\liminf\limits_{T\rightarrow \infty}\frac{\E\left[ n^i_{T}~\mathbbm{1}_{\{I^* = 1\}}\right]}{\log T} \geq \frac{\alpha}{d_i},
	\end{align}
	which implies that 
	\begin{align}
		\liminf\limits_{T\rightarrow \infty}\frac{\E\left[ n^i_{T}~\mathbbm{1}_{\{I^* = 1\}}\right]}{\log T} \geq \frac{1}{d_i}\label{eqn:regfirst}.
	\end{align}
	Finally, putting everything together, from \ref{eqn:regretlwb}, \ref{eqn:regsecond}, and \ref{eqn:regfirst}, we have
	\begin{align}
		\liminf\limits_{T\rightarrow \infty}\frac{\PSR_T(\pi, \bnu)}{\log T} &\geq \liminf\limits_{T\rightarrow \infty} \mu^*\sum_{i\neq 1}\frac{\E\left[ n^i_T~\mathbbm{1}_{\{I^* = 1\}} \right]}{\log T} -\liminf\limits_{T\rightarrow \infty} \sum_{i\neq 1}\frac{\textup{O}(1)}{\log T}\notag\\
		&\geq  \sum_{i\neq 1}\frac{\mu^*}{d_i}.
	\end{align}
	Plugging in the definition of $d_i$ and substituting $k^*$ back in place give the desired result.
	\hfill\Halmos
	\endproof

	\subsection{Proof of Theorem \ref{thm:lb}}\label{apx:thm1}
	First we fix a policy $\pi \in \Pi$. Let $\Delta \triangleqq (K-1)^{1/3}/(2T^{1/3})$. We construct two bandit environments with different reward distributions for each of the arms and show that $\pi$ cannot perform well in both environments simultaneously.
	
	We first specify the reward distribution for the arms in the base environment, denoted as the bandit $\boldsymbol{\nu} = \left\{ \nu_1, \dots, \nu_K \right\}$. Assume that the reward for all of the arms have the Bernoulli distribution, i.e., $\nu_i \sim \textup{Bernoulli}(\mu_i)$. We let $\mu_1 = \frac{1}{2}+\Delta$, and $\mu_i = \frac{1}{2}$ for $2 \leq i \leq K$. We let $\p_{\boldsymbol{\nu}}$ denote the probability distribution induced over events until time $T$ under policy $\pi$ in this first environment, i.e., in bandit $\bnu$. Let $\E_{\boldsymbol{\nu}}$ denote the expectation under $\p_{\boldsymbol{\nu}}$.
	
	Define $n^i_{\lceil\Delta T\rceil}$ as the (random) number of pulls spent on arm $i \in \{1,\dots,K\}$ until time $\lceil\Delta T\rceil$ (note that $\sum_{i = 1}^K n^i_{\lceil\Delta T\rceil} = \lceil\Delta T\rceil$) under policy $\pi$. Specifically, $n^1_{\lceil\Delta T\rceil}$ is the total (random) number of pulls spent on the first arm under policy $\pi$ until time $\lceil\Delta T\rceil$. Under policy $\pi$, let $l^*$ denote the arm in the set $[K]\setminus \{1\}$ that is pulled the least in expectation until time $\lceil\Delta T\rceil$, i.e., $l^* \in \arg\min_{2\leq i\leq K} \E_{\bnu}(n^i_{\lceil\Delta T\rceil}).$
	Then clearly, we have that $\E_{\bnu}(n^{l^*}_{\lceil\Delta T\rceil})\leq \frac{\lceil\Delta T\rceil}{K-1}$. 
	
	Having defined $l^*$, we can now define the second environment, denoted as the bandit $\boldsymbol{\nu'} = \left\{ \nu'_1, \dots, \nu'_K \right\}$. Again, assume that the reward for all of the arms have the Bernoulli distribution, i.e., $\nu'_i \sim \textup{Bernoulli}(\mu'_i)$. We let $\mu'_1 = \frac{1}{2}+\Delta$, $\mu'_i = \frac{1}{2}$ for $[2 \leq i \leq K]\setminus \{l^*\}$, and $\mu'_{l^*} = \frac{1}{2}+2\Delta$. We let $\p_{\bnu'}$ denote the probability distribution induced over events until time $T$ under policy $\pi$ in this second environment, i.e., in bandit $\bnu'$. Let $\E_{\bnu'}$ denote the expectation under $\p_{\bnu'}$.
	
	With some abuse of notation, for any event $B$, we define:
	\begin{align}
		\PSR_T(\pi, \bnu, B) = \mu^*T\p_{\bnu}(B)  -\E_{\bnu}\big(\max\big(\overline{U}^1_{T}, \overline{U}^2_{T}, \dots, \overline{U}^K_{T}\big)\mathbbm{1}_{B}\big).
	\end{align}
	It is then clear that $\PSR_T(\pi, \bnu) = \PSR_T(\pi, \bnu, B)+ \PSR_T(\pi, \bnu, B^c).$ 
	We need the following two results for our proof.

	
	\begin{lemma}\label{lma:1}
		Fix a policy $\pi$. Consider the K-armed bandit instance $\bnu$ with Bernoulli rewards and mean vector $\bmu =(\frac{1}{2} +\Delta, \frac{1}{2}, \frac{1}{2},\cdots,\frac{1}{2})$, where $\Delta <\frac{1}{2}$. Consider the event $ A =\{n^1_{\lceil\Delta T\rceil}\leq\frac{\Delta T}{2}\}$. Then we have,
		\[\PSR_T(\pi, \bnu, A) \geq\frac{\Delta T}{4}\p_{\bnu}(A)-2\sqrt{T\log(KT)}-2.\]
	\end{lemma}
	
	The proof of Lemma~\ref{lma:1} is presented below in this section. A similar argument shows the following.
	\begin{lemma}\label{lma:2}
		Fix a policy $\pi$. Consider the K-armed bandit instance $\bnu'$ with Bernoulli rewards and mean vector $\bmu' =(\frac{1}{2} +\Delta, \frac{1}{2}, \frac{1}{2},\cdots,\frac{1}{2}, \frac{1}{2}+2\Delta)$, where $\Delta <\frac{1}{4}$. Consider the event $A^c=\{n^1_{\lceil\Delta T\rceil}>\frac{\Delta T}{2}\}$. Then we have,
		\[\PSR_T(\pi, \bnu', A^c) \geq\frac{\Delta T}{4}\p_{\bnu'}(A^c)-2\sqrt{T\log(KT)}-2.\]
	\end{lemma}
	The proof of Lemma~\ref{lma:2} is omitted since it is almost identical to that of Lemma~\ref{lma:1}. 
	These two facts result in the following two inequalities: 
	\begin{align}
		\PSR_T(\pi, \bnu, A) &\geq \p_{\bnu}\left( n^1_{\lceil\Delta T\rceil} \leq \frac{\Delta T}{2} \right) \Omega(\Delta T), \textrm{ and }\\
		\PSR_T(\pi, \bnu', A^c) &\geq \p_{\bnu'}\left( n^1_{\lceil\Delta T\rceil} > \frac{\Delta T}{2} \right) \Omega(\Delta T).
	\end{align}
	
	Note that here we have ignored the lower order $\sqrt{T\log(KT)}$ terms since $\Delta T = \Theta(T^{2/3}K^{1/3})$. Now, using the Bretagnolle-Huber inequality (see Theorem 14.2 in \cite{lattimore2018bandit}), we have,
	\begin{align}
		\PSR_T(\pi, \bnu, A) + \PSR_T(\pi, \bnu', A^c) & \geq \Omega(\Delta T) \left( \p_{\bnu}\left( n^1_{\lceil\Delta T\rceil} \leq \frac{\Delta T}{2}\right) + \p_{\bnu'}\left( n^1_{\lceil\Delta T\rceil} > \frac{\Delta T}{2}\right) \right)\\
		&= \Omega(\Delta T) \left( \overline{\p}_{\bnu}\left( n^1_{\lceil\Delta T\rceil} \leq \frac{\Delta T}{2}\right) + \overline{\p}_{\bnu'}\left( n^1_{\lceil\Delta T\rceil} > \frac{\Delta T}{2}\right) \right)\\
		&\geq \Omega(\Delta T) \exp\left( - \text{D}\left( \overline{\p}_{\bnu}, \overline{\p}_{\bnu'} \right) \right).
	\end{align}
	Here, $\overline{\p}_{\bnu}$ ($\overline{\p}_{\bnu'}$) is the probability distribution induced by the policy $\pi$ on events until time $\lceil\Delta T\rceil$ under bandit $\bnu$ ($\bnu'$). The first equality then results from the fact that the two events $\{n^1_{\lceil\Delta T\rceil} \leq \frac{\Delta T}{2}\}$ and $\{n^1_{\lceil\Delta T\rceil} > \frac{\Delta T}{2}\}$ depend only on the play until time $\lceil\Delta T\rceil$. In the second inequality, which results from the Bretagnolle-Huber inequality, $\text{D}\left( \overline{\p}_{\bnu}, \overline{\p}_{\bnu'} \right)$ is the relative entropy, or the Kullback-Leibler (KL) divergence between the distributions $\overline{\p}_{\bnu}$ and $\overline{\p}_{\bnu'}$ respectively.  
	We can upper bound $ \text{D}\left( \overline{\p}_{\bnu}, \overline{\p}_{\bnu'} \right)$ as,
	\begin{align}
		\text{D}\left( \overline{\p}_{\bnu}, \overline{\p}_{\bnu'} \right) = \E_{\bnu}(n^{l^*}_{\lceil\Delta T\rceil}) \text{D}\left( \nu_{l^*}, \nu'_{l^*} \right) \leq \dfrac{\lceil\Delta T\rceil}{K-1} \text{D}\left( \nu_{l^*}, \nu'_{l^*} \right)\lessapprox \dfrac{8\Delta^3 T}{K-1}, \label{ub_on_KL}
	\end{align}
	where $\nu_{l^*}$ ($ \nu'_{l^*} $) denotes the reward distribution of arm $l^*$ in the first (second) environment. The first equality results from divergence decomposition (see Lemma 15.1 in \cite{lattimore2018bandit}) and the fact  no arm other than $l^*$ offers any distinguishability between $\bnu$ and $\bnu'$. The next inequality follows from the fact that $\E_{\bnu}[n^{l^*}_{\lceil\Delta T\rceil}]\leq (\lceil\Delta T\rceil)/(K-1)$, since by definition, $l^*$ is the arm that is pulled the least in expectation until time $\lceil\Delta T\rceil$ in bandit $\bnu$ under $\pi$. Now $\text{D}\left( \nu_{l^*}, \nu'_{l^*} \right)$ is simply the relative entropy between the distributions $\textup{Bernoulli}(1/2)$ and $\textup{Bernoulli}(1/2+2\Delta)$, which, by elementary calculations, can be shown to be at most $8\Delta^2$, resulting in the final inequality.
	Thus, we finally have,
	\[
	\PSR_T(\pi, \bnu, A) + \PSR_T(\pi, \bnu', A^c)  \geq  \Omega(\Delta T)\text{exp}\left( - \frac{8\Delta^3T}{K-1} \right).
	\]
	
	Substituting $\Delta = (K-1)^{1/3}/(2T^{1/3})$ gives
	\begin{align}
		\PSR_T(\pi, \bnu, A) + \PSR_T(\pi, \bnu', A^c)  \geq \Omega\left((K-1)^{1/3} T^{2/3} \right).\label{eqn:part1}
	\end{align}
	
	Equation~\ref{eqn:part1} along with
	\begin{align}
		\PSR_T(\pi, \bnu, A^c) \geq - \textup{O} (\sqrt{T\log (KT)})&\text{ and }\label{eqn:regretlb_1}\\
		\PSR_T(\pi, \bnu', A) \geq - \textup{O} (\sqrt{T\log (KT)})&,\label{eqn:regretlb_2}
	\end{align}
	imply that
	\begin{align}
		\PSR_T(\pi, \bnu) + \PSR_T(\pi, \bnu')  \geq \Omega\left((K-1)^{1/3} T^{2/3} \right).
	\end{align}
	Finally, using $2\max\{a,b\} \geq a +b$ gives the desired lower bound on the regret.
	
	Showing Equations~\ref{eqn:regretlb_1} and~\ref{eqn:regretlb_2} is an easy exercise:
	\begin{align}
		\PSR_T(\pi, \bnu, A^c) &= \mu^*T\p_{\bnu}(A^c)  -\E_{\bnu}\big(\max\big(\overline{U}^1_{T}, \overline{U}^2_{T}, \dots, \overline{U}^K_{T}\big)\mathbbm{1}_{A^c}\big)\nonumber\\
		&\geq \mu^*T\p_{\bnu}(A^c)  - \E_{\bnu}\big(\max\big(\sum_{t=1}^{T}U^1_t, \sum_{t=1}^{T}U^2_t, \dots, \sum_{t=1}^{T}U^K_t\big)\mathbbm{1}_{A^c}\big)\nonumber\\
		&\overset{(a)}{\geq} \mu^*T\p_{\bnu}(A^c)  - \mu^*T\p_{\bnu}(A^c)-2\sqrt{T\log(KT)}-2\nonumber\\
		&= -2\sqrt{T\log(KT)}-2.\label{eqn:part2a} 
	\end{align}
	Here, (a) follows from an argument essentially identical to the one in the proof of Lemma~\ref{lma:1} below and we do not repeat it here for brevity. Similarly, we can show that 
	\begin{align}
		\PSR_T(\pi, \bnu', A)\geq -2\sqrt{T\log(KT)}-2.\label{eqn:part2b}
	\end{align}
	\hfill\Halmos
	
\proof{Proof of Lemma~\ref{lma:1}.}
	We first have that
	\begin{align}
		\E_{\bnu}\big(\max\big(\overline{U}^1_{T}, \overline{U}^2_{T}, \dots, \overline{U}^K_{T}\big)\mathbbm{1}_{A}\big) &= \E_{\bnu}\big(\max\big(\sum_{t=1}^{n^1_T}U^1_t, \sum_{t=1}^{n^2_T}U^2_t, \dots, \sum_{t=1}^{n^K_T}U^K_t\big)\mathbbm{1}_{A}\big)\\
		&\leq \E_{\bnu}\big(\max\big(\sum_{t=1}^{ T-\lceil\frac{T\Delta}{2}\rceil}U^1_t, \sum_{t=1}^{T}U^2_t, \dots, \sum_{t=1}^{T}U^K_t\big)\mathbbm{1}_{A}\big).\label{eqn:here1}
	\end{align}
	Defining $T_1 = T-\lceil\frac{T\Delta }{2}\rceil$, and $T_i = T$ for all $i>1$, consider the ``good'' event 
	\[G=\left\{ |\sum_{t=1}^{T_j}U^j_t - \mu_jT_j| \leq \sqrt{T\log (KT)} \textrm{ for all }j \right\}.\]
	Since $U^j_t\in[0,1]$, by Hoeffding's inequality, we have that for any $T'\leq T$, 
	\begin{align*}
		\p_{\bnu}\left( |\sum_{t=1}^{T'}U^j_t - \mu_jT'| \leq \sqrt{T\log (KT)}\right)&\geq 1-2\exp(-\frac{2(\sqrt{T\log (KT)})^2}{T'})\\
		&\geq 1-2\exp(-\frac{2(\sqrt{T\log (KT)})^2}{T})\\
		&= 1-\frac{2}{K^2T^2}\geq 1-\frac{2}{KT}.
	\end{align*}
	Hence, by the union bound we have that $P\left( G \right)\geq 1-\frac{2}{T}$.  Thus we finally have,
	
	\begin{align}
		&\E_{\bnu}\big(\max\big(\sum_{t=1}^{ T-\lceil\frac{T\Delta}{2}\rceil}U^1_t, \sum_{t=1}^{T}U^2_t, \dots, \sum_{t=1}^{T}U^K_t\big)\mathbbm{1}_{A}\big)\nonumber\\
		&~~= \E_{\bnu}\big(\max\big(\sum_{t=1}^{ T-\lceil\frac{T\Delta}{2}\rceil}U^1_t, \sum_{t=1}^{T}U^2_t, \dots, \sum_{t=1}^{T}U^K_t\big)\mathbbm{1}_{A,\,G}\big)\nonumber\\
		&~~~~+\E_{\bnu}\big(\max\big(\sum_{t=1}^{ T-\lceil\frac{T\Delta}{2}\rceil}U^1_t, \sum_{t=1}^{T}U^2_t, \dots, \sum_{t=1}^{T}U^K_t\big)\mathbbm{1}_{A}\mid G^c\big)\p_{\bnu}(G^c)\nonumber\\
		&~~\leq \E_{\bnu}\left(\left(\max_{i\in[K]}\mu_i T_i +  2\sqrt{T\log (KT)}\right)\mathbbm{1}_{A,\,G}\right)+ \frac{2}{T}\times T\nonumber\\
		&~~\leq \max\left((\frac{1}{2}+\Delta)(T-\frac{\Delta  T}{2}), \frac{T}{2}\right) \p_{\bnu}(A)+  2\sqrt{T\log (KT)} + 2\nonumber\\
		&~~\overset{(a)}{=}(\frac{1}{2}+\Delta)(T-\frac{\Delta T}{2})\p_{\bnu}(A) +  2\sqrt{T\log (KT)} + 2.\label{eqn:here2}
	\end{align}
	Here, (a) follows from the fact that $\Delta<\frac{1}{2}$. Thus, from Equations~\ref{eqn:here1} and~\ref{eqn:here2}, we finally have,
	\begin{align}
		\PSR_T(\pi, \bnu, A) &= (\frac{1}{2}+\Delta)T\p_{\bnu}(A) -\E_{\bnu}\big(\max\big(\overline{U}^1_{T}, \overline{U}^2_{T}, \dots, \overline{U}^K_{T}\big)\mathbbm{1}_{A}\big)\nonumber\\
		&\geq (\frac{1}{2}+\Delta)T\p_{\bnu}(A) - (\frac{1}{2}+\Delta)(T-\frac{\Delta T}{2})\p_{\bnu}(A) -  2\sqrt{T\log (KT)} - 2\nonumber\\
		&= (\frac{1}{2}+\Delta)\frac{\Delta T}{2}\p_{\bnu}(A)-  2\sqrt{T\log (KT)} - 2\nonumber\\
		&\geq \frac{\Delta T}{4}\p_{\bnu}(A)-  2\sqrt{T\log (KT)} - 2.
	\end{align}
	\hfill\Halmos
	\endproof
	
	\subsection{Proof of Theorem~\ref{thm:sumbad}}\label{apx:sumbad}
	
	To prove this result, we first need the following result, which bounds the expected reward for the max objective in a two-armed bandit instance by the expected maximum number of times either of the arms is pulled.
	\begin{lemma}\label{lma:forUCB}
		Consider a $2$-armed stochastic bandit instance $\bnu$ with means $\mu_1>\mu_2$, where $\mu_1 - \mu_2 = \Delta$. Then, the expected reward in this instance under the max objective for any policy $\pi$, given $T$ pulls, can be upper bounded by $T \Delta + \mu_2~ \E_{\bnu} \left[ \max\{n^1_T,  n^2_T\} \right] + \textup{O}(\sqrt{T\log(T)})$.
	\end{lemma}
	\proof{Proof.}
Recalling that $n^i_t$ denotes the number of pulls spent on arm $i$ at time $t$, for any policy $\pi$ on bandit $\bnu$, we have the following expression for the reward under the max objective after all $T$ pulls are depleted,
\[
\max \left\{ \sum_{t=1}^{n^1_T} U^1_t, \sum_{t=1}^{n^2_T} U^2_t \right\}.
\]

Consider the expected reward of policy $\pi$ on bandit $\bnu$
\begin{align}
	\E_{\bnu}\left[ \max \left\{ \sum_{t=1}^{n^1_T} U^1_t, \sum_{t=1}^{n^2_T} U^2_t \right\} \right] &= \E_{\bnu}\left[ \max \left\{ \sum_{t=1}^{T} U^1_t - \sum_{t=n^1_T+1}^{T} U^1_t,  \sum_{t=1}^{T} U^2_t - \sum_{t=n^2_T+1}^{T} U^2_t \right\} \right] \notag\\
	&\overset{(a)}{\leq} \E_{\bnu}\left[ \max \left\{ \sum_{t=1}^{T} U^1_t - \sum_{t=n_T+1}^{T} U^1_t,  \sum_{t=1}^{T} U^2_t - \sum_{t=n_T+1}^{T} U^2_t \right\} \right] \notag\\
	&\leq \E_{\bnu}\left[ \max \left\{ \sum_{t=1}^{T} U^1_t,  \sum_{t=1}^{T} U^2_t \right\} - \min \left\{ \sum_{t=n_T+1}^{T} U^1_t, \sum_{t=n_T+1}^{T} U^2_t \right\} \right] 
\end{align}
We let $n_T = \max\{n^1_T,  n^2_T\}$ in obtaining inequality (a). Next, we upper bound 
\[ 
 \E_{\bnu}\left[ \max \left\{ \sum_{t=1}^{T} U^1_t,  \sum_{t=1}^{T} U^2_t \right\}\right],
\]
and lower bound
\[
 \E_{\bnu}\left[\min \left\{ \sum_{t=n_T+1}^{T} U^1_t, \sum_{t=n_T+1}^{T} U^2_t \right\} \right].
\]
Consider the ``good'' event 
\[G=\left\{ |\sum_{t=1}^{T}U^j_t - \mu_jT| \leq \sqrt{T\log (T)} \textrm{ for all }j \in [2] \right\}.\]
Since $U^j_t\in[0,1]$, by Hoeffding's inequality, we have that for any $T'\leq T$, 
\begin{align*}
	\p_{\bnu}\left( G\right)&\geq 1-2\exp(-\frac{2(\sqrt{T\log (T)})^2}{T})= 1-\frac{2}{T^2}.
\end{align*}
Therefore,
\begin{align}
 \E_{\bnu}\left[ \max \left\{ \sum_{t=1}^{T} U^1_t,  \sum_{t=1}^{T} U^2_t \right\}\right] &\leq  \E_{\bnu}\left[ \max \left\{ T\mu_1 + \sqrt{T\log (T)},  T\mu_2 + \sqrt{T\log (T)} \right\}\right] + T\p_\nu(G^c)\\
 & \leq T\mu_1 + \sqrt{T\log (T)} +\textup{o}(1).
\end{align}
Also,
\begin{align}
	\E_{\bnu}\left[\min \left\{ \sum_{t=n_T+1}^{T} U^1_t, \sum_{t=n_T+1}^{T} U^2_t \right\} \right] 
	&= \E_{\bnu}\left[ \E_{\bnu}\left[\min \left\{ \sum_{t=n_T+1}^{T} U^1_t, \sum_{t=n_T+1}^{T} U^2_t \right\} \mid n_T \right]  \right]\notag\\
	& \overset{(a)}{\geq} \E_{\bnu}\left[ \min \left\{ (T-n_T)\mu_1 - \sqrt{T\log (T)}, (T-n_T)\mu_2 - \sqrt{T\log (T)} \right\}  \right]\notag\\
	&\geq  (T-\E_{\bnu}\left[n_T\right])\mu_2 - \sqrt{T\log (T)}.\notag
\end{align}
Here for (a), we again condition on the good event that $\sum_{t=n_T+1}^{T} U^i_t \geq \mu_i(T-n_T) - \sqrt{T\log T}$ for $i=1,2$, and, lower bound the probability of this event by $1-1/T^2$ using Hoeffding's inequality. This latter step is valid since, conditioned on $n_T$, $U^i_t$ for $t>n_T$ are i.i.d. with mean $\mu_i$ for $i=1,2$. 
Combining everything, we obtain
\begin{align}
	\E_{\bnu}\left[ \max \left\{ \sum_{t=1}^{n^1_T} U^1_t, \sum_{t=1}^{n^2_T} U^2_t \right\} \right] 
	&\leq  T\Delta + \E_{\bnu}\left[n_T\right]\mu_2 + \textup{O}(\sqrt{T\log (T)}).
\end{align}
	\hfill\Halmos
	
\begin{proof}{Proof of Theorem~\ref{thm:sumbad}.}
For a fixed $T$ large enough so that the sum-regret bound is valid, and for a fixed $\alpha \in (1/2,1)$, consider two two-armed bandit instances, $\bnu$ with means $(1/2, 1/2+1/T^{\alpha/2})$ and $\bnu'$ with means $(1/2, 1/2 - 1/T^{\alpha/2})$. Consider a time $T' = \lfloor T^{\alpha}\rfloor$. Let $T_i$ be the number of times that arm $i$ is pulled until time $T$ and $T'_i$ be the number of times it is pulled until time $T'$ by the policy $\pi$. Let $\p_i$ denote the probabilities of events under policy $\pi$ and instance $i$. Then, by the  Bretagnolle-Huber inequality, we have that 
$$\p_{\bnu}(T'_1>T'_2) + \p_{\bnu'}(T'_2>T'_1) = \overline \p_{\bnu}(T'_1>T'_2) +  \overline\p_{\bnu'}(T'_2>T'_1)
 \geq  \frac{1}{2}\exp\left( - \text{D}\left( \overline{\p}_{\bnu}, \overline{\p}_{\bnu'} \right) \right),$$
where  $\overline\p_{\bnu} ()$ ($\overline{\p}_{\bnu'}()$) denotes the probability distribution on events until time $T'$ under policy $\pi$ and bandit instance $\bnu$ ($\bnu'$). We can upper bound the relative entropy $ \text{D}\left( \overline{\p}_{\bnu}, \overline{\p}_{\bnu'} \right)$ as,
	\begin{align}
		\text{D}\left( \overline{\p}_{\bnu}, \overline{\p}_{\bnu'} \right) = T' \text{D}\left( \nu_{2}, \nu'_{2} \right) \leq T^\alpha \text{D}\left( \nu_{2}, \nu'_{2} \right) \leq \textup{O}(1),
	\end{align}
where the last inequality results from the fact that $\text{D}\left( \nu_{2}, \nu'_{2} \right)$ is the KL-divergence between two Bernoulli random variables with means $1/2 - 1/T^{\alpha/2}$ and $1/2 +1/T^{\alpha/2}$, which is $\textup{O}(1/T^{\alpha})$.
Thus, we have that
$$\p_{\bnu}(T'_1>T'_2) + \p_{\bnu'}(T'_2>T'_1) \geq C_1,$$
for some constant $C_1>0$. This means that there is an instance between the two for which the probability that the suboptimal arm is pulled more until time $T'$ is at least a constant $C_1/2 = C_2$. Let's suppose that instance is $\bnu$ without loss of generality, i.e., we have that,  
$$ \p_{\bnu}(T'_1>T'_2)\geq C_2.$$
Additionally, since, 
$$C\sqrt{T}(\log T)^g\geq \textup{SumRegret} \geq \p_{\bnu}(T_2<T/2)\times T/2\times 1/T^{\alpha/2},$$
We have that 
$$\p_{\bnu}(T_2<T/2)\leq 2CT^{-1/2 +\alpha/2}(\log T)^g= \textup{o}(1)$$
since $\alpha<1$. Thus, we have that 
$$\p_{\bnu}(\{T'_1>T'_2\} \cap \{T_2>T/2\}) \geq C_2 - \textup{o}(1).$$
But  $T'_1>T'_2$ and  $T_2>T/2$ implies that $\min(T_1,T_2)\geq T'/2$. Thus, we have that,
$$\textup{E}_{\bnu}(\min(T_1,T_2)) \geq  \p_{\bnu}(\min(T_1,T_2)\geq T'/2) \times T'/2 \geq  C_2T'/2- \textup{o}(T') = \Omega(T').$$
This means that $\textup{E}_{\pi,\bnu}(\max(T_1,T_2)) =  T - \textup{E}_{\pi,\bnu}(\min(T_1,T_2))\leq T-\Omega(T^\alpha)$. Thus the expected reward under the max objective for instance $\bnu$ under policy $\pi$ is at most
\begin{align}
&T/T^{\alpha/2}  + 1/2(T-\Omega(T^\alpha)) + \textup{O}(\sqrt{T\log(T)})\\
&~~=T(\frac{1}{2}+\frac{1}{T^{\alpha/2}}) - \Omega(T^\alpha) + \textup{O}(\sqrt{T\log(T)})\\
&~~= T(\frac{1}{2}+\frac{1}{T^{\alpha/2}}) - \Omega(T^\alpha)
\end{align}
for any $\alpha>1/2$.
Thus the max regret under this policy is at least $ \Omega(T^\alpha)$.
\hfill\Halmos
\end{proof}

\subsection{Proof of Theorem \ref{thm_general:lb}} \label{apx_general:thm_lb}
	First we fix a policy $\pi \in \Pi$. Let $\Delta \triangleqq (K-m)^{1/3}/(2m^{1/3}T^{1/3})$. We construct two bandit environments with different reward distributions for each of the arms and show that $\pi$ cannot perform well in both environments simultaneously.
	
	We first specify the reward distribution for the arms in the base environment, denoted as the bandit $\boldsymbol{\nu} = \left( \nu_1, \dots, \nu_K \right)$. Assume that the reward for all of the arms have the Bernoulli distribution, i.e., $\nu_i \sim \textup{Bernoulli}(\mu_i)$.  We let $\mu_1 = \mu_2 = \dots = \mu_m = \frac{1}{2}+\Delta$, and $\mu_i = \frac{1}{2}$ for $m + 1 \leq i \leq K$. We let $\p_{\boldsymbol{\nu}}$ denote the probability distribution induced over events until time $\frac{T}{m}$ under policy $\pi$ in this first environment, i.e., in bandit $\bnu$. Let $\E_{\boldsymbol{\nu}}$ denote the expectation under $\p_{\boldsymbol{\nu}}$.
	
	Let $\tau = \lceil\frac{\Delta T}{m}\rceil$ and define $n^i_\tau$ as the (random) number of pulls spent on arm $i \in \{1,\dots,K\}$ until period $\tau$ (note that $\sum_{i = 1}^K n^i_\tau = m\tau$ until period $\tau$) under policy $\pi$. 
	Also, under policy $\pi$, let $l^*$ denote the set of $m$ arms in the set $[K]\setminus \{1, \dots, m\}$ that is pulled the least in expectation until period $\lceil\frac{\Delta T}{m}\rceil$. Then, we must have that $\E_{\bnu}(\sum_{i \in l^*} n^i_\tau)\leq \frac{\lceil\frac{\Delta T}{m}\rceil m^2}{K-m} \approxeq \frac{\Delta T m}{K-m}$. 
	
	Having defined $l^*$, we can now define the second environment, denoted as the bandit $\boldsymbol{\nu'} = \left\{ \nu'_1, \dots, \nu'_K \right\}$. Without loss of generality, for ease of notation, we can let $l^*$ to be the last $m$ arms, i.e., $l^* = \{k_m, \dots, K \}$, where $k_m = K-m+1$. 
	Again, assume that the reward for all of the arms have the Bernoulli distribution, i.e., $\nu'_i \sim \textup{Bernoulli}(\mu'_i)$. We let $\mu'_1 = \mu'_2 = \dots = \mu'_m = \frac{1}{2}+\Delta$, $\mu'_i = \frac{1}{2}$ for $m + 1 \leq i \leq k_m-1$, and $\mu'_{k_m} = \mu'_{k_m+1} = \dots = \mu'_K = \frac{1}{2}+2\Delta$. We let $\p_{\bnu'}$ denote the probability distribution induced over events until time $\frac{T}{m}$ under policy $\pi$ in this second environment, i.e., in bandit $\bnu'$. Let $\E_{\bnu'}$ denote the expectation under $\p_{\bnu'}$.
	
	With some abuse of notation, for any event $B$, we define:
	\begin{align}
		\PSR_T(\pi, \bnu, B) = \bestm \frac{T}{m} \p_{\bnu}(B)  -\E_{\bnu}\big(\obj\big(\overline{U}^1_{\Tminline}, \overline{U}^2_{\Tminline}, \dots, \overline{U}^K_{\Tminline}\big)\mathbbm{1}_{B}\big).
	\end{align}
	It is then clear that $\PSR_T(\pi, \bnu) = \PSR_T(\pi, \bnu, B)+ \PSR_T(\pi, \bnu, B^c).$ 
	
	We define event $A =\{\sum_{t=1}^\tau \mathbbm{1}_{ \{ \boldi_t \cap \{1,\dots,m\} > \lceil\frac{m}{2}\rceil \} } < \frac{\tau}{2}\}$. In words, $A$ is the event where, until time $\tau$, there are at least $\tau/2$ periods in which at most half of the first $m$ arms are pulled. Then, building on the event $A$, we need the following two results for our proof.
	
	\begin{lemma}\label{lma_general:1}
		Fix a policy $\pi$. Consider the K-armed bandit instance $\bnu$ with Bernoulli rewards and mean vector $\bmu = (\mu_1 ,\mu_2, \dots, \mu_K)$ with $\mu_1 = \mu_2 = \dots = \mu_m = \frac{1}{2} +\Delta$ and $\mu_{m+1} = \dots = \mu_K = \frac{1}{2}$, where $\Delta <\frac{1}{2}$. Consider the event $ A =\{\sum_{t=1}^\tau \mathbbm{1}_{ \{ \boldi_t \cap \{1,\dots,m\} > \lceil\frac{m}{2}\rceil \} } < \frac{\tau}{2}\}$. Then we have,
		\[\PSR_T(\pi, \bnu, A) \geq\frac{\Delta T}{4m}\p_{\bnu}(A)-2\sqrt{ \frac{T}{m} \log\left(\frac{KT}{m}\right) }-2.\]
	\end{lemma}
	
	The proof of Lemma~\ref{lma_general:1} is presented below in this section. A similar argument shows the following.
	\begin{lemma}\label{lma_general:2}
		Fix a policy $\pi$. Consider the K-armed bandit instance $\bnu'$ with Bernoulli rewards and mean vector $\bmu' = (\mu'_1 ,\mu'_2, \dots, \mu'_K)$ with $\mu'_1 = \mu'_2 = \dots = \mu'_m = \frac{1}{2} +\Delta$, $\mu'_{m+1} = \dots = \mu'_{k_m - 1} = \frac{1}{2}$ and $\mu'_{k_m} = \dots = \mu'_K = \frac{1}{2} + 2\Delta$, where $\Delta <\frac{1}{4}$. Consider the event $A^c = \{\sum_{t=1}^\tau \mathbbm{1}_{ \{ \boldi_t \cap \{1,\dots,m\} > \lceil\frac{m}{2}\rceil \} } \geq \frac{\tau}{2}\}$. Then we have,
		\[\PSR_T(\pi, \bnu', A^c) \geq\frac{\Delta T}{4m}\p_{\bnu'}(A^c)-2\sqrt{ \frac{T}{m} \log\left(\frac{KT}{m}\right) }-2.\]
	\end{lemma}
	The proof of Lemma~\ref{lma_general:2} is omitted since it is almost identical to that of Lemma~\ref{lma_general:1}. 
	
	These two facts result in the following two inequalities: 
	\begin{align}
		\PSR_T(\pi, \bnu, A) &\geq \p_{\bnu}\left( \sum_{t=1}^\tau \mathbbm{1}_{ \{ \boldi_t \cap \{1,\dots,m\} > \lceil\frac{m}{2}\rceil\} } < \frac{\tau}{2} \right) \Omega\left(\frac{\Delta T}{m}\right), \textrm{ and }\\
		\PSR_T(\pi, \bnu', A^c) &\geq \p_{\bnu'}\left( \sum_{t=1}^\tau \mathbbm{1}_{ \{ \boldi_t \cap \{1,\dots,m\} > \lceil\frac{m}{2}\rceil \} } \geq \frac{\tau}{2} \right) \Omega\left(\frac{\Delta T}{m}\right).
	\end{align}
	
	As above, we let $A = \{ \sum_{t=1}^\tau \mathbbm{1}_{ \{ \boldi_t \cap \{1,\dots,m\} > \lceil\frac{m}{2}\rceil \} } < \frac{\tau}{2}  \}$. Now, using the Bretagnolle-Huber inequality (see Theorem 14.2 in \cite{lattimore2018bandit}), we have,
	\begin{align}
		\PSR_T(\pi, \bnu, A) + \PSR_T(\pi, \bnu', A^c) & \geq \Omega\left(\frac{\Delta T}{m}\right) \left( \p_{\bnu}\left( A \right) + \p_{\bnu'}\left( A^c \right) \right)\\
		&= \Omega\left(\frac{\Delta T}{m}\right) \left( \overline{\p}_{\bnu}\left( A \right) + \overline{\p}_{\bnu'}\left( A^c \right) \right)\\
		&\geq \Omega\left(\frac{\Delta T}{m}\right) \exp\left( - \text{D}\left( \overline{\p}_{\bnu}, \overline{\p}_{\bnu'} \right) \right).
	\end{align}
	Here, $\overline{\p}_{\bnu}$ ($\overline{\p}_{\bnu'}$) is the probability distribution induced by the policy $\pi$ on events until time $\lceil\frac{\Delta T}{m}\rceil$ under bandit $\bnu$ ($\bnu'$). The equality then results from the fact that the two events $\{ \sum_{t=1}^\tau \mathbbm{1}_{ \{ \boldi_t \cap \{1,\dots,m\} > \lceil\frac{m}{2}\rceil \} } < \frac{\tau}{2} \}$ and $\{ \sum_{t=1}^\tau \mathbbm{1}_{ \{ \boldi_t \cap \{1,\dots,m\} > \lceil\frac{m}{2}\rceil \} } \geq \frac{\tau}{2} \}$ depend only on the play until time $\lceil\frac{\Delta T}{m}\rceil$. In the second inequality, which results from the Bretagnolle-Huber inequality, $\text{D}\left( \overline{\p}_{\bnu}, \overline{\p}_{\bnu'} \right)$ is the relative entropy, or the Kullback-Leibler (KL) divergence between the distributions $\overline{\p}_{\bnu}$ and $\overline{\p}_{\bnu'}$ respectively.  
	We can upper bound $ \text{D}\left( \overline{\p}_{\bnu}, \overline{\p}_{\bnu'} \right)$ as,
	\begin{align*}
		\text{D}\left( \overline{\p}_{\bnu}, \overline{\p}_{\bnu'} \right) = \sum_{i = k_m}^K \E_{\bnu}(n^i_\tau) \text{D}\left( \nu_i,\nu'_i \right)  &= \text{D}\left( \nu_K,\nu'_K \right) \sum_{i = k_m}^K \E_{\bnu}(n^i_\tau)\\  &~\leq \text{D}\left( \nu_K,\nu'_K\right) \frac{\lceil\frac{\Delta T}{m}\rceil m^2}{K-m} \lessapprox \frac{8\Delta^3 Tm}{K-m},
	\end{align*}
	where $\nu_i$ ($ \nu'_i $) denotes the reward distribution of arm $i$ in the first (second) environment. 
	The first equality results from the fact that only the last $m$ arms differ between $\bnu$ and $\bnu'$. 
	The second equality follows since the reward distribution of the last $m$ arms are identical. The first inequality follows from the fact that $\sum_{i = k_m}^K \E_{\bnu}(n^i_\tau) \leq \frac{\lceil\frac{\Delta T}{m}\rceil m^2}{K-m}$. Now, $\text{D}\left( \nu_K,\nu'_K\right) $ is simply the relative entropy between the distributions $\textup{Bernoulli}(1/2)$ and $\textup{Bernoulli}(1/2+2\Delta)$, which, by elementary calculations, can be shown to be at most $8\Delta^2$, resulting in the final inequality.
	
	Thus, we finally have,
	\[
	\PSR_T(\pi, \bnu, A) + \PSR_T(\pi, \bnu', A^c)  \geq  \Omega\left(\frac{\Delta T}{m}\right)  \text{exp}\left( - \frac{8\Delta^3 Tm}{K-m} \right).
	\]
	
	Substituting $\Delta = (K-m)^{1/3}/(2m^{1/3}T^{1/3})$ gives
	\begin{align}
		\PSR_T(\pi, \bnu, A) + \PSR_T(\pi, \bnu', A^c)  \geq \Omega\left( \frac{(K-m)^{1/3} T^{2/3}}{m^{4/3}} \right). \label{eqn_general:p1}
	\end{align}
	
	Equation~\ref{eqn_general:p1} along with
	\begin{align}
		\PSR_T(\pi, \bnu, A^c) \geq - \tilde{\textup{O}} \left( \sqrt{\Tminline} \right)&\text{ and }\label{eqn_general:regretlb_1}\\
		\PSR_T(\pi, \bnu', A) \geq - \tilde{\textup{O}} \left( \sqrt{\Tminline} \right)&,\label{eqn_general:regretlb_2}  
	\end{align}
	imply that
	\begin{align}
		\PSR_T(\pi, \bnu) + \PSR_T(\pi, \bnu')  \geq \Omega\left( \frac{(K-m)^{1/3} T^{2/3}}{m^{4/3}} \right).
	\end{align}
	
	Finally, using $2\max\{a,b\} \geq a +b$ gives the desired lower bound on the regret.

	Showing Equations~\ref{eqn_general:regretlb_1} and~\ref{eqn_general:regretlb_2} is an easy exercise:
	\begin{align}
		\PSR_T(\pi, \bnu, A^c) &= \bestm \frac{T}{m} \p_{\bnu}(A^c)  -\E_{\bnu}\big(\obj\big(\overline{U}^1_{\Tminline}, \dots, \overline{U}^K_{\Tminline}\big)\mathbbm{1}_{A^c}\big)\nonumber\\
		&\geq \bestm \frac{T}{m} \p_{\bnu}(A^c)  -\E_{\bnu}\big(\obj\big(\sum_{t=1}^{\Tminline}U^1_t, \dots, \sum_{t=1}^{\Tminline}U^K_t\big)\mathbbm{1}_{A^c}\big)\nonumber\\
		&\overset{(a)}{\geq} \bestm \frac{T}{m}\p_{\bnu}(A^c)  - \bestm \frac{T}{m}\p_{\bnu}(A^c)-2\sqrt{ \frac{T}{m} \log\left(\frac{KT}{m}\right) }-2\nonumber\\
		&= -2\sqrt{ \frac{T}{m} \log\left(\frac{KT}{m}\right) }-2.\label{eqn_general:part2a}
	\end{align}
	Here (a) follows from an argument essentially identical to the one in the proof of Lemma~\ref{lma_general:1} below and we do not repeat it here for brevity. Similarly, we can show that 
	\begin{align}
		\PSR_T(\pi, \bnu', A)\geq -2\sqrt{ \frac{T}{m} \log\left(\frac{KT}{m}\right) }-2.\label{eqn_general:part2b}
	\end{align}
	\hfill\Halmos
	
	\proof{Proof of Lemma~\ref{lma_general:1}.}
	We first have that
	\begin{align}
		\E_{\bnu}\big(\obj\big(\overline{U}^1_{\Tminline}, \dots, \overline{U}^K_{\Tminline}\big)\mathbbm{1}_{A}\big)  &= \E_{\bnu}\big(\obj\big(\sum_{t=1}^{n^1_{\Tminline}} U^1_t, \dots, \sum_{t=1}^{n^K_{\Tminline}} U^K_t\big)\mathbbm{1}_{A}\big)\\
		&\leq \E_{\bnu}\big(\obj\big(\sum_{t=1}^{ T_1} U^1_t, \dots, \sum_{t=1}^{ T_m} U^m_t, \sum_{t=1}^{\Tminline} U^{m+1}_t, \dots \sum_{t=1}^{\Tminline} U^K_t\big)\mathbbm{1}_{A}\big),\label{eqn_general:here1}
	\end{align}
	where $T_i = \frac{T}{m} - t_i$. The event $A$ states that, until time $\tau$, there are at least $\tau/2$ periods in which at most half of the first $m$ arms are pulled, so we have that $\sum_{i = 1}^m t_i \geq \frac{m\tau}{2}$ and $t_i \leq \tau$ for any $i \in [m]$.
	
	Consider the event 
	\[G=\left\{ |\sum_{t=1}^{T_j}U^j_t - \mu_j T_j| \leq \sqrt{ \frac{T}{m} \log\left(\frac{KT}{m}\right) } \textrm{ for all } j \right\},\]
	for $T_j \leq T/m$ for all $j \in [K]$.
	Since $U^j_t\in[0,1]$, by Hoeffding's inequality, we have that for any $T'\leq T/m$, 
	\begin{align*}
		\p_{\bnu}\left( |\sum_{t=1}^{T'}U^j_t - \mu_jT'| \leq \sqrt{ \frac{T}{m} \log\left(\frac{KT}{m}\right) } \right) & \geq 1 - 2\exp( -\frac{2 \left(\sqrt{ \frac{T}{m} \log\left(\frac{KT}{m}\right) }\right)^2}{T'})\\
		&\geq 1 - 2\exp( -\frac{2 \left(\sqrt{ \frac{T}{m} \log\left(\frac{KT}{m}\right) }\right)^2}{T/m})\\
		&= 1-\frac{2m^2}{K^2T^2}\geq 1-\frac{2m}{KT}. 
	\end{align*}
	Hence, by the union bound we have that $P\left( G \right)\geq 1-\frac{2m}{T}$. 
	
	Recall that, for $i \in [m]$, we have $T_i = \frac{T}{m} - t_i$ with $\sum_{i = 1}^m t_i \geq \frac{m\tau}{2}$ and $t_i \leq \tau$ for any $i \in [m]$. Thus we finally have,
	\begin{align}
		&\E_{\bnu}\big(\obj\big(\sum_{t=1}^{T_1} U^1_t, \dots, \sum_{t=1}^{ T_m} U^m_t, \sum_{t=1}^{\Tminline} U^{m+1}_t, \dots \sum_{t=1}^{\Tminline} U^K_t\big)\mathbbm{1}_{A}\big)\nonumber\\
		&~~= \E_{\bnu}\big(\obj\big(\sum_{t=1}^{ T_1} U^1_t,  \dots, \sum_{t=1}^{ T_m} U^m_t, \sum_{t=1}^{\Tminline} U^{m+1}_t, \dots \sum_{t=1}^{\Tminline} U^K_t\big)\mathbbm{1}_{A,\,G}\big)\nonumber\\
		&~~~~+\E_{\bnu}\big(\obj\big(\sum_{t=1}^{ T_1} U^1_t,  \dots, \sum_{t=1}^{ T_m} U^m_t, \sum_{t=1}^{\Tminline} U^{m+1}_t, \dots \sum_{t=1}^{\Tminline} U^K_t\big)\mathbbm{1}_{A}\mid G^c\big)\p_{\bnu}(G^c)\nonumber\\
		&~~\leq \E_{\bnu}\left(\left( \obj\left( \mu_1 T_1, \dots, \mu_m T_m, \mu_{m+1} \frac{T}{m}, \dots, \mu_K \frac{T}{m} \right) +  2\sqrt{ \frac{T}{m} \log\left(\frac{KT}{m}\right) }\right)\mathbbm{1}_{A,\,G}\right)+ \frac{2m}{T}\times \frac{T}{m}\nonumber\\
		&~~\overset{(a)}{\leq} \max\left((\frac{1}{2}+\Delta)(\frac{T}{m} - \lceil\frac{\tau}{2}\rceil), (\frac{1+\Delta}{2})\frac{T}{m}\right) \p_{\bnu}(A)+  2\sqrt{ \frac{T}{m} \log\left(\frac{KT}{m}\right) } + 2\nonumber\\
		&~~\overset{(b)}{=}(\frac{1}{2}+\Delta)(\frac{T}{m} - \lceil\frac{\tau}{2}\rceil)\p_{\bnu}(A) +  2\sqrt{ \frac{T}{m} \log\left(\frac{KT}{m}\right) } + 2.\label{eqn_general:here2}
	\end{align}
	Here (a) follows from the observation of two extreme cases: (i) losing $\lceil \tau/2 \rceil$ pulls from each of the first $m$ arms or (ii) pulling $\lfloor m/2 \rfloor$ of the arms in $\{1,\dots,m\}$ and $\lceil m/2 \rceil$ of the arms in $\{m+1,\dots,K\}$ for all $\frac{T}{m}$ periods. (b) follows from the fact that $\Delta<\frac{1}{2}$. Thus, from Equations~\ref{eqn_general:here1} and~\ref{eqn_general:here2}, we finally have,
	\begin{align}
		\PSR_T(\pi, \bnu, A) &= (\frac{1}{2} +\Delta) \frac{T}{m} \p_{\bnu}(A) -\E_{\bnu}\big(\obj\big(\sum_{t=1}^{ T_1} U^1_t, \dots, \sum_{t=1}^{ T_m} U^m_t, \sum_{t=1}^{\Tminline} U^{m+1}_t, \dots \sum_{t=1}^{\Tminline} U^K_t\big)\mathbbm{1}_{A}\big)\nonumber\\
		&\geq (\frac{1}{2} +\Delta) \frac{T}{m} \p_{\bnu}(A) - (\frac{1}{2}+\Delta)(\frac{T}{m} - \lceil\frac{\tau}{2}\rceil)\p_{\bnu}(A) -  2\sqrt{ \frac{T}{m} \log\left(\frac{KT}{m}\right) } - 2\nonumber\\
		&= (\frac{1}{2} +\Delta) \lceil\frac{\Delta T}{2m}\rceil \p_{\bnu}(A) -  2\sqrt{ \frac{T}{m} \log\left(\frac{KT}{m}\right) } - 2\nonumber\\
		&\geq \frac{\Delta T}{4m}\p_{\bnu}(A) -  2\sqrt{ \frac{T}{m} \log\left(\frac{KT}{m}\right) } - 2.
	\end{align}
	\hfill\Halmos
	\endproof

	\newpage
	\section{Proofs of upper bounds}\label{apx:upper}
	
	\subsection{Proof of Theorem \ref{thm_stopping}}\label{apx:thm2}
	The proof of Theorem \ref{thm_stopping} utilizes two technical lemmas. The first one is the following. 
	\begin{lemma}\label{lma:peeling}
		Let $\delta\in(0,1)$, and $X_1$, $X_2$, $\dots$, be a sequence of independent $0$-mean 1-Sub-Gaussian random variables. Let $\bar{\mu}_t =\frac{1}{t}\sum_{s=1}^tX_s$. Then for any $x>0$, 
		\[P\left(\exists\, t>0: \bar{\mu}_t +\sqrt{\frac{4}{t}\log^+\left( \frac{1}{\delta t^{3/2}} \right)} +x <0\right)\leq \frac{39\delta}{x^3}.\]
	\end{lemma}
	
	Its proof is similar to the proof of Lemma 9.3 in \cite{lattimore2018bandit}, which we present below.
	
\proof{Proof of Lemma~\ref{lma:peeling}.}
	We have,
	\begin{align}
		&P\left(\exists\, t>0: \bar{\mu}_t +\sqrt{\frac{4}{t}\log^+\left( \frac{1}{\delta t^{3/2}} \right)} +x <0\right) \nonumber\\
		&~~= P\left(\exists\, t>0: t\bar{\mu}_t +\sqrt{4t\log^+\left( \frac{1}{\delta t^{3/2}} \right)} +tx <0\right)\nonumber\\
		&~~\leq \sum_{i = 0}^\infty P\left(\exists\, t \in \left[ 2^i, 2^{i+1} \right]: t\bar{\mu}_t +\sqrt{4t\log^+\left( \frac{1}{\delta t^{3/2}} \right)} +tx <0\right)\nonumber\\
		&~~\leq \sum_{i = 0}^\infty P\left(\exists\, t \in \left[ 0, 2^{i+1} \right]: t\bar{\mu}_t +\sqrt{ 2^{i+2} \log^+\left( \frac{1}{\delta 2^{(i+1) \cdot 3/2}} \right)} + 2^ix <0\right)\nonumber\\
		&~~\leq \sum_{i = 0}^\infty \exp\left(-\frac{\left( \sqrt{ 2^{i+2} \log^+\left( \frac{1}{\delta 2^{(i+1) \cdot 3/2}} \right)} + 2^ix \right)^2}{2^{i+2}}\right)\nonumber\\
		&~~\leq \delta \sum_{i = 0}^\infty 2^{(i+1) \cdot 3/2} \exp\left( - 2^{i-2}x^2 \right), \label{eq:unimod}
	\end{align}
	where the first inequality follows from a union bound on a geometric grid. The second inequality is used to set up the argument to apply Theorem 9.2 in \cite{lattimore2018bandit} and the third inequality is due to its application. The fourth inequality follows from $(a+b)^2 \geq a^2+b^2$ for $a,b\geq 0$. Then, using a property of unimodal functions ($\sum_{j = c}^d f(j) \leq \max_{i \in [c,d]} f(i) + \int_c^d f(i)di$ for a unimodal function $f$), the Expression~\ref{eq:unimod} can be upper bounded by $\frac{42\delta}{e^{3/2} x^3} + \delta \int_0^\infty (2^{3/2})^{i+1} \exp(-x^2 2^{i-2} )di$. Evaluating the integral to $\frac{8\sqrt{2\pi}}{\log(2)}\frac{1}{x^3}$, we get 
	\begin{align}
		P\left(\exists\, t>0: \bar{\mu}_t +\sqrt{\frac{4}{t}\log\frac{1}{\delta t^{3/2}}} +x <0\right)\leq \frac{39\delta}{x^3}.
	\end{align}
	\hfill\Halmos\\
	\endproof
	
	The second result we need is Lemma 8.2 from \cite{lattimore2018bandit}, which we present below for completeness.
	\begin{lemma}\cite{lattimore2018bandit}\label{lma:bound2}
		Let $X_1$, $X_2$, $\cdots$, be a sequence of independent $0$-mean 1-Sub-Gaussian random variables. Let $\bar{\mu}_t =\frac{1}{t}\sum_{s=1}^tX_s$. Let $\epsilon>0$, and $a>0$, and define
		\[
		\kappa = \sum_{t=1}^T\mathbbm{1}\{\bar{\mu}_t +\sqrt{\frac{2a}{t}}>\epsilon\}.
		\]
		Then $\E[\kappa]\leq 1+\frac{2}{\epsilon^2}(a+\sqrt{a\pi}+1)$.
	\end{lemma}
	
\proof{Proof of Theorem \ref{thm_stopping}.}
	
Let $1$ denote the first arm and $i^*$ denote the arm used in the Commit phase of ADA-ETC. 
We first define a random variable that quantifies the lowest value of the index of arm $1$ can take with respect to its true mean across $\tau$ pulls. 
\[ \Delta \triangleqq \left( \mu_1 - \min_{n \leq \tau} \left(  \bar{\mu}_{n}^1 + \sqrt{\frac{4}{n}\log\left(\frac{T}{Kn^{3/2}}\right)}\mathbbm{1}_{\left\{ n < \tau \right\}} \right) \right)^+.\]
The following bound is instrumental for our analysis. For any $x\geq 0$, 
\begin{align}
	P(\Delta>x) &= P\left(\exists\, n \leq \tau: \bar{\mu}^1_{n} + \sqrt{\frac{4}{n}\log\left(\frac{T}{Kn^{3/2}}\right)} \mathbbm{1}_{\left\{ n < \tau \right\}}< \mu_1 - x\right)\notag\\
	&\leq P\left(\exists\, n < \tau: \bar{\mu}^1_{n} + \sqrt{\frac{4}{n}\log\left(\frac{T}{Kn^{3/2}}\right)}< \mu_1 - x\right) + P\left(\bar{\mu}^1_{\tau} < \mu_1 - x\right)\notag\\
	&\overset{(a)}{\leq} \min(1,\frac{39K}{Tx^3} +\exp(-2\tau x^2))\\
	&\overset{(b)}{\leq} \min(1,\frac{40K}{Tx^3}).\label{eq:tailbound}
\end{align}
Here, (a) follows from Lemma~\ref{lma:peeling} and Hoeffding's inequality, and (b) follows by the definition of $\tau$ and since $\exp(-2\alpha^{2/3})\leq 1/\alpha$ for all $\alpha \geq 0$.

We next decompose the regret into the regret from wasted pulls in the Explore phase and the regret from committing to a suboptimal arm in the Commit phase. Let $\omega$ be the random time when the Explore phase ends. Let $r^i_\omega$ be the reward earned from arm $i$ until time $\omega$. Then the expected regret in the event that $\{i^* = i\}$ is bounded by:\\
\begin{align}
	&\E\left( \Big(T\mu_1 - (T- \sum_{j\neq i} n^j_\omega -n^i_\omega)\mu_i -r^i_\omega\Big)\mathbbm{1}_{\{i^* = i\}}\right). \label{eq:regretdecompose_1}
\end{align}
Note that this expression assumes that the cumulative reward of arm $i$ will be chosen to compete against $T\mu_1$ at the end of time $T$; however, if there is an arm with a higher cumulative reward, then the resulting regret can only be lower. Thus the total expected regret is bounded by:
\begin{align}
	&\sum_{i=1}^K\E\left( \Big(T\mu_1 - (T- \sum_{j\neq i} n^j_\omega -n^i_\omega)\mu_i -r^i_\omega\Big)\mathbbm{1}_{\{i^* = i\}}\right)\notag\\
	&\overset{(a)}{\leq} \sum_{i=1}^K\E(T\Delta_i\mathbbm{1}_{\{i^* = i\}}) + \mu_1\sum_{i=1}^K\E(n^i_\omega\mathbbm{1}_{\{i^* \neq i\}}) +  \sum_{i=1}^K\E\Big((n^i_\omega\mu_i - r^i_\omega)\mathbbm{1}_{\{i^* = i\}}\Big)\notag\\
	&\overset{(b)}{=} \sum_{i=1}^K\E(T\Delta_i\mathbbm{1}_{\{i^* = i\}}) + \mu_1\sum_{i=1}^K\E(n^i_\omega\mathbbm{1}_{\{i^* \neq i\}}) +  \sum_{i=1}^K\E\Big((\tau\mu_i - r^i_\omega)\mathbbm{1}_{\{i^* = i\}}\Big)\notag\\
	&=\sum_{i=1}^K\E(T\Delta_i\mathbbm{1}_{\{i^* = i\}}) + \mu_1\sum_{i=1}^K\E(n^i_\omega\mathbbm{1}_{\{i^* \neq i\}}) +  \sum_{i=1}^K P(i^*=i) (\tau\mu_i - \E(r^i_\omega\mid i^* = i))\notag\\
	&\overset{(c)}{=}\sum_{i=1}^K\E(T\Delta_i\mathbbm{1}_{\{i^* = i\}}) + \mu_1\sum_{i=1}^K\E(n^i_\omega\mathbbm{1}_{\{i^* \neq i\}}) +  \sum_{i=1}^K P(i^*=i) (\tau\mu_i - \sum_{n=1}^{\tau}\E(U^i_n\mid i^* = i))\notag\\
	&\overset{(d)}{\leq}\underbrace{\sum_{i=1}^K\E(T\Delta_i\mathbbm{1}_{\{i^* = i\}})}_{\textup{ Regret from misidentifications in Commit phase }} + \underbrace{\mu_1\sum_{i=1}^K\E(n^i_\omega\mathbbm{1}_{\{i^* \neq i\}}).}_{\textup{ Regret from wasted pulls in the Explore phase }} \label{eq:regretdecompose}
\end{align}
Here, (a) results from rearranging terms, and from the fact that $\mu_i\leq \mu_1$. Both (b) and (c) result from the fact that in the event that $\{i^* = i\}$, $n^i_\omega=\tau$. (d) holds since, by a standard stochastic dominance argument, $\tau\mu_i \leq  \sum_{n=1}^{\tau}\E(U^i_n\mid i^* = i)$.

We bound these two terms one by one. \\

{\bf Regret from Explore.} First, note that an instance-independent bound on the regret from Explore is simply $K\tau =K\lceil\frac{T^{2/3}}{K^{2/3}}\rceil= \textup{O}(K^{1/3}T^{2/3})$, which is the maximum number of pulls possible before ADA-ETC enters the Commit phase. Hence, we now focus on deriving an instance-dependent bound. We have that
\begin{align}
	\E( \sum_{i=1}^K n^i_\omega\mathbbm{1}_{\{i^* \neq i\}} )&\leq  \E( \sum_{i=2}^K n^i_\omega )+ \tau P(i^*\neq 1)\nonumber\\
	& =  \E(\sum_{i\geq 2:\Delta \leq\frac{\Delta_i}{2}}n^i_\omega)+\E(\sum_{i\geq 2:\Delta> \frac{\Delta_i}{2}}n^i_\omega) + \tau P(i^*\neq 1). \label{eq:pulls-ub}
\end{align}

We first bound the first term. Define the random variable \[\eta_i = \sum_{n = 1}^{\tau} \mathbbm{1}\left\{\bar{\mu}^i_{n} +  \sqrt{\frac{4}{n}\log\left(\frac{T}{Kn^{3/2}}\right)}\mathbbm{1}_{\left\{ n < \tau \right\}}  \geq \mu_i + \frac{\Delta_i}{2} \right\}.\] Then in the event that $\Delta \leq\frac{\Delta_i}{2}$, we have that $n^i_\omega\leq \eta_i$. We also have that $n^i_\omega\leq \tau$. And thus in the event that $\Delta \leq\frac{\Delta_i}{2}$, we have $n^i_\omega\leq \min(\eta_i,\tau)$. Hence the first term above is bounded as:

\[\sum_{i=2}^K P(\Delta\leq \frac{\Delta_i}{2})\E(\min(\eta_i,\tau)) \leq \sum_{i=2}^K P(\Delta\leq\frac{\Delta_i}{2})\min(\E(\eta_i),\tau)\leq  \sum_{i=2}^K \min(\E(\eta_i),\tau). \]
We can now bound $\E(\eta_i)$ as follows:
\begin{align}
	\E(\eta_i) &\leq 1+ \E\left( \sum_{n = 1}^{\tau-1} \mathbbm{1}\left\{\bar{\mu}^i_{n} +  \sqrt{\frac{4}{n}\log\left(\frac{T}{Kn^{3/2}}\right)}  \geq \mu_i + \frac{\Delta_i}{2} \right\} \right) \notag\\
	&= 1+ \E\left( \sum_{n = 1}^{\tau-1} \mathbbm{1}\left\{\bar{\mu}^i_{n} +  \sqrt{\frac{4}{n}\log^+\left(\frac{T}{Kn^{3/2}}\right)}  \geq \mu_i + \frac{\Delta_i}{2} \right\} \right)\notag\\
	&\overset{(a)}{\leq} 1+ \frac{1}{\Delta_i^2} + \E\left(\sum_{n = 1}^{\tau-1} \mathbbm{1}\left\{\bar{\mu}^i_{n} +  \sqrt{\frac{4}{n}\log^+\left(\frac{T\Delta_i^3}{K}\right)}  \geq \mu_i + \frac{\Delta_i}{2} \right\} \right)\notag\\
	&\overset{(b)}{\leq} 2+ \frac{1}{\Delta_i^2} + \frac{8}{\Delta_i^2} \left( 2 \log^+\left(\frac{T\Delta_i^3}{K}\right)+ \sqrt{2\pi \log^+\left(\frac{T\Delta_i^3}{K}\right)} + 1 \right)\notag\\
	&\leq  \frac{11}{\Delta_i^2} + \frac{16}{\Delta_i^2} \log^+\left(\frac{T\Delta_i^3}{K}\right) +\frac{24}{\Delta_i^2}\sqrt{\log^+\left(\frac{T\Delta_i^3}{K}\right)} .
\end{align}

Here, (a) is due to lower bounding $1/n^{3/2}$ by $\Delta_i^3$, and adding $1/\Delta_i^2$ for the first $1/\Delta_i^2$ time periods where this lower bound doesn't hold. (b) is due to Lemma~\ref{lma:bound2}. 
The final inequality results from the fact that $\Delta_i\leq 1$ and from trivially bounding $2\pi\leq 9$. 
Thus, we finally have,
\begin{align}
	\E(\sum_{i\geq 2:\Delta \leq \frac{\Delta_i}{2}}n^i_\omega)\leq \sum_{i=2}^K \min\left(\frac{11}{\Delta_i^2} + \frac{16}{\Delta_i^2} \log^+\left(\frac{T\Delta_i^3}{K}\right)+\frac{24}{\Delta_i^2}\sqrt{\log^+\left(\frac{T\Delta_i^3}{K}\right)},\tau\right). \label{ub-explore-1}
\end{align}

We now focus on the second term in Equation~\ref{eq:pulls-ub}. Note that we have $n^i_\omega\leq \tau$, and hence,
\begin{align}
	\E(\sum_{i\geq 2:\Delta > \frac{\Delta_i}{2}}n^i_\omega)\leq \tau\sum_{i=2}^K P(\Delta> \frac{\Delta_i}{2})\leq \tau \sum_{i=2}^K\min(1,\frac{320K}{T\Delta_i^3}). \label{ub-explore-2}
\end{align}
Here the second inequality follows from Equation~\ref{eq:tailbound}. 
Next, we focus on the third term in Equation~\ref{eq:pulls-ub}. We have:
\begin{align}
	P(i^* \neq 1) &= P(i^* \neq 1 \textrm{ and } \Delta \leq \frac{\Delta_2}{2}) + P(i^* \neq 1 \textrm{ and } \Delta > \frac{\Delta_2}{2})\notag\\
	&\leq \min\left(1,\sum_{i=2}^KP(i^* = i \textrm{ and } \Delta \leq \frac{\Delta_2}{2}) + P(\Delta > \frac{\Delta_2}{2})\right)\notag\\
	&\leq \min\left(1,\sum_{i=2}^KP(i^* = i \textrm{ and } \Delta \leq \frac{\Delta_2}{2}) + \frac{320K}{T\Delta_2^3}\right).\label{ub-explore-mid}
\end{align}
Here the final inequality again follows from Equation~\ref{eq:tailbound}. Now in the event that $\Delta\leq \Delta_2/2$, $i^*=i$ implies that there is some $n\leq \tau$ such that $\lcb^i_n=\bar{\mu}^i_n-\bar{\mu}_{n}^{i}\mathbbm{1}_{\{n< \tau\}} > \mu_i+ \Delta_i/2$.
Thus, we have,
\begin{align}
	\sum_{i=2}^KP(i^* = i \textrm{ and } \Delta \leq \frac{\Delta_2}{2})&\leq \sum_{i=2}^KP\left(\exists\, n\leq \tau: \bar{\mu}^i_n-\bar{\mu}_{n}^{i}\mathbbm{1}_{\{n< \tau\}} > \mu_i+\Delta_i/2\right)\nonumber\\
	&= \sum_{i=2}^KP(\bar{\mu}^i_\tau> \mu_i+ \Delta_i/2)\nonumber\\
	&\overset{(a)}{\leq} \sum_{i=2}^K \exp(-\frac{\tau \Delta_i^2}{2})\overset{(b)}{\leq} \sum_{i=2}^K \frac{8K}{T\Delta_i^3}. \label{ub-explore-3.1}
\end{align}
Here, (a) follows from Hoeffding's inequality, and (b) follows from the definition of $\tau$ and the fact that $\exp(-\alpha^{2/3}/2)\leq 8/\alpha$ for $\alpha\geq0$. 
Thus we finally have
\begin{align}
	\tau P(i^* \neq 1)&\leq \tau\min(1,\sum_{i=2}^K \frac{8K}{T\Delta_i^3}+\frac{320K}{T\Delta_2^3})\nonumber\\
	&\leq \tau\min(1,\sum_{i=2}^K \frac{328K}{T\Delta_i^3}). \label{ub-explore-3}
\end{align}

Thus, combining Equations~\ref{ub-explore-1}, \ref{ub-explore-2}, and \ref{ub-explore-3}, we have that the regret from the Explore phase is bounded by 
\begin{align}
	&\mu_1\sum_{i=2}^K \min\left(\frac{11}{\Delta_i^2} + \frac{16}{\Delta_i^2} \log^+\left(\frac{T\Delta_i^3}{K}\right)+\frac{24}{\Delta_i^2}\sqrt{\log^+\left(\frac{T\Delta_i^3}{K}\right)},\tau\right)\notag\\
	&+ \mu_1\tau \sum_{i=2}^K\min(1,\frac{320K}{T\Delta_i^3})+  \mu_1\tau\min(1,\sum_{i=2}^K \frac{328K}{T\Delta_i^3})\notag\\
	&~~\leq \mu_1\sum_{i=2}^K \min\left(\frac{10}{\Delta_i^2} + \frac{16}{\Delta_i^2} \log^+\left(\frac{T\Delta_i^3}{K}\right)+\frac{24}{\Delta_i^2}\sqrt{\log^+\left(\frac{T\Delta_i^3}{K}\right)},\tau\right)\notag\\
	&~~~~+ \mu_1\tau \sum_{i=2}^K\min(2,\frac{648K}{T\Delta_i^3})\label{ub-explore}.
\end{align}
Here the inequality results from the fact that $\min(1,a) + \min(1,b) \leq \min(2,a+b)$ for $a,\, b>0$.
This finishes our derivation of a distribution dependent bound on the regret from the Explore phase. We next focus on the regret arising from misidentification in the Commit phase.\\

{\bf Regret from Commit.} 
This regret is upper bounded by 
\begin{align}
	&\E(\sum_{i:\Delta \leq\frac{\Delta_i}{2}}\mathbbm{1}_{\{i^* = i\}} T\Delta_i)+ \E(\sum_{i:\Delta > \frac{\Delta_i}{2}}\mathbbm{1}_{\{i^* = i\}}T\Delta_i). 
\end{align}
We now get instance dependent and independent bounds on each of the above two terms.

{\bf An instance dependent bound on $\E(\sum_{i:\Delta \leq \frac{\Delta_i}{2}}\mathbbm{1}_{\{i^* = i\}}T\Delta_i)$.}  In the event that $\Delta\leq\Delta_i/2$, $i^*=i$ implies that there is some $n\leq \tau$ such that 
$\lcb^i_n=\bar{\mu}^i_n-\bar{\mu}_{n}^{i}\mathbbm{1}_{\{n< \tau\}} > \mu_i+ \Delta_i/2$.
Thus, we have,
\begin{align}
	\E(\sum_{i:\Delta \leq\frac{\Delta_i}{2}}\mathbbm{1}_{\{i^* = i\}}T\Delta_i)\leq \sum_{i=2}^KP\left(\exists\, n\leq \tau: \bar{\mu}^i_n-\bar{\mu}_{n}^{i}\mathbbm{1}_{\{n< \tau\}} > \mu_i+\Delta_i/2\right)T\Delta_i.
\end{align}
Now, we have,
\begin{align}
	P(\exists\, n\leq \tau: \bar{\mu}^i_n-\bar{\mu}_{n}^{i}\mathbbm{1}_{\{n< \tau\}} > \mu_i+ \Delta_i/2)&= P(\bar{\mu}^i_\tau> \mu_i+ \Delta_i/2)\leq\exp(-\frac{\tau \Delta_i^2}{2}).
\end{align}
Here the final inequality follows from Hoeffding's inequality. Thus we finally have,
\begin{align}
	\E(\sum_{i:\Delta \leq\frac{\Delta_i}{2}}\mathbbm{1}_{\{i^* = i\}}T\Delta_i)\leq \sum_{i=2}^K\exp(-\frac{\tau \Delta_i^2}{2})T\Delta_i.\label{ub-commit-1}
\end{align}

{\bf An instance independent bound on $\E(\sum_{i:\Delta \leq \frac{\Delta_i}{2}}\mathbbm{1}_{\{i^* = i\}}T\Delta_i)$.} 
We have
\begin{align}
	&\E(\sum_{i:\Delta <\frac{\Delta_i}{2}}\mathbbm{1}_{\{i^* = i\}}T\Delta_i)= T^{2/3}K^{1/3}\sqrt{2\log K}+\E(\sum_{i:\Delta <\frac{\Delta_i}{2}; \Delta_i \geq \frac{K^{1/3}\sqrt{2\log K}}{T^{1/3}}}\mathbbm{1}_{\{i^* = i\}}T\Delta_i)\nonumber\\
	&~~\overset{(a)}{\leq} T^{2/3}K^{1/3}\sqrt{2\log K}+ \E(\sum_{i: \Delta_i\geq \frac{K^{1/3}\sqrt{2\log K}}{T^{1/3}}}\exp(-\frac{\tau \Delta_i^2}{2})T\Delta_i)\nonumber\\
	&~~\overset{(b)}{\leq} T^{2/3}K^{1/3}\sqrt{2\log K}+ T^{2/3}K^{1/3}\sqrt{2\log K}.\label{ind-bound-1}
\end{align}
Here, (a) follows for the same reason as the derivation of the bound in Equation~\ref{ub-commit-1}. Next, observe that the function $\exp(-\frac{\tau x^2}{2})x$ is maximized at $x^* = \sqrt{2/\tau}= \sqrt{2}K^{1/3}/T^{1/3}$. But since $\Delta_i\geq \sqrt{2\log K}K^{1/3}/T^{1/3}\geq  \sqrt{2}K^{1/3}/T^{1/3}$, by the unimodality of $\exp(-\frac{\tau x^2}{2})x$,  we have 
\[\exp(-\frac{\tau \Delta_i^2}{2})T\Delta_i\leq \exp(-\log K)T^{2/3}K^{1/3}\sqrt{2\log K}= \frac{1}{K}T^{2/3}K^{1/3}\sqrt{2\log K}.\]
Hence (b) follows.

{\bf An instance dependent bound on $\E(\sum_{i:\Delta > \frac{\Delta_i}{2}}\mathbbm{1}_{\{i^* = i\}}T\Delta_i)$.} 

\begin{align}
	\E(\sum_{i:\Delta > \frac{\Delta_i}{2}}\mathbbm{1}_{\{i^* = i\}}T\Delta_i)&\leq\E(\max_{i\in[K]} T\Delta_i \mathbbm{1}_{\{\Delta > \frac{\Delta_i}{2}\}}) \notag\\ 
	&=P(\Delta > \frac{\Delta_K}{2})T\Delta_K + \sum_{i=1}^{K-1} P( \frac{\Delta_{i+1}}{2}\geq \Delta > \frac{\Delta_i}{2})T\Delta_i\notag\\
	&=P(\Delta > \frac{\Delta_K}{2})T\Delta_K + \sum_{i=1}^{K-1} \left(P(\Delta > \frac{\Delta_i}{2})- P(\Delta > \frac{\Delta_{i+1}}{2})\right)T\Delta_i\notag\\
	&=\sum_{i=2}^KP(\Delta>\frac{\Delta_i}{2})T(\Delta_i-\Delta_{i-1})\notag\\
	&\leq \sum_{i=2}^K\min(1,\frac{320K}{T\Delta_i^3})T(\Delta_i-\Delta_{i-1}).\label{ub-commit-2}
\end{align}
Here the final inequality again follows from Equation~\ref{eq:tailbound}. 

{\bf An instance independent bound on $\E(\sum_{i:\Delta > \frac{\Delta_i}{2}}\mathbbm{1}_{\{i^* = i\}}T\Delta_i)$.} 
We have,
\begin{align}
	\E(\sum_{i:\Delta > \frac{\Delta_i}{2}}\mathbbm{1}_{\{i^* = i\}}T\Delta_i)&\leq  \E(2T\Delta\sum_{i=1}^K\mathbbm{1}_{\{i^* = i\}}) =\E(2T\Delta) = 2T \E(\Delta).
\end{align}
We then look at $\E( \Delta )$. We have,

\[ \E( \Delta ) = \int_{0}^{\infty} \p\left( \Delta > x \right) dx \leq \int_{0}^{\infty} \min \left(1, \frac{40K}{Tx^3}\right)dx.\] 

This integral evaluates to \[ \int_0^{\frac{(40K)^{1/3}}{T^{1/3}}} dx + \int_{\frac{(40K)^{1/3}}{T^{1/3}}}^\infty \frac{40K}{Tx^3} dx \leq 2\frac{(40K)^{1/3}}{T^{1/3}}. \] 

Combining these results, we have 
\begin{align}
	\E( \Delta ) \leq 2\frac{(40K)^{1/3}}{T^{1/3}}. 
\end{align}
Thus we finally have, 

\begin{align}
	\E(\sum_{i:\Delta > \frac{\Delta_i}{2}}\mathbbm{1}_{\{i^* = i\}}T\Delta_i)&\leq 4(40K)^{1/3}T^{2/3}. \label{ind-bound-2} 
\end{align}
The final instance-dependent bound follows from Equations~\ref{ub-explore}, \ref{ub-commit-1}, and \ref{ub-commit-2}. The instance-independent bound follows from the fact that the regret from the Explore phase is at most $K\tau = \textup{O}(T^{2/3}K^{1/3})$ and from Equations~\ref{ind-bound-1} and \ref{ind-bound-2}.
	\hfill\Halmos
	\endproof


	\subsection{Proof of Theorem~\ref{thm_stopping_generalm}}\label{apx_general:thm_ub_gen}
	
	
	Let $\boldi^*$ denote the arms used in the Commit phase of $m\textup{-ADA-ETC}$. 
	We first define $m$ random variables, each quantifying the lowest value of the index of arm $i \in [m]$ can take with respect to its true mean across $\tau$ pulls. Recall that the empirical average reward of arm $i$ remains fixed after $\tau$ pulls.
	\[ \delta_i  \triangleqq \left( \mu_i - \min_{n \leq \tau} \left(  \bar{\mu}_{n}^i + \sqrt{\frac{4}{n}\log\left(\frac{T}{(K-m)n^{3/2}}\right)}\mathbbm{1}_{\left\{ n < \tau \right\}} \right) \right)^+.\] 
	
	
	We also define
	\[ \bar{\delta} \triangleqq \max_{i\leq m} \delta_i. \]
	
	The following bound, which follows from Equation~\ref{eq:tailbound}, is instrumental for our analysis. For any $x\geq 0$ and $i \in [m]$, 
	\begin{align}
		P(\delta_i>x) &= P\left(\exists\, n \leq \tau: \bar{\mu}_{n}^i + \sqrt{\frac{4}{n}\log\left(\frac{T}{(K-m)n^{3/2}}\right)} \mathbbm{1}_{\left\{ n < \tau \right\}}< \mu_i - x\right)\notag\\
		&\leq P\left(\exists\, n < \tau: \bar{\mu}_{n}^i + \sqrt{\frac{4}{n}\log\left(\frac{T}{(K-m)n^{3/2}}\right)} < \mu_i - x\right) + P\left(\bar{\mu}^i_{\tau} < \mu_i - x\right)\notag\\
		&\overset{(a)}{\leq} \min(1,\frac{39(K-m)}{Tx^3} +\exp(-2\tau x^2))\\
		&\overset{(b)}{\leq} \min(1,\frac{40(K-m)}{Tx^3}).\label{eq:tailbound_gen}
	\end{align}
	Again, (a) follows from Lemma~\ref{lma:peeling} and Hoeffding's inequality, and (b) follows by the definition of $\tau$ and since $\exp(-2\alpha^{2/3})\leq 1/\alpha$ for all $\alpha \geq 0$. Notice that the expression in \ref{eq:tailbound_gen} does not depend on arm $i$.
	
	We then decompose the regret into the regret from wasted pulls in the Explore phase and the regret from committing to one or more suboptimal arms in the Commit phase. In contrast to the $m=1$ case, not all arms {\it enter} the exploitation phase at the same time. If there is a time $t$ 
	for arm $i$ where and $n_{t+1}^{i} = \tau + 1$, then arm $i$ belongs to the set of exploited arms from time $(t+1)$ onwards, i.e., $i \in \boldi^*$ (Lemma~\ref{lma:exploit},  presented at the end of this proof).
	
	To that end, we define arm specific stopping times. For $i \in \boldi^*$, let $\omega_i$ be the time period prior to arm $i$ being pulled $(\tau + 1)-\textup{st}$ time, i.e., $\omega_i = \min \left\{ t \leq \frac{T}{m}: n_{t+1}^{i} = \tau + 1 \right\}$. Note that if $i \notin \boldi^*$ we set $\omega_i = \Tminline$. 
	
	For $i \in \boldi^*$, let $r^i_{\omega_i}$ be the reward earned from arm $i$ during its exploration. Define the number of missed pulls from arm $i$ during its exploration as $n^{i,\textup{miss}}_{\omega_i} \triangleqq \omega_i - \tau$, $i \in \boldi^*$. Then, the expected regret in the event that $\{\boldi^* = \boldi\}$ is bounded by:\\
	\begin{align}
		&\E\left( \Big(\frac{T}{m} \bestm - \frac{1}{m} \sum_{i\in \boldi}\Big(\frac{T}{m} -  n^{i,\textup{miss}}_{\omega_i} - n^i_{\omega_i}\Big)\mu_i - \frac{1}{m} \sum_{i\in \boldi}r^i_{\omega_i}\Big)\mathbbm{1}_{\{\boldi^* = \boldi\}}\right). 
	\end{align}
	Note that this expression assumes that the average of the cumulative rewards of the arms in set $\boldi$ will be chosen to compete against $\frac{T}{m}\bestm$ at the end of time $\Tminline$; however, if there are arms with higher cumulative rewards than the arms in $\boldi$, then the resulting regret can only be lower. Thus the total expected regret is bounded by:
	\begin{align}
		&\sum_{\boldi \in {K\choose m}}\E\left( \Big(\frac{T}{m} \bestm - \frac{1}{m} \sum_{i\in \boldi}\Big(\frac{T}{m} -  n^{i,\textup{miss}}_{\omega_i} - n^i_{\omega_i}\Big)\mu_i - \frac{1}{m} \sum_{i\in \boldi}r^i_{\omega_i}\Big)\mathbbm{1}_{\{\boldi^* = \boldi\}}\right)\notag\\
		&\overset{(a)}{=}  \sum_{\boldi \in {K\choose m}}\E\Big(\frac{T}{m}\Delta_\boldi \mathbbm{1}_{\{\boldi^* = \boldi\}}\Big) + \frac{1}{m} \sum_{i=1}^K\E(n^{i,\textup{miss}}_{\omega_i}\mu_i\mathbbm{1}_{\{ i\in \boldi^*\}}) + \frac{1}{m} \sum_{i=1}^K\E\Big((n^i_{\omega_i} \mu_i - r^i_{\omega_i})\mathbbm{1}_{\{i\in \boldi^*\}}\Big)\notag\\
		&\overset{(b)}{=}  \sum_{\boldi \in {K\choose m}}\E\Big(\frac{T}{m}\Delta_\boldi \mathbbm{1}_{\{\boldi^* = \boldi\}}\Big) + \frac{1}{m} \sum_{i=1}^K\E(n^{i,\textup{miss}}_{\omega_i}\mu_i\mathbbm{1}_{\{ i\in \boldi^*\}}) + \frac{1}{m} \sum_{i=1}^K\E\Big((\tau \mu_i - r^i_\tau)\mathbbm{1}_{\{i\in \boldi^*\}}\Big)\notag\\
		&=\sum_{\boldi \in {K\choose m}}\E\Big(\frac{T}{m}\Delta_\boldi \mathbbm{1}_{\{\boldi^* = \boldi\}}\Big) + \frac{1}{m} \sum_{i=1}^K\E(n^{i,\textup{miss}}_{\omega_i}\mu_i\mathbbm{1}_{\{ i\in \boldi^*\}}) + \frac{1}{m}  \sum_{i=1}^K P(i\in \boldi^*) (\tau\mu_i - \sum_{n=1}^{\tau}\E(U^i_n\mid i\in \boldi^*))\notag\\
		&\overset{(c)}{\leq}\underbrace{\sum_{\boldi \in {K\choose m}}\E\Big(\frac{T}{m}\Delta_\boldi \mathbbm{1}_{\{\boldi^* = \boldi\}}\Big)}_{\textup{ Regret from misidentifications in Commit phase }} + \underbrace{\frac{1}{m} \sum_{i=1}^K\E(n^{i,\textup{miss}}_{\omega_i}\mu_i\mathbbm{1}_{\{ i\in \boldi^*\}}).}_{\textup{ Regret from wasted pulls in the Explore phase }} \label{eq:decompose}
	\end{align}
	Here, (a) results from rearranging terms. (b) follows from the fact that in the event of $\{i \in \boldi^*\}$, $n^i_{\omega_i} = \tau$ and $r^i_{\omega_i} = r^i_{\tau}$ by the definition of $\omega_i$. And (c) holds since, by a standard stochastic dominance argument, $\tau\mu_i \leq  \sum_{n=1}^{\tau}\E(U^i_n\mid i \in \boldi^*)$. Here, we let $\Delta_\boldi$ denote $\bestm - \frac{1}{m} \sum_{i \in \boldi} \mu_i$.
	
	We bound these two terms one by one.\\
	
	{\bf Regret from Explore.} First, note that an instance-independent bound on the regret from Explore is simply $\frac{K}{m}\tau - \tau = \frac{K-m}{m}\lceil\frac{T^{2/3}}{\left(K-m\right)^{2/3}}\rceil = \textup{O}(\frac{(K-m)^{1/3}T^{2/3}}{m})$, 
	which is the maximum number of allotted pulls on arms not in set $\boldi^*$ before $m\textup{-ADA-ETC}$ enters the Commit phase. Hence, we now focus on deriving an instance-dependent bound. We have that
	\begin{align}
		\frac{1}{m} \sum_{i=1}^K\E(n^{i,\textup{miss}}_{\omega_i}\mu_i\mathbbm{1}_{\{ i\in \boldi^*\}}) &\leq \frac{1}{m} \sum_{i=1}^m\mu_i\E(n^{i,\textup{miss}}_{\omega_i}) + \frac{1}{m} \sum_{j=m+1}^K\E(n^{j,\textup{miss}}_{\omega_j}\mu_j\mathbbm{1}_{\{ j\in \boldi^*\}})\\
		&\overset{(a)}{\leq} \frac{1}{m}\sum_{i=1}^m\mu_i\E(n^{i,\textup{miss}}_{\omega_i}) + \frac{K-m}{m}\frac{\tau}{m}\sum_{j=m+1}^K \mu_j P( j \in \boldi^*). \label{eq:explore_gen}
	\end{align}
	Here, (a) follows from the fact that the highest number of pulls missed from arm $j \in \boldi^*$ is $\frac{K}{m} \tau - \tau$.

	We first bound the first term in Equation~\ref{eq:explore_gen}. Recall that $\Delta_j = \mu_{m} - \mu_j$ for $j \geq m+1$ and $\bar{\Delta}_i = \mu_i -\mu_{m+1}$ for $i \in [m]$. Then,
	\begin{align}
		\frac{1}{m}\sum_{i=1}^m\mu_i\E(n^{i,\textup{miss}}_{\omega_i}) = \frac{1}{m}\E(\sum_{i \in [m]:\delta_i \leq \bar{\Delta}_i/2} \mu_i n^{i,\textup{miss}}_{\omega_i}) + \frac{1}{m}\E(\sum_{i \in [m]: \delta_i > \bar{\Delta}_i/2} \mu_i n^{i,\textup{miss}}_{\omega_i}). \label{eq:explore_first}
	\end{align}
	
	We now bound the first term in Equation~\ref{eq:explore_first}. Define the random variable
	\begin{align}
		\kappa_j = \sum_{n=1}^{\tau} \mathbbm{1}\left\{ \bar{\mu}_{n}^j + \sqrt{\frac{4}{n}\log\left(\frac{T}{(K-m)n^{3/2}}\right)}\mathbbm{1}_{\left\{ n < \tau \right\}} > \mu_j + \frac{\Delta_j}{2} \right\}. \label{defn:kappa}
	\end{align}
	Then, in the event that $\delta_i \leq \bar{\Delta}_i/2$, we have that $n^{i,\textup{miss}}_{\omega_i} \leq \sum_{j=m+1}^{K} \kappa_j$. 
	From the previous discussion, we also have that $n^{i,\textup{miss}}_{\omega_i} \leq \frac{K-m}{m} \tau$. Hence the first term in Equation~\ref{eq:explore_first} is bounded as:
	\begin{align}
		\frac{1}{m} \sum_{i =1}^{m} \mu_i P\left(\delta_i \leq \bar{\Delta}_i/2\right) \E( \min(\sum_{j=m+1}^{K} \kappa_j ,  \frac{K-m}{m} \tau) ) \leq  \frac{1}{m} \sum_{i =1}^{m} \mu_i   \min\big(\sum_{j=m+1}^{K} \E(\kappa_j),  \frac{K-m}{m} \tau\big).
	\end{align}
	
	We can now bound $\E(\kappa_j)$ as follows:
	\begin{align}
		\E(\kappa_j) &\leq 1 + \E \left( \sum_{n=1}^{\tau-1} \mathbbm{1}\left\{ \bar{\mu}_{n}^j + \sqrt{\frac{4}{n}\log\left(\frac{T}{(K-m)n^{3/2}}\right)} > \mu_j + \frac{\Delta_j}{2}   \right\} \right)\notag\\
		&= 1 + \E \left( \sum_{n=1}^{\tau-1} \mathbbm{1}\left\{ \bar{\mu}_{n}^j + \sqrt{\frac{4}{n}\log^+\left(\frac{T}{(K-m)n^{3/2}}\right)} > \mu_j + \frac{\Delta_j}{2}   \right\} \right)\notag\\
		&\overset{(a)}{\leq}  1 + \frac{1}{\Delta_j^2} + \E \left( \sum_{n=1}^{\tau-1} \mathbbm{1}\left\{ \bar{\mu}_{n}^j + \sqrt{\frac{4}{n}\log^+\left(\frac{T\Delta_j^3}{K-m}\right)} > \mu_j + \frac{\Delta_j}{2}  \right\} \right)\notag\\
		&\overset{(b)}{\leq}  2 + \frac{1}{\Delta_j^2} + \frac{8}{\Delta_j^2} \left(2\log^+\left(\frac{T\Delta_j^3}{K-m}\right) + \sqrt{2\pi\log^+\left(\frac{T\Delta_j^3}{K-m}\right)} +1  \right) \notag\\
		&\leq \frac{11}{\Delta_j^2} + \frac{16}{\Delta_j^2} \log^+\left(\frac{T\Delta_j^3}{K-m}\right) + \frac{24}{\Delta_j^2}\sqrt{\log^+\left(\frac{T\Delta_j^3}{K-m}\right)}.
	\end{align}
	Here, (a) is due to lower bounding $1/n^{3/2}$ by $\Delta_j^3$, and adding $1/\Delta_j^2$ for the first $1/\Delta_j^2$ time periods where this lower bound doesn't hold. (b) is due to Lemma~\ref{lma:bound2} and reorganizing terms. 
	The final inequality results from the fact that $\Delta_j \leq 1$ and from trivially bounding $2\pi\leq 9$. Thus, the first term in Equation~\ref{eq:explore_first} is bounded by
	\begin{align}
		\frac{1}{m}\sum_{i=1}^m\mu_i \min\left( \sum_{j = m+1}^K \min\Big(\frac{11}{\Delta_j^2} + \frac{16}{\Delta_j^2} \log^+\left(\frac{T\Delta_j^3}{K-m}\right) + \frac{24}{\Delta_j^2}\sqrt{\log^+\left(\frac{T\Delta_j^3}{K-m}\right)}, \tau\Big) , ~\frac{K-m}{m}\tau \right). \label{eq:explore_first_p1}
	\end{align}		
	Note that we have $\kappa_j \leq \tau$ for $j \geq m+1$ from Equation~\ref{defn:kappa}.
	
	Finally, we bound the second term in Equation~\ref{eq:explore_first}. Note that we have $n^{i,\textup{miss}}_{\omega_i} \leq \frac{K-m}{m} \tau$, and hence, 
	\begin{align}
		\frac{1}{m}\E(\sum_{i \in [m]: \delta_i > \bar{\Delta}_i/2} \mu_i n^{i,\textup{miss}}_{\omega_i}) \leq \frac{K-m}{m}\frac{\tau}{m} \sum_{i=1}^m \mu_i P\left( \delta_i > \frac{\bar{\Delta}_i}{2} \right) \leq \frac{K-m}{m}\frac{\tau}{m} \sum_{i=1}^m \mu_i \min(1, \frac{320(K-m)}{T \bar{\Delta}_i^3}). \label{eq:explore_first_p2}
	\end{align}
	Here, the last inequality follows from Equation~\ref{eq:tailbound_gen}.
	
	Thus, combining Equations \ref{eq:explore_first_p1} and \ref{eq:explore_first_p2}, we have that the first term in Equation~\ref{eq:explore_gen} is bounded by
	\begin{align}
		&\frac{1}{m}\sum_{i=1}^m\mu_i \min\left( \sum_{j = m+1}^K \min\Big(\frac{11}{\Delta_j^2} + \frac{16}{\Delta_j^2} \log^+\left(\frac{T\Delta_j^3}{K-m}\right) + \frac{24}{\Delta_j^2}\sqrt{\log^+\left(\frac{T\Delta_j^3}{K-m}\right)}, \tau\Big) , ~\frac{K-m}{m}\tau \right) \notag\\
		&~~~+\frac{K-m}{m}\frac{\tau}{m} \sum_{i=1}^m \mu_i \min(1, \frac{320(K-m)}{T \bar{\Delta}_i^3}).
	\end{align}


	Next, we focus on the second term in Equation~\ref{eq:explore_gen}. For $j \in \{m+1, \dots,K\}$,
	\begin{align*}
		P( j \in \boldi^*) &= P\left( j \in \boldi^* \text{ and } \frac{\Delta_j}{2} < \bar{\delta} \right) + P\left( j \in \boldi^* \text{ and } \frac{\Delta_j}{2} \geq \bar{\delta} \right)\\
		&\leq \min \left(1,  ~P\left(\frac{\Delta_j}{2} < \bar{\delta} \right) + P\left( j \in \boldi^* \text{ and } \frac{\Delta_j}{2} \geq \bar{\delta} \right) \right).
	\end{align*}
	Recall that $\bar{\delta} = \max_{i\leq m} \delta_i$. Then, 
	\begin{align}
		P\left(\frac{\Delta_j}{2} < \bar{\delta}\right) &\leq P\left(\bigcup_{i=1}^m \left\{ \delta_i > \frac{\Delta_j}{2}  \right\} \right)\notag\\
		&\leq \min( 1, \sum_{i = 1}^m  P\left( \delta_i > \frac{\Delta_j}{2} \right) )\notag\\
		&\leq \min( 1, \frac{320 m(K-m)}{T \Delta_j^3} ). \label{eq:commit_misident_1}
	\end{align}
	Here, the last inequality follows from Equation~\ref{eq:tailbound_gen}.

	
	The event $\Delta_j/2 \geq \bar{\delta}$ together with the event $j \in \boldi^*$ for some $j \in \{m+1, \dots,K\}$ imply that there exists an arm $i \in [m]$ such that $i \notin \boldi^*$ and $\Delta_j/2 \geq \delta_i$. Therefore, these two events imply that there is some $n\leq \tau$ such that $\lcb_{n}^{j} = \bar{\mu}_n^j - \bar{\mu}_n^j \mathbbm{1}_{\{n< \tau\}} > \mu_i - \frac{\Delta_j}{2} \geq \mu_m - \frac{\Delta_j}{2} = \mu_j + \frac{\Delta_j}{2}$. Hence, 
	\begin{align}
		P\left( j \in \boldi^* \text{ and } \frac{\Delta_j}{2} \geq \delta_i \right) &\leq P\left(\exists\, n\leq \tau: \bar{\mu}^j_n-\bar{\mu}_{n}^{j}\mathbbm{1}_{\{n< \tau\}} > \mu_j + \Delta_j/2 \right)\notag\\
		&= P\left(\bar{\mu}^j_{\tau} > \mu_j + \Delta_j/2 \right)\notag\\
		&\overset{(a)}{\leq} \exp(-\frac{\tau \Delta_j^2}{2})\overset{(b)}{\leq} \frac{8(K-m)}{T\Delta_j^3}. \label{eq:commit_misident}
	\end{align}
	Here, (a) follows from Hoeffding's inequality, and (b) follows from the definition of $\tau$ and the fact that $\exp(-\alpha^{2/3}/2)\leq 8/\alpha$ for $\alpha\geq0$. 
	
	Thus, combining Equations~\ref{eq:commit_misident_1} and~\ref{eq:commit_misident}, we finally have 
	\begin{align}
		P( j \in \boldi^*) &\leq \min (1,  \frac{320 m(K-m)}{T \Delta_j^3} + \frac{8(K-m)}{T\Delta_j^3} )\notag\\
		&\leq \min (1,  \frac{328 m(K-m)}{T \Delta_j^3}),
	\end{align}
	$j \in \{m+1, \dots,K\}$. Additionally, since we have $\sum_{\boldi \in {K\choose m}}P(\boldi^* = \boldi) = 1$ and $| \boldi | = m$, we can bound $\sum_{j=m+1}^K P( j \in \boldi^*)$ by $m$. This is because the latter expression is counting each subset of $m$ arms at most $m$ times. 
	Then, 
	\begin{align}
		\sum_{j=m+1}^K \mu_j P( j \in \boldi^*) \leq \min\big(\sum_{j=m+1}^K \mu_j \min (1,  \frac{328 m(K-m)}{T \Delta_j^3}), ~m\mu_m  \big).
	\end{align}
	
	Bringing everything together, the regret from the Explore phase is bounded by
	\begin{align}
		&\frac{1}{m}\sum_{i=1}^m\mu_i \min\left( \sum_{j = m+1}^K \min\Big(\frac{11}{\Delta_j^2} + \frac{16}{\Delta_j^2} \log^+\left(\frac{T\Delta_j^3}{K-m}\right) + \frac{24}{\Delta_j^2}\sqrt{\log^+\left(\frac{T\Delta_j^3}{K-m}\right)}, \tau\Big) , ~\frac{K-m}{m}\tau \right) \notag\\
		&~~+\frac{K-m}{m}\frac{\tau}{m} \sum_{i=1}^m \mu_i \min(1, \frac{320(K-m)}{T \bar{\Delta}_i^3})\notag \\
		&~~+ \frac{K-m}{m}\frac{\tau}{m}\min\big(\sum_{j=m+1}^K \mu_j \min (1,  \frac{328 m(K-m)}{T \Delta_j^3}), ~m\mu_m  \big).
	\end{align}
	
	This finishes our derivation of a distribution dependent bound on the regret from the Explore phase. \\

	The following result will be useful in the coming parts. We aim to bound the probability of misidentifying an optimal arm, under the event $\frac{\bar{\Delta}_i}{2} \geq \delta_i$. 
	Under event $\frac{\bar{\Delta}_i}{2} \geq \delta_i$,  $i \notin \boldi^*$ but $j \in \boldi^*$ implies that there is some $n\leq \tau$ such that $\lcb_{n}^{j} = \bar{\mu}_n^j - \bar{\mu}_n^j \mathbbm{1}_{\{n< \tau\}} >  \mu_j + \frac{\bar{\Delta}_i}{2}$. Hence, 
	\begin{align}
		P\left( i \notin \boldi^*, ~j \in \boldi^*, ~\frac{\bar{\Delta}_i}{2} \geq \delta_i  \right) &\leq P\left( \exists\, n\leq \tau: \bar{\mu}^j_n-\bar{\mu}_{n}^j\mathbbm{1}_{\{n< \tau\}} > \mu_j+\frac{\bar{\Delta}_i}{2} \right)\notag\\
		&= P\left( \bar{\mu}^j_{\tau} > \mu_j+\frac{\bar{\Delta}_i}{2} \right)\notag\\
		&\overset{(a)}{\leq} \exp(-\frac{\tau \bar{\Delta}_i^2}{2}). \label{eq:missfromm_1}
	\end{align}
	Here, (a) follows from Hoeffding's inequality.\\
	
	We next focus on the regret arising from misidentification in the Commit phase.
	
	{\bf Regret from Commit.} This regret is upper bounded by 
	\begin{align}
		\sum_{\boldi \in {K\choose m}}\E\Big(\frac{T}{m}\Delta_{\boldi}\mathbbm{1}_{\{\boldi^* = \boldi\}}\Big).
	\end{align}
	We now get instance dependent and independent bounds on term above.
	
	{\bf An instance dependent bound on $\sum_{\boldi \in {K\choose m}}\E\Big(\frac{T}{m}\Delta_{\boldi}\mathbbm{1}_{\{\boldi^* = \boldi\}}\Big)$.} We have
	\begin{align}
		\sum_{\boldi \in {K\choose m}}\E\Big(\frac{T}{m}\Delta_{\boldi}\mathbbm{1}_{\{\boldi^* = \boldi\}}\Big) &= \frac{T}{m} \sum_{\boldi \in {K\choose m}}\E\Big(\frac{1}{m} \big( \sum_{i=1}^m \mu_i - \sum_{j \in \boldi} \mu_j \big) \mathbbm{1}_{\{\boldi^* = \boldi\}}\Big) \notag\\
		&=\frac{T}{m} \frac{1}{m} \E\left( \sum_{i=1}^m \mu_i - \sum_{j = 1}^K \mu_j \mathbbm{1}_{\{j \in \boldi^*\}}  \right). \label{eq:focusing}
	\end{align}
	We focus on the term inside the paranthesis in Expression~\ref{eq:focusing}. To that end, we define $s^* \triangleqq \min(\frac{1}{\mid [m]\setminus\boldi^* \mid}, 1)$, the reciprocal of the number of suboptimal arms in the exploitation set $\boldi^*$, given that there are any. Otherwise, we set $s^*=1$. If there is more than one misidentified arm, this definition of $s^*$ will ensure that we are not counting respective arms multiple times in the below expression.
	\begin{align}
		\E\left( \sum_{i=1}^m \mu_i - \sum_{j = 1}^K \mu_j \mathbbm{1}_{\{j \in \boldi^*\}}  \right)&\overset{(a)}{=} \E\left( \sum_{i=1}^m \mu_i \left( s^* \sum_{j = m+1}^K \mathbbm{1}_{\{j \in \boldi^*\}} \right) \right) - \E\left( \sum_{j = 1}^K \mu_j \mathbbm{1}_{\{j \in \boldi^*\}} \left(  s^* \sum_{i = 1}^m \mathbbm{1}_{\{i \notin \boldi^*\}} \right) \right) \label{eq:commit_start}\\
		&\overset{(b)}{=} \E\left( s^* \sum_{i=1}^m \sum_{j = m+1}^K \mu_i \mathbbm{1}_{\{j \in \boldi^*\}} \right)- \E\left( s^* \sum_{i=1}^m \sum_{j = 1}^K \mu_j \mathbbm{1}_{\{i \notin \boldi^*, ~j \in \boldi^*\}}\right)\notag\\
		&= \E\left( s^* \sum_{i=1}^m \sum_{j = m+1}^K \mu_i \mathbbm{1}_{\{i \notin \boldi^*, ~j \in \boldi^*\}}\right) + \E\left( s^* \sum_{i=1}^m \sum_{j = m+1}^K \mu_i \mathbbm{1}_{\{i \in \boldi^*, ~j \in \boldi^*\}} \right)\notag\\
		&~~ - \E\left( s^*\sum_{i=1}^m \sum_{j = m+1}^K \mu_j \mathbbm{1}_{\{i \notin \boldi^*, ~j \in \boldi^*\}}\right) - \E\left( s^*\sum_{i=1}^m \sum_{j = 1}^m \mu_j \mathbbm{1}_{\{i \notin \boldi^*, ~j \in \boldi^*\}} \right)\\
		&\overset{(c)}{=} \E\left( s^* \sum_{i=1}^m \sum_{j = m+1}^K \mu_i \mathbbm{1}_{\{i \notin \boldi^*, ~j \in \boldi^*\}}\right) - \E\left( s^*\sum_{i=1}^m \sum_{j = m+1}^K \mu_j \mathbbm{1}_{\{i \notin \boldi^*, ~j \in \boldi^*\}}\right) \notag\\
		&=\E\left(s^* \sum_{i = 1}^m \sum_{j = m+1}^K (\mu_i-\mu_j) \mathbbm{1}_{\{ i \notin \boldi^*,~ j \in \boldi^*\}} \right).
	\end{align}
	Here, (a) follows from the definition of $s^*$. Note that, if $\boldi^* = [m]$, then $s^* = 1$ but $s^* \sum_{j = m+1}^K \mathbbm{1}_{\{j \in \boldi^*\}} = s^* \sum_{i = 1}^m \mathbbm{1}_{\{i \notin \boldi^*\}} = 0$. Nevertheless, this does not imply that (a) is invalid. If $\boldi^* = [m]$, then we have zero in the left-hand side of Equation~\ref{eq:commit_start} too. (b) follows from reorganizing terms. (c) is due to the following fact:
	\begin{align}
		\E\left( s^* \sum_{i=1}^m \sum_{j = m+1}^K \mu_i \mathbbm{1}_{\{i \in \boldi^*, ~j \in \boldi^*\}} \right) & = \E\left( s^* \sum_{i=1}^m \mu_i \mathbbm{1}_{\{i \in \boldi^*\}} \sum_{j = m+1}^K \mathbbm{1}_{\{j \in \boldi^*\}} \right) \notag\\
		&\overset{(a)}{=}\E\left( s^* \sum_{j=1}^m \mu_j \mathbbm{1}_{\{j \in \boldi^*\}} \sum_{i = m+1}^K \mathbbm{1}_{\{i \in \boldi^*\}} \right) \notag\\
		&\overset{(b)}{=}\E\left( s^* \sum_{j=1}^m \mu_j \mathbbm{1}_{\{j \in \boldi^*\}} \sum_{i = 1}^m \mathbbm{1}_{\{i \notin \boldi^*\}} \right)\notag \\
		& =\E\left( s^*\sum_{i=1}^m \sum_{j = 1}^m \mu_j \mathbbm{1}_{\{i \notin \boldi^*, ~j \in \boldi^*\}} \right).
	\end{align}
	We swap indices in (a). (b) follows from the fact that the number of the suboptimal arms in $\boldi^*$ must be same as the number of optimal arms missing from $\boldi^*$.
	Hence, we have that
	\begin{align}
		\sum_{\boldi \in {K\choose m}}\E\Big(\frac{T}{m}\Delta_{\boldi}\mathbbm{1}_{\{\boldi^* = \boldi\}}\Big)=\frac{T}{m^2} \E\left(s^* \sum_{i = 1}^m \sum_{j = m+1}^K (\mu_i-\mu_j) \mathbbm{1}_{\{ i \notin \boldi^*,~ j \in \boldi^*\}} \right). \label{eq:rewriteregret}
	\end{align}
	
	Since $s^*$ is at most $1$, 
	\begin{align}
		\sum_{\boldi \in {K\choose m}}\E\Big(\frac{T}{m}\Delta_{\boldi}\mathbbm{1}_{\{\boldi^* = \boldi\}}\Big) &\leq \frac{T}{m^2} \E\left(\sum_{i = 1}^m \sum_{j = m+1}^K (\mu_i-\mu_j) \mathbbm{1}_{\{ i \notin \boldi^*,~ j \in \boldi^*\}} \right)\\
		&= \frac{T}{m^2} \sum_{i = 1}^m \sum_{j = m+1}^K (\mu_i-\mu_j) P(i \notin \boldi^*, ~j \in \boldi^*).
	\end{align}
	
	We can further break $P(i \notin \boldi^*, ~j \in \boldi^*)$ down as follows:
	\begin{align}
		P(i \notin \boldi^*, ~j \in \boldi^*) &= P(i \notin \boldi^*, ~j \in \boldi^*, \delta_i > \frac{\bar{\Delta}_i}{2}, \delta_i > \frac{\Delta_j}{2})\notag\\ 
		&~~+  P(i \notin \boldi^*, ~j \in \boldi^*, \delta_i \leq \frac{\bar{\Delta}_i}{2}, \delta_i > \frac{\Delta_j}{2})\notag\\
		&~~+ P(i \notin \boldi^*, ~j \in \boldi^*, \delta_i > \frac{\bar{\Delta}_i}{2}, \delta_i \leq \frac{\Delta_j}{2})\notag\\ 
		&~~+  P(i \notin \boldi^*, ~j \in \boldi^*, \delta_i \leq \frac{\bar{\Delta}_i}{2}, \delta_i \leq \frac{\Delta_j}{2}).
	\end{align}
	Then, using Equations~\ref{eq:tailbound_gen},~\ref{eq:commit_misident} and~\ref{eq:missfromm_1}, we have
	\begin{align}
		P(i \notin \boldi^*, ~j \in \boldi^*) &\leq \min\Bigg( \min\left(1, \frac{320 (K-m)}{T \bar{\Delta}_i^3}, \frac{320 (K-m)}{T \Delta_j^3} \right) \notag\\
		&~~+ \exp(-\frac{\tau \bar{\Delta}_i^2}{2}) + \exp(-\frac{\tau \Delta_j^2}{2}) + \min\Big(\exp(-\frac{\tau \bar{\Delta}_i^2}{2}), \exp(-\frac{\tau \Delta_j^2}{2})\Big), 1 \Bigg)\notag\\
		&\leq \min\left( \min\left(\frac{320 (K-m)}{T \bar{\Delta}_i^3}, \frac{320 (K-m)}{T \Delta_j^3} \right) + \frac{3}{2}\exp(-\frac{\tau \bar{\Delta}_i^2}{2}) + \frac{3}{2}\exp(-\frac{\tau \Delta_j^2}{2}), 1 \right).
	\end{align}
	
	Then, the instance dependent bound on the regret from Commit is
	\begin{align}
		\frac{T}{m^2} \sum_{i = 1}^m \sum_{j = m+1}^K (\mu_i-\mu_j) \min\left( \min\left(\frac{320 (K-m)}{T \bar{\Delta}_i^3}, \frac{320 (K-m)}{T \Delta_j^3} \right) + \frac{3}{2}\exp(-\frac{\tau \bar{\Delta}_i^2}{2}) + \frac{3}{2}\exp(-\frac{\tau \Delta_j^2}{2}), 1 \right).
	\end{align}
	
	\textbf{An instance independent bound on $\sum_{\boldi \in {K\choose m}}\E\Big(\frac{T}{m}\Delta_{\boldi}\mathbbm{1}_{\{\boldi^* = \boldi\}}\Big)$.} Consider Equation~\ref{eq:rewriteregret} again:
	\begin{align}
		\sum_{\boldi \in {K\choose m}}\E\Big(\frac{T}{m}\Delta_{\boldi}\mathbbm{1}_{\{\boldi^* = \boldi\}}\Big)=\frac{T}{m^2} \E\left(s^* \sum_{i = 1}^m \sum_{j = m+1}^K (\mu_i-\mu_j) \mathbbm{1}_{\{ i \notin \boldi^*,~ j \in \boldi^*\}} \right).
	\end{align}
	
	Then,
	\begin{align}
		\sum_{\boldi \in {K\choose m}}\E\Big(\frac{T}{m}\Delta_{\boldi}\mathbbm{1}_{\{\boldi^* = \boldi\}}\Big)
		&= \frac{T}{m^2} \E\left(s^* \sum_{i = 1}^m \sum_{j = m+1}^K (\mu_i-\mu_j) \mathbbm{1}_{\{ i \notin \boldi^*,~ j \in \boldi^*\}} \right) \label{eq:count2}\\
		&\overset{(a)}{\leq} \frac{T}{m^2} \E\left(s^* \sum_{i = 1}^m \sum_{j = m+1}^K (\bar{\Delta}_i+\Delta_j) \mathbbm{1}_{\{ i \notin \boldi^* , ~j \in \boldi^*\}} \right)\notag\\
		&= \frac{T}{m^2} \E\left(s^*  \sum_{i = 1}^m \sum_{j = m+1}^K (\bar{\Delta}_i+\Delta_j) \mathbbm{1}_{\{ i \notin \boldi^*, ~j \in \boldi^*, ~\delta_i > \bar{\Delta}_i/2, ~\delta_i > \Delta_j/2\}} \right) \label{eq:commitreg_4}\\
		&~+ \frac{T}{m^2} \E\left(s^* \sum_{i = 1}^m \sum_{j = m+1}^K (\bar{\Delta}_i+\Delta_j) \mathbbm{1}_{\{ i \notin \boldi^*, ~j \in \boldi^* , ~\delta_i \leq \bar{\Delta}_i/2, ~\delta_i > \Delta_j/2\}} \right)\label{eq:commitreg_2}\\
		&~+ \frac{T}{m^2} \E\left(s^* \sum_{i = 1}^m \sum_{j = m+1}^K (\bar{\Delta}_i+\Delta_j) \mathbbm{1}_{\{ i \notin \boldi^*, ~j \in \boldi^* , ~\delta_i > \bar{\Delta}_i/2, ~\delta_i \leq \Delta_j/2\}} \right)\label{eq:commitreg_3}\\
		&~+ \frac{T}{m^2} \E\left(s^* \sum_{i = 1}^m \sum_{j = m+1}^K (\bar{\Delta}_i+\Delta_j) \mathbbm{1}_{\{ i \notin \boldi^*,~ j \in \boldi^* , ~\delta_i \leq \bar{\Delta}_i/2, ~\delta_i \leq \Delta_j/2\}} \right). \label{eq:commitreg_1}
	\end{align}
	Here, (a)  follows from the definition of $\bar{\Delta}_i$ and $\Delta_j$, i.e., $\mu_i -\mu_j  = \mu_i - \mu_{m+1} + \mu_m -\mu_j + \mu_{m+1} - \mu_m \leq \bar{\Delta}_i + \Delta_j$ for $i \in [m]$, $j\geq m+1$. \\

	We proceed by analyzing each term in Expressions~\ref{eq:commitreg_4}--\ref{eq:commitreg_1}. We start with the term in Expression~\ref{eq:commitreg_4}:
	\begin{align}
		\frac{T}{m^2} \E\left(s^*  \sum_{i = 1}^m \sum_{j = m+1}^K (\bar{\Delta}_i+\Delta_j) \mathbbm{1}_{\{ i \notin \boldi^*, ~j \in \boldi^*, ~\delta_i > \bar{\Delta}_i/2, ~\delta_i > \Delta_j/2\}} \right) &\leq \frac{T}{m^2} \E\left( 4 \sum_{i = 1}^m \delta_i \sum_{j = m+1}^K s^*  \mathbbm{1}_{\{ i \notin \boldi^*, ~j \in \boldi^*\}} \right)\notag\\
		&\leq 4 \frac{T}{m^2} \sum_{i=1}^{m}\E(\delta_i).
	\end{align}
	Here, the second inequality follows from the definition of $s^*$ since $\sum_{j = m+1}^K s^*  \mathbbm{1}_{\{ i \notin \boldi^*, ~j \in \boldi^*\}} \leq 1$  for each $i \in[m]$. 
	We then look at $\E( \delta_i )$ for $i \in [m]$. We have,
	
	\[ \E( \delta_i ) = \int_{0}^{\infty} \p\left( \delta_i > x \right) dx \leq \int_{0}^{\infty} \min \left(1, \frac{40(K-m)}{Tx^3}\right)dx.\] 
	This integral evaluates to \[ \int_0^{\frac{(40(K-m))^{1/3}}{T^{1/3}}} dx + \int_{\frac{(40(K-m))^{1/3}}{T^{1/3}}}^\infty \frac{40(K-m)}{Tx^3} dx \leq 2\frac{(40(K-m))^{1/3}}{T^{1/3}}. \] 
	
	Combining these results, we have 
	\begin{align}
		\E( \delta_i ) \leq 2\frac{(40(K-m))^{1/3}}{T^{1/3}},
	\end{align}
	for each $i \in [m]$. 
	Thus we have,
	\begin{align}
		\frac{T}{m^2} \E\left(s^*  \sum_{i = 1}^m \sum_{j = m+1}^K (\bar{\Delta}_i+\Delta_j) \mathbbm{1}_{\{ i \notin \boldi^*, ~j \in \boldi^*, ~\delta_i > \bar{\Delta}_i/2, ~\delta_i > \Delta_j/2\}} \right) \leq 8 \frac{(40(K-m))^{1/3}T^{2/3}}{m}. \label{eq:commitreg_first}
	\end{align}\\
	
	Next, we look at the term in Expression~\ref{eq:commitreg_2}:
	\begin{align}
		&\frac{T}{m^2} \E\left(s^* \sum_{i = 1}^m \sum_{j = m+1}^K (\bar{\Delta}_i+\Delta_j) \mathbbm{1}_{\{ i \notin \boldi^*, ~j \in \boldi^* , ~\delta_i \leq \bar{\Delta}_i/2, ~\delta_i > \Delta_j/2\}} \right) \notag\\
		&~~\overset{(a)}{\leq} 2\frac{T}{m^2}  \E\left( \sum_{i = 1}^m \bar{\Delta}_i \sum_{j = m+1}^K s^* \mathbbm{1}_{\{ i \notin \boldi^*, ~j \in \boldi^* , ~\delta_i \leq \bar{\Delta}_i/2\}} \right)\label{eq:midstep}\\
		&~~=2\frac{T}{m^2}  \E\left( \sum_{i = 1:\bar{\Delta}_i < \frac{(K-m)^{1/3}\sqrt{2\log(K-m)}}{T^{1/3}}}^m \bar{\Delta}_i \sum_{j = m+1}^K s^* \mathbbm{1}_{\{ i \notin \boldi^*, ~j \in \boldi^* , ~\delta_i \leq \bar{\Delta}_i/2\}} \right)\notag\\
		&~~~~+ 2\frac{T}{m^2}  \E\left( \sum_{i = 1:\bar{\Delta}_i \geq \frac{(K-m)^{1/3}\sqrt{2\log(K-m)}}{T^{1/3}}}^m \bar{\Delta}_i \sum_{j = m+1}^K s^* \mathbbm{1}_{\{ i \notin \boldi^*, ~j \in \boldi^* , ~\delta_i \leq \bar{\Delta}_i/2\}} \right)\notag\\
		&~~\overset{(b)}{\leq} 2\frac{T}{m^2} \frac{(K-m)^{1/3}\sqrt{2\log(K-m)}}{T^{1/3}} \E\left( \sum_{i = 1}^m \sum_{j = m+1}^K s^* \mathbbm{1}_{\{ i \notin \boldi^*, ~j \in \boldi^* , ~\delta_i \leq \bar{\Delta}_i/2\}} \right)\notag\\
		&~~~~+ 2\frac{T}{m^2}   \sum_{i = 1:\bar{\Delta}_i \geq \frac{(K-m)^{1/3}\sqrt{2\log(K-m)}}{T^{1/3}}}^m \bar{\Delta}_i ~\E\left(\sum_{j = m+1}^K \mathbbm{1}_{\{ i \notin \boldi^*, ~j \in \boldi^* , ~\delta_i \leq \bar{\Delta}_i/2\}} \right)\notag\\
		&~~\overset{(c)}{\leq} 2\frac{T}{m^2} ~\frac{(K-m)^{1/3}\sqrt{2\log(K-m)}}{T^{1/3}} ~m\notag\\
		&~~~~+ 2\frac{T}{m^2}   \sum_{i = 1:\bar{\Delta}_i \geq \frac{(K-m)^{1/3}\sqrt{2\log(K-m)}}{T^{1/3}}}^m \bar{\Delta}_i ~\sum_{j = m+1}^K P(i \notin \boldi^*, ~j \in \boldi^* , ~\delta_i \leq \bar{\Delta}_i/2)\notag\\
		&~~\overset{(d)}{\leq} 2\frac{(K-m)^{1/3}T^{2/3}\sqrt{2\log(K-m)}}{m}\notag\\
		&~~~~+ 2\frac{T}{m^2} (K-m)  \sum_{i = 1:\bar{\Delta}_i \geq \frac{(K-m)^{1/3}\sqrt{2\log(K-m)}}{T^{1/3}}}^m \bar{\Delta}_i  \exp(-\frac{\tau \bar{\Delta}_i^2}{2}) \notag\\
		&~~\overset{(e)}{\leq} 4\frac{(K-m)^{1/3}T^{2/3}\sqrt{2\log(K-m)}}{m}. \label{eq:commit_term2}
	\end{align}
	Here, (a) follows from the conditions of the indicator, that is, $\Delta_j/2 < \delta_i \leq \bar{\Delta}_i/2$. (b) follows from the condition on $\bar{\Delta}_i$ and the fact that $s^* \leq 1$. Now, (c) follows from 
	\begin{align}
		\sum_{i = 1}^m \sum_{j = m+1}^K s^* \mathbbm{1}_{\{ i \notin \boldi^*, ~j \in \boldi^* , ~\delta_i \leq \bar{\Delta}_i/2\}} \leq \sum_{i = 1}^m \sum_{j = m+1}^K s^* \mathbbm{1}_{\{ i \notin \boldi^*, ~j \in \boldi^*\}} \leq m, \label{eq:maxcommit}
	\end{align} that is, we can at most have $m$ suboptimal arms that we eventually commit to. (d) is due to Equation~\ref{eq:missfromm_1}. Finally, (e) follows from the unimodality of $\exp(-\frac{\tau x^2}{2})x$, an argument similar to that of Equation~\ref{ind-bound-1}. Note that in the case of $2m > K$, we have $\sum_{i = 1}^m \sum_{j = m+1}^K s^* \mathbbm{1}_{\{ i \notin \boldi^*, ~j \in \boldi^*\}} \leq K-m.$ \\
	
	Notice that we can directly bound the term in Expression~\ref{eq:commitreg_3} as it is symmetric to the term in Expression~\ref{eq:commitreg_2}. Hence,
	\begin{align}
		\frac{T}{m^2} \E\left(s^* \sum_{i = 1}^m \sum_{j = m+1}^K (\bar{\Delta}_i+\Delta_j) \mathbbm{1}_{\{ i \notin \boldi^*, ~j \in \boldi^* , ~\delta_i > \bar{\Delta}_i/2, ~\delta_i \leq \Delta_j/2\}} \right) \leq 4\frac{(K-m)^{1/3}T^{2/3}\sqrt{2\log(K-m)}}{m}. \label{eq:commit_term3}
	\end{align}\\
	
	Finally, we can bound the term in Expression~\ref{eq:commitreg_1} referring to the techniques we used for bounding the term in Expression~\ref{eq:commitreg_2}, and Expression~\ref{eq:commitreg_3} as well:
	\begin{align}
		&\frac{T}{m^2} \E\left( \sum_{i = 1}^m \sum_{j = m+1}^K (\bar{\Delta}_i+\Delta_j) s^* \mathbbm{1}_{\{ i \notin \boldi^*,~ j \in \boldi^* , ~\delta_i \leq \bar{\Delta}_i/2, ~\delta_i \leq \Delta_j/2\}} \right)\notag\\
		&~~\leq \frac{T}{m^2} \E\left( \sum_{i = 1}^m \sum_{j = m+1}^K \bar{\Delta}_i s^* \mathbbm{1}_{\{ i \notin \boldi^*,~ j \in \boldi^* , ~\delta_i \leq \bar{\Delta}_i/2\}} \right) +\frac{T}{m^2} \E\left( \sum_{i = 1}^m \sum_{j = m+1}^K \Delta_j s^* \mathbbm{1}_{\{ i \notin \boldi^*,~ j \in \boldi^* , ~\delta_i \leq \Delta_j/2\}} \right)\notag\\
		&~~\leq4\frac{(K-m)^{1/3}T^{2/3}\sqrt{2\log(K-m)}}{m}. \label{eq:commit_term4}
	\end{align}
	Here, the final inequality is due to the series of bounds on Expression~\ref{eq:midstep}.\\

	Combining Equations~\ref{eq:commitreg_first},~\ref{eq:commit_term2},~\ref{eq:commit_term3}, and~\ref{eq:commit_term4}, we have
	\begin{align}
		\frac{T}{m^2} \E\left( \sum_{i = 1}^m \mu_i \mathbbm{1}_{\{ i \notin \boldi^* \}} - \sum_{j = m+1}^K \mu_j \mathbbm{1}_{\{ j \in \boldi^* \}}\right) &\leq \frac{8 (40(K-m))^{1/3}T^{2/3} + 12 (K-m)^{1/3}T^{2/3}\sqrt{2\log(K-m)}}{m} \notag \\
		&\leq 45 \frac{(K-m)^{1/3}T^{2/3}\sqrt{\log(K-m)}}{m}.	\label{eq:commit_final}
	\end{align}
	
	The instance independent bound follows from the fact that the regret from the Explore phase is at most $\frac{K}{m}\tau - \tau = \textup{O}(\frac{(K-m)^{1/3}T^{2/3}}{m})$ and from Equation~\ref{eq:commit_final}.
	\hfill\Halmos
	\vspace{1cm}
	\begin{lemma}\label{lma:exploit}
		Assume that $\exists t \in \{ \tau+1, \dots, \lfloor\Tminline\rfloor-1\}$ such that $n_{t-1}^i = \tau$ and $n_{t}^i = \tau+1$ for some arm $i \in [K]$. Then, $i \in \textup{\boldi}^*$ and $n_{\Tminline}^i = \lfloor\Tminline\rfloor+\tau+1-t$, i.e., arm $i$ is pulled in all time periods following $t$.
	\end{lemma}
	\proof{Proof of Lemma~\ref{lma:exploit}.}
	If there is an arm $i$ satisfying the conditions of the lemma, then 
	\[ \ucb^i_{n^i_{t-1}} = \ucb^i_{\tau} = \bar{\mu}_{\tau}^i = \lcb^i_{\tau} = \lcb^i_{n^i_{t-1}}. \]
	Following the definition of the empirical average reward for arm $j$, 
	\[ \bar{\mu}_{\tau + s}^i = \bar{\mu}_{\tau}^i \]
	for all $j \in [K]$ and $s \geq 1$. Hence, for arms with at least $\tau$ pulls on them, the empirical average reward and upper/lower confidence bounds are the same.
	
	Define the set $l_s = \{j \in [K]: n_{s}^j \geq \tau+1 \}$, the set of the arms with at least $\tau+1$ pulls on them at time $s$. Now, we have that $i \in l_t$, and since it is pulled at time $t$, (1) it is among the $m$ arms with the highest upper confidence bounds at time $t$, and (2) its empirical average reward is not updated after it is pulled. In fact, none of the arms in $l_t$ have their empirical average rewards updated after they are pulled, and all are in $l_{t+1}$, that is, $l_s \subset l_{s+1}$ for $s \in \{1,\dots, \lfloor\Tminline\rfloor-1\}$, which we prove next.
	
	Let $\boldi_s$ denote the set of arms pulled at time $s$. We claim that $l_s \subset \boldi_s$ for $s \leq \lfloor\Tminline\rfloor$.
	
	Without loss of generality, and for ease of discussion, assume that arm $i$ is the first element of set $l_t$, i.e., $l_{t-1} = \emptyset$ and $l_t = i$. Therefore we also have that $n_{t-1}^i = \tau$ and $n_{t}^i = \tau + 1$ since $i \in \boldi_t$. At time $t$, there are $(m-1)$ other arms that are being pulled. Then, arm $i$ will be among the $m$ arms with the highest upper confidence bounds at time $(t+1)$ as well. This is because at most $(m-1)$ arms' upper confidence bounds are updated after the pulls at time $t$ and might exceed the empirical average reward of arm $i$. Therefore, $i \in \boldi_{t+1}$, and repeating the same argument gives $i \in \boldi_{s}$ for $s \geq t+2$. 
	
	Now, assume that $\exists t_1 > t$ such that $l_t = i$ and $\{i,j\} \in l_{t_1}$. As before, we have that $n_{t_1-1}^j = \tau$ and $n_{t_1}^j = \tau + 1$ since $j \in \boldi_{t_1}$. Per the same argument as above, at time $t_1$, there are $(m-2)$ other arms that are being pulled. Then, arms $i$ and $j$ will both be among the $m$ arms with the highest upper confidence bounds at time $(t_1+1)$ as well. Therefore, $i, j \in \boldi_{t_1+1}$, and, as before, $i, j \in \boldi_{s}$ for $s \geq t_1+2$. 
	
	Following in this fashion, we get $l_s = \boldi_s$ for $s \geq \omega+1$, where $\omega = \max_{i \in \boldi^*} \omega_i$. Recall that, for $i \in \boldi^*$, we defined $\omega_i$ as the time period prior to arm $i$ being pulled $(\tau + 1)-\textup{st}$ time, i.e., $\omega_i = \min \left\{ t \leq \frac{T}{m}: n_{t+1}^{i} = \tau + 1 \right\}$. That is, when there are exactly $m$ arms with at least $\tau+1$ pulls on them, $m\textup{-ADA-ETC}$ is surely in the Commit phase. 
	\hfill\Halmos\\
	\endproof
	
\section{Auxiliary results.}\label{apx:aux}
\subsection{Asymptotic near-optimality of common policies for the max objective}\label{apxsec:asym}
We show that any policy limiting the number of times a suboptimal arm is pulled in a multi-armed bandit problem with distinct mean rewards must perform well for the max objective.
\begin{prop}
Consider a $K$-armed bandit instance $\nu$ with distinct means $\mu_i$ for $i=1,\cdots, K$, where arm 1 is the optimal arm. For arm $i$, let $T_i$ be the number of times it is pulled by a policy $\pi$. Suppose that the policy $\pi$ ensures that 
$E(T_i) \leq Q_i(T)$ for some quantity $Q_i(T) = \textup{o}(T)$ for any $i\geq 2$. Then the max regret under this policy is bounded as:
$$\mathcal{\textup{Reg}}_{\textup{MAX},T}(\pi, \bnu) \leq \sum_{i\geq 2} Q_i(T)\mu_i$$
for any $T$ large enough.
\end{prop}
\proof{Proof.}
The expected max reward under policy $\pi$ is bounded as:
\begin{align}
		\mathcal{R}_T(\pi,\bnu) &= \textup{E}\big(\max_{i\in[K]} \overline{U}^i_{T}\big)\nonumber\\
		&\geq \max_{i\in[K]}\textup{E}\big(\overline{U}^i_{T}\big)\\
		&=\max_{i\in[K]}\mu_i\textup{E}\big(T_i\big)\\
		&= \max\big(\mu_1(T-\sum_{i\geq 2} E(T_i)), \max_{i\geq 2}\mu_i\textup{E}\big(T_i\big)\big)\\
		&\overset{(a)}{=} \mu_1(T-\sum_{i\geq 2} E(T_i)) \textup{ (for large enough $T$)}\\
		&\geq \mu_1(T-\sum_{i\geq 2} Q_i(T))  \textup{ (for large enough $T$).}
\end{align}
Here (a) follows from the fact that $\textup{E}\big(T_i\big) \leq Q_i(T) = \textup{o}(T)$ for $i\geq 2$. This implies that the max regret is bounded by $\mu_1T - \mu_1(T-\sum_{i\geq 2} Q_i(T)) = \sum_{i\geq 2} Q_i(T)\mu_i$ for any $T$ large enough.
\hfill \Halmos
\begin{corollary}
Consider a $K$-armed bandit instance $\nu$ with distinct means $\mu_i$ for $i=1,\cdots, K$, where arm 1 is the optimal arm.  Suppose that the policy $\pi$ ensures that 
$$\limsup_{T\rightarrow\infty}\frac{E(T_i)}{\log T} \leq \frac{C_1}{d_i}$$
for any suboptimal arm $i$ for some $C_1\geq 1$. Then this policy asymptotically achieves the instance-optimal regret upto a constant factor for the max objective. 
\end{corollary}
Since this property is satisfied by many UCB policies, including UCB1, this implies that these policies are asymptotically instance-optimal up to a constant factor for the max objective. 

\subsection{Proof of Proposition \ref{prop:ucbbad}}\label{apx:ucbbad}
	First we fix a $T$ and assume it is even for convenience. 
	Let arm $1$ be the arm with Bernoulli(0.5 + $\nicefrac{1}{\sqrt{T}}$) rewards and arm $2$ be the arm with Bernoulli(0.5) rewards, i.e., $\mu_1 = 0.5 + \nicefrac{1}{\sqrt{T}}$, $\mu_2 = 0.5$. 
	Let $\ucb_i(t)$ denote the upper confidence bound (UCB) of arm $i$ after it receives $t$-th pull. We use the definition
	\begin{align}
		\ucb_i(t) \triangleqq \bar{\mu}^i_{t} + c \sqrt{\frac{\log(T)}{t}}, \label{eq:UCB1define}
	\end{align}
	for some $c > \sqrt{\frac{1}{2}}$. Under the UCB1 policy, the arm $i$ with the largest UCB is pulled at time $t$.
	
	Next, consider the following event $G_i$ for arm $i$, $i \in \{1, 2\}$, for some $d \in (\sqrt{\frac{1}{2}}, ~c)$,
	\begin{align}
		G_i = \Big\{ \mid \mu_i - \bar{\mu}^i_t \mid \leq d \sqrt{\frac{\log(T)}{t}}  ~\forall t \in [T] \Big\}.  \label{eq:UCB1defineEvent} 
	\end{align}
	Using Hoeffding's lemma, we have that $\p\left( G_i \right) \geq 1 - \frac{2}{T^{(2d^2-1)}}$ for $i \in \{1, 2\}$. Hence, the good event 
	\begin{align}
		G = G_1 \cap G_2 \notag
	\end{align}
	occurs with probability at least $1 - \frac{4}{T^{(2d^2-1)}}$. We condition our following arguments on this good event $G$.
	
	Notice that, independent of the choice for a policy, after $T$ pulls are depleted, we can either have (i) both arms receiving $T/2$ pulls, (ii) arm $1$ receiving strictly more pulls than arm $2$, or (iii) arm $2$ receiving strictly more pulls than arm $1$. Since there is nothing to prove in the first case (the max objective regret would be linear in $T$ due to Lemma~\ref{lma:forUCB}), we will focus on the other two cases and construct the following two events:
	\begin{align}
		B_1(\tau) &= \{ \min_{t \in [\tau]} \ucb_1(t) >  \ucb_2(T/2) \} \label{eq:ucbComparing11},\\
		B_2(\tau) &= \{ \min_{t \in [\tau]} \ucb_2(t) >  \ucb_1(T/2) \} \label{eq:ucbComparing21}.
	\end{align}
	
	The key observation is that if $B_i(\tau)$ occurs, then arm $i$ gets pulled at least $\tau$ times by the time the arm $-i$ gets pulled $T/2$ times. We will show that both events $B_1(b T)$ and $B_2(bT)$ occur with high probability for an appropriate constant $b>0$, so that irrespective of which arm gets $T/2$ pulls (at least one arm must), the other arm obtains at least $bT$ pulls. We can therefore conclude that the less pulled arm receives a constant fraction of $T$ pulls with high probability.
Considering event $B_2(\tau)$, $\tau$ to be specified later,
	\begin{align}
		\p\left( (B_2(\tau))^c,\, G \right) &= \p\left( \exists t \leq \tau \textup{ such that } \ucb_2(t) \leq  \ucb_1(T/2),\, G \right) \notag\\
		& \leq \sum_{t=1}^{\tau} \p\left(\ucb_2(t) \leq  \ucb_1(T/2), \, G \right) \notag\\
		& = \sum_{t=1}^{\tau} \p\left(\bar{\mu}^2_{t} + c\sqrt{\frac{\log(T)}{t}} \leq \bar{\mu}^1_{T/2} + c\sqrt{\frac{2\log(T)}{T}}, \, G \right) \notag\\
		& \overset{(a)}{\leq} \sum_{t=1}^{\tau} \p\left(\bar{\mu}^2_{t} + c\sqrt{\frac{\log(T)}{t}} \leq \mu_1 + (c + d)\sqrt{2}\sqrt{\frac{\log(T)}{T}},\, G \right) \notag\\
		& \overset{(b)}{=} \sum_{t=1}^{\tau} \p\left(\bar{\mu}^2_{t} + c\sqrt{\frac{\log(T)}{t}} \leq \mu_2 + \frac{1}{\sqrt{T}} + (c + d)\sqrt{2}\sqrt{\frac{\log(T)}{T}},\, G \right) \notag\\
		& \leq \sum_{t=1}^{\tau} \p\left(\bar{\mu}^2_{t} + c\sqrt{\frac{\log(T)}{t}} \leq \mu_2 + (c + d+1)\sqrt{2}\sqrt{\frac{\log(T)}{T}},\, G \right) \notag\\
		& = \sum_{t=1}^{\tau} \p\left( c\sqrt{\frac{\log(T)}{t}} -(c + d+1)\sqrt{2}\sqrt{\frac{\log(T)}{T}} \leq \mu_2 - \bar{\mu}^2_{t},\, G \right) \notag\\
		& \overset{(c)}{\leq} \sum_{t=1}^{\tau} \p\left( c\sqrt{\frac{\log(T)}{t}} -(c + d+1)\sqrt{2}\sqrt{\frac{\log(T)}{T}} \leq d\sqrt{\frac{\log(T)}{t}},\, G \right) \notag\\
		& \leq  \sum_{t=1}^{\tau} \p\left((c-d)\sqrt{\frac{\log(T)}{t}} \leq (c + d+1)\sqrt{2}\sqrt{\frac{\log(T)}{T}} \right). \label{eq:ucbCompare}
	\end{align}
	Here, the second equality follows from using the definition of arm indices for the UCB1 policy. 
	Inequality (a) holds since, on the event $G$, the empirical mean of arm $1$ is always within $d\sqrt{\frac{\log(T)}{T/2}}$ of the actual mean after $T/2$ pulls. 
	(b) holds since arm $2$ is the suboptimal arm. 
	(c) is due to the fact that on event G, the empirical mean of arm $2$ is always within $d\sqrt{\frac{\log(T)}{t}}$ of the actual mean after $t$ pulls. Letting $\kappa = \frac{(c-d)^2}{2(c + d + 1)^2}$, Expression~\ref{eq:ucbCompare} is zero for $\tau < \kappa T$. Hence, we have that $\p(B_2(\kappa T - 1)^c,G) =0$, implying that 
	$$P(B_2(\kappa T - 1))\geq P(G) \geq 1 - \frac{4}{T^{(2d^2-1)}}= 1-\textup{o}(1).$$ 
	A similar argument can be made for event $B_1$ to conclude that the event $B_1(\kappa T - 1)$ also occurs with probability $1-\textup{o}(1)$. 
	Therefore, the least pulled arm, whether it is arm $1$ or arm $2$, will receive at least $\kappa T - 1$ with probability $1-\textup{o}(1)$.
	We record this result as 
	\begin{align}
		\p(\max\{n^1_T, n^2_T\} \leq T - \kappa T + 1)\geq 1-\textup{o}(1), 
	\end{align}
	and hence,
		\begin{align}
		\E(\max\{n^1_T, n^2_T\}) \leq  T - \kappa T + 1 + \textup{o}(T).\label{eq:maxPulls}
	\end{align}

	
	Then, using Lemma~\ref{lma:forUCB}, the expected reward under good event $G$, where policy $\pi$ is the UCB1 policy as defined in Expression~\ref{eq:UCB1define}, can be upper bounded as follows.
	\begin{align*}
		\E_{\bnu}\left[ \max \left\{ \sum_{t=1}^{n^1_T} U^1_t, \sum_{t=1}^{n^2_T} U^2_t \right\} \right] &\leq T \Delta + \mu_2(\bnu)~ \E_{\bnu} \left[ \max\{n^1_T,  n^2_T\} \right] + \textup{O}(\sqrt{T\log(T)})\\
		&\leq \sqrt{T} + \frac{1}{2} (T - \kappa T + 1+ \textup{o}(T)) + \textup{O}(\sqrt{T\log(T)})\\
		&\leq \frac{T}{2} - \frac{\kappa}{2}T + \textup{o}(T).
	\end{align*}
	The first inequality is due to Lemma~\ref{lma:forUCB} and the second inequality is due to Expression~\ref{eq:maxPulls}. Since the best expected reward is $(0.5 + \frac{1}{\sqrt{T}}) T = \frac{T}{2} + \sqrt{T}$, the expected regret is at least $\frac{\kappa}{2}T-\textup{o}(T)$.
	\hfill\Halmos
	
\subsection{Additional figures for reference}\label{apxsec:moreFigs}

{\bf General $m$ experiment with larger $T$ value for detailed comparison:} The example with $m=4$, $K=8$, and $\alpha = 0$ with a longer horizon is presented here to show that the performance gains of $m\textup{-UCB1}$ over $m\textup{-NADA-ETC}$ do not last long and the infrequent but constant sampling of suboptimal arms in $m\textup{-UCB1}$ drives regret higher as $T$ grows.

\begin{figure}[h]
	\centering
	\centering
	\includegraphics[width=0.65\textwidth]{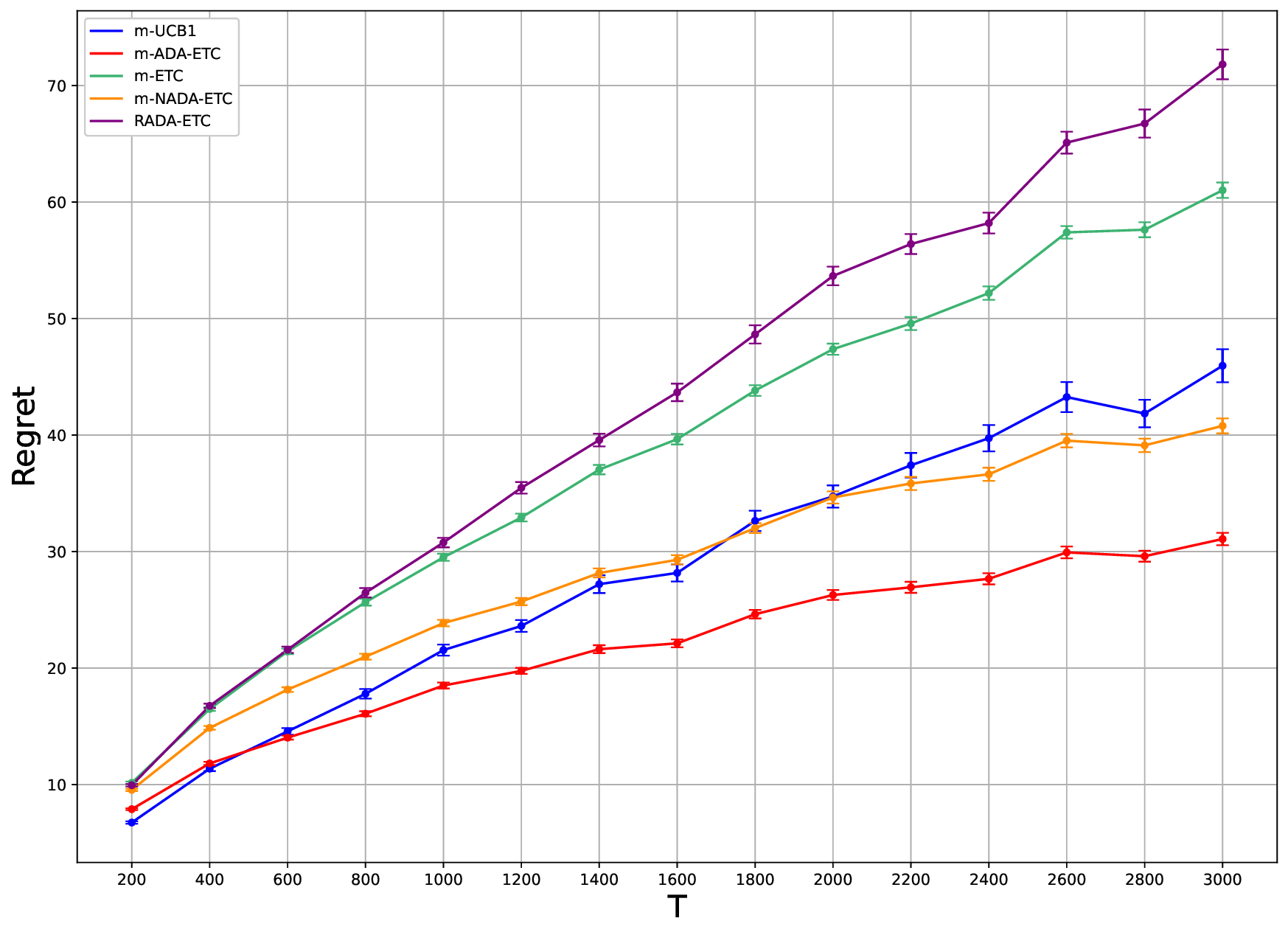}
	\caption{$m=4$, $K=8$, $\alpha = 0$}\label{fig:experiments_m_longer}
\end{figure}%

{\bf Product grooming application:} We present a figure with UCB1 included in here in an attempt to present the differences between algorithms that perform consistently well in the main paper. 

\begin{figure}[h]
		\centering
		\includegraphics[width=0.55\textwidth]{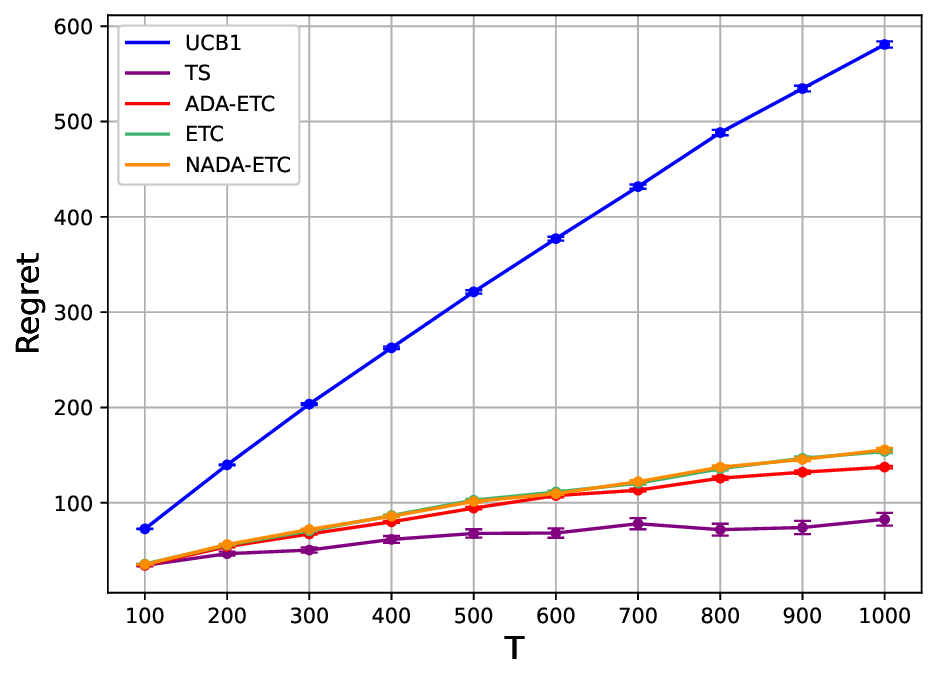}
		\caption{Snow shovel}
%
	
	 \label{fig:experiments_amazon_w_UCB1}
\end{figure}

\begin{table}[h]
	\parbox{.4\linewidth}{
		\caption{Dash Cams}
		\begin{tabular}{|c|c|c|c|c|c|c|}
			\hline
			Product & $1 \star$ & $2 \star$& $3 \star$ & $4 \star$& $5 \star$& Avg. Rating  \\
			\hline
			1 & 0.08 & 0.05 & 0.06 & 0.19 & 0.62 & 4.22 \\
			2 & 0.12 & 0.06 & 0.08 & 0.18 & 0.56 & 4.00 \\
			3 & 0.07 & 0.02 & 0.06 & 0.22 & 0.63 & 4.32 \\
			4 & 0.07 & 0.03 & 0.06 & 0.14 & 0.70 & 4.37 \\
			5 & 0.06 & 0.05 & 0.09 & 0.19 & 0.61 & 4.24 \\
			6 & 0.05 & 0.01 & 0.06 & 0.26 & 0.62 & 4.39 \\
			\hline
		\end{tabular}
		\label{table:1}
	}
\hspace{2.5cm}	
	\parbox{.4\linewidth}{
		\caption{Snow Shovels}
		\begin{tabular}{|c|c|c|c|c|c|c|}
			\hline
			Product & $1 \star$ & $2 \star$& $3 \star$ & $4 \star$& $5 \star$& Avg. Rating  \\
			\hline
			1 & 0.13 & 0.09 & 0.04 & 0.18 & 0.56 & 3.95 \\
			2 & 0.08 & 0.08 & 0.10 & 0.13 & 0.61 & 4.11 \\
			3 & 0.03 & 0.03 & 0.06 & 0.18 & 0.70 & 4.49 \\
			4 & 0.02 & 0.01 & 0.02 & 0.07 & 0.88 & 4.78 \\
			5 & 0.03 & 0.03 & 0.08 & 0.17 & 0.69 & 4.46 \\
			6 & 0.13 & 0.06 & 0.18 & 0.18 & 0.45 & 3.76 \\
			\hline
		\end{tabular}
		\label{table:2}
	}

\vspace{1cm}

\parbox{.4\linewidth}{
	\caption{Leaf Blowers}
	\begin{tabular}{|c|c|c|c|c|c|c|}
		\hline
		Product & $1 \star$ & $2 \star$& $3 \star$ & $4 \star$& $5 \star$& Avg. Rating  \\
		\hline
		1 & 0.03 & 0.02 & 0.07 & 0.19 & 0.69 & 4.49 \\
		2 & 0.06 & 0.06 & 0.08 & 0.19 & 0.61 & 4.23 \\
		3 & 0.15 & 0.07 & 0.10 & 0.19 & 0.49 & 3.80 \\
		4 & 0.10 & 0.04 & 0.07 & 0.15 & 0.64 & 4.19 \\
		5 & 0.07 & 0.03 & 0.08 & 0.17 & 0.65 & 4.30 \\
		6 & 0.12 & 0.07 & 0.10 & 0.19 & 0.52 & 3.92 \\
		\hline
	\end{tabular}
	\label{table:3}
}
\hspace{2.5cm}
\parbox{.4\linewidth}{
	\caption{Humidifiers}
	\begin{tabular}{|c|c|c|c|c|c|c|}
		\hline
		Product & $1 \star$ & $2 \star$& $3 \star$ & $4 \star$& $5 \star$& Avg. Rating  \\
		\hline
		1 & 0.05 & 0.03 & 0.05 & 0.15 & 0.72 & 4.46 \\
		2 & 0.04 & 0.02 & 0.05 & 0.13 & 0.76 & 4.55 \\
		3 & 0.15 & 0.06 & 0.08 & 0.14 & 0.57 & 3.92 \\
		4 & 0.08 & 0.04 & 0.08 & 0.16 & 0.64 & 4.24 \\
		5 & 0.09 & 0.03 & 0.05 & 0.12 & 0.61 & 4.33 \\
		6 & 0.11 & 0.03 & 0.07 & 0.14 & 0.65 & 4.19 \\
		\hline
	\end{tabular}
	\label{table:4}
}
\end{table}

\end{APPENDICES}

%% file: Bandit_Labor_Training.bbl
\begin{thebibliography}{}

\bibitem[Agrawal, 1995]{agrawal1995sample}
Agrawal, R. (1995).
\newblock Sample mean based index policies with {O}(log n) regret for the
  multi-armed bandit problem.
\newblock {\em Advances in Applied Probability}, pages 1054--1078.

\bibitem[Akbarpour et~al., 2020]{akbarpour2020thickness}
Akbarpour, M., Li, S., and Gharan, S.~O. (2020).
\newblock Thickness and information in dynamic matching markets.
\newblock {\em Journal of Political Economy}, 128(3):783--815.

\bibitem[Audibert and Bubeck, 2009]{audibertminimax}
Audibert, J.-Y. and Bubeck, S. (2009).
\newblock Minimax policies for adversarial and stochastic bandits.
\newblock In {\em COLT}, volume~7, pages 1--122.

\bibitem[Audibert and Bubeck, 2018]{audibertminimax2009}
Audibert, J.-Y. and Bubeck, S. (2018).
\newblock Minimax policies for adversarial and stochastic bandits.

\bibitem[Audibert et~al., 2010]{audibert2010best}
Audibert, J.-Y., Bubeck, S., and Munos, R. (2010).
\newblock Best arm identification in multi-armed bandits.
\newblock In {\em COLT}, pages 41--53.

\bibitem[Auer et~al., 2002]{auer2002finite}
Auer, P., Cesa-Bianchi, N., and Fischer, P. (2002).
\newblock Finite-time analysis of the multiarmed bandit problem.
\newblock {\em Machine learning}, 47(2-3):235--256.

\bibitem[Auer and Ortner, 2010]{auer2010ucb}
Auer, P. and Ortner, R. (2010).
\newblock Ucb revisited: Improved regret bounds for the stochastic multi-armed
  bandit problem.
\newblock {\em Periodica Mathematica Hungarica}, 61(1-2):55--65.

\bibitem[Baccara et~al., 2020]{baccara2020optimal}
Baccara, M., Lee, S., and Yariv, L. (2020).
\newblock Optimal dynamic matching.
\newblock {\em Theoretical Economics}, 15(3):1221--1278.

\bibitem[Baudry et~al., 2022]{baudry2022efficient}
Baudry, D., Russac, Y., and Kaufmann, E. (2022).
\newblock Efficient algorithms for extreme bandits.
\newblock {\em arXiv preprint arXiv:2203.10883}.

\bibitem[Bhatt et~al., 2022]{bhatt2022extreme}
Bhatt, S., Li, P., and Samorodnitsky, G. (2022).
\newblock Extreme bandits using robust statistics.
\newblock {\em IEEE Transactions on Information Theory}.

\bibitem[Bubeck and Cesa-Bianchi, 2012]{bubeck2012regret}
Bubeck, S. and Cesa-Bianchi, N. (2012).
\newblock Regret analysis of stochastic and nonstochastic multi-armed bandit
  problems.
\newblock {\em arXiv preprint arXiv:1204.5721}.

\bibitem[Bubeck et~al., 2009]{bubeck2009pure}
Bubeck, S., Munos, R., and Stoltz, G. (2009).
\newblock Pure exploration in multi-armed bandits problems.
\newblock In {\em International conference on Algorithmic learning theory},
  pages 23--37. Springer.

\bibitem[Bubeck et~al., 2011]{bubeck2011pure}
Bubeck, S., Munos, R., and Stoltz, G. (2011).
\newblock Pure exploration in finitely-armed and continuous-armed bandits.
\newblock {\em Theoretical Computer Science}, 412(19):1832--1852.

\bibitem[Bubeck et~al., 2013]{bubeck2013multiple}
Bubeck, S., Wang, T., and Viswanathan, N. (2013).
\newblock Multiple identifications in multi-armed bandits.
\newblock In {\em International Conference on Machine Learning}, pages
  258--265.

\bibitem[Carpentier and Locatelli, 2016]{carpentier2016tight}
Carpentier, A. and Locatelli, A. (2016).
\newblock Tight (lower) bounds for the fixed budget best arm identification
  bandit problem.
\newblock In {\em Conference on Learning Theory}, pages 590--604.

\bibitem[Carpentier and Valko, 2014]{carpentier2014extreme}
Carpentier, A. and Valko, M. (2014).
\newblock Extreme bandits.
\newblock In {\em Advances in Neural Information Processing Systems}, pages
  1089--1097.

\bibitem[Carpentier and Valko, 2015]{carpentier2015simple}
Carpentier, A. and Valko, M. (2015).
\newblock Simple regret for infinitely many armed bandits.
\newblock In {\em International Conference on Machine Learning}, pages
  1133--1141.

\bibitem[Cesa-Bianchi et~al., 2013]{cesa2013online}
Cesa-Bianchi, N., Dekel, O., and Shamir, O. (2013).
\newblock Online learning with switching costs and other adaptive adversaries.
\newblock In {\em Advances in Neural Information Processing Systems}, pages
  1160--1168.

\bibitem[Cicirello and Smith, 2005]{cicirello2005max}
Cicirello, V.~A. and Smith, S.~F. (2005).
\newblock The max k-armed bandit: A new model of exploration applied to search
  heuristic selection.
\newblock In {\em The Proceedings of the Twentieth National Conference on
  Artificial Intelligence}, volume~3, pages 1355--1361.

\bibitem[Dekel et~al., 2014]{dekel2014bandits}
Dekel, O., Ding, J., Koren, T., and Peres, Y. (2014).
\newblock Bandits with switching costs: {T$^{2/3}$} regret.
\newblock In {\em Proceedings of the forty-sixth annual ACM symposium on Theory
  of computing}, pages 459--467. ACM.

\bibitem[Donaker et~al., 2019]{luca2019designing}
Donaker, G., Kim, H., and Luca, M. (2019).
\newblock Designing better online review systems.
\newblock {\em Harvard Business Review}.

\bibitem[Esfandiari et~al., 2021]{esfandiari}
Esfandiari, H., Karbasi, A., Mehrabian, A., and Mirrokni, V. (2021).
\newblock Regret bounds for batched bandits.
\newblock In {\em Proceedings of the AAAI Conference on Artificial
  Intelligence}, volume~35, pages 7340--7348.

\bibitem[Even-Dar et~al., 2009]{even2009online}
Even-Dar, E., Kleinberg, R., Mannor, S., and Mansour, Y. (2009).
\newblock Online learning for global cost functions.
\newblock In {\em Conference on Learning Theory (COLT)}.

\bibitem[Even-Dar et~al., 2002]{even2002pac}
Even-Dar, E., Mannor, S., and Mansour, Y. (2002).
\newblock Pac bounds for multi-armed bandit and markov decision processes.
\newblock In {\em International Conference on Computational Learning Theory},
  pages 255--270. Springer.

\bibitem[Even-Dar et~al., 2006]{even2006action}
Even-Dar, E., Mannor, S., and Mansour, Y. (2006).
\newblock Action elimination and stopping conditions for the multi-armed bandit
  and reinforcement learning problems.
\newblock {\em Journal of machine learning research}, 7(Jun):1079--1105.

\bibitem[Even-Dar et~al., 2010]{even2010learning}
Even-Dar, E., Mannor, S., and Mansour, Y. (2010).
\newblock Learning with global cost in stochastic environments.

\bibitem[Gao et~al., 2019]{gao2019batched}
Gao, Z., Han, Y., Ren, Z., and Zhou, Z. (2019).
\newblock Batched multi-armed bandits problem.
\newblock {\em arXiv preprint arXiv:1904.01763}.

\bibitem[Garivier and Capp{\'e}, 2011]{garivier2011kl}
Garivier, A. and Capp{\'e}, O. (2011).
\newblock The kl-ucb algorithm for bounded stochastic bandits and beyond.
\newblock In {\em Proceedings of the 24th annual conference on learning
  theory}, pages 359--376. JMLR Workshop and Conference Proceedings.

\bibitem[Garivier et~al., 2016]{garivier2016explore}
Garivier, A., Lattimore, T., and Kaufmann, E. (2016).
\newblock On explore-then-commit strategies.
\newblock {\em Advances in Neural Information Processing Systems}, 29.

\bibitem[Gittins et~al., 2011]{gittins2011multi}
Gittins, J., Glazebrook, K., and Weber, R. (2011).
\newblock {\em Multi-armed bandit allocation indices}.
\newblock John Wiley \& Sons.

\bibitem[Gourville and Soman, 2005]{gourville2005overchoice}
Gourville, J.~T. and Soman, D. (2005).
\newblock Overchoice and assortment type: When and why variety backfires.
\newblock {\em Marketing science}, 24(3):382--395.

\bibitem[Hsu et~al., 2021]{hsu2021integrated}
Hsu, W.-K., Xu, J., Lin, X., and Bell, M.~R. (2021).
\newblock Integrated online learning and adaptive control in queueing systems
  with uncertain payoffs.
\newblock {\em Operations Research}.

\bibitem[Jamieson et~al., 2014]{jamieson2014lil}
Jamieson, K., Malloy, M., Nowak, R., and Bubeck, S. (2014).
\newblock lil'ucb: An optimal exploration algorithm for multi-armed bandits.
\newblock In {\em Conference on Learning Theory}, pages 423--439.

\bibitem[Jin et~al., 2019]{jin2019efficient}
Jin, T., Shi, J., Xiao, X., and Chen, E. (2019).
\newblock Efficient pure exploration in adaptive round model.
\newblock {\em Advances in Neural Information Processing Systems}, 32.

\bibitem[Jin et~al., 2021]{jin2021double}
Jin, T., Xu, P., Xiao, X., and Gu, Q. (2021).
\newblock Double explore-then-commit: Asymptotic optimality and beyond.
\newblock In {\em Conference on Learning Theory}, pages 2584--2633. PMLR.

\bibitem[Johari et~al., 2021]{johari2021matching}
Johari, R., Kamble, V., and Kanoria, Y. (2021).
\newblock Matching while learning.
\newblock {\em Operations Research}, 69(2):655--681.

\bibitem[Jun et~al., 2016]{jun2016top}
Jun, K.-S., Jamieson, K., Nowak, R., and Zhu, X. (2016).
\newblock Top arm identification in multi-armed bandits with batch arm pulls.
\newblock In {\em Artificial Intelligence and Statistics}, pages 139--148.
  PMLR.

\bibitem[Kalyanakrishnan et~al., 2012]{kalyanakrishnan2012pac}
Kalyanakrishnan, S., Tewari, A., Auer, P., and Stone, P. (2012).
\newblock Pac subset selection in stochastic multi-armed bandits.
\newblock In {\em Proceedings of the 29th International Conference on Machine
  Learning}, pages 227--234.

\bibitem[Kamble and Ozbay, 2022]{kamble2022exploration}
Kamble, V. and Ozbay, E. (2022).
\newblock Exploration in markets under local congestion-based pricing.
\newblock {\em Available at SSRN 4041075}.

\bibitem[Karnin et~al., 2013]{karnin2013almost}
Karnin, Z., Koren, T., and Somekh, O. (2013).
\newblock Almost optimal exploration in multi-armed bandits.
\newblock In {\em International Conference on Machine Learning}, pages
  1238--1246.

\bibitem[Kaufmann et~al., 2016]{kaufmann2016complexity}
Kaufmann, E., Capp{\'e}, O., and Garivier, A. (2016).
\newblock On the complexity of best-arm identification in multi-armed bandit
  models.
\newblock {\em The Journal of Machine Learning Research}, 17(1):1--42.

\bibitem[Kaufmann and Kalyanakrishnan, 2013]{kaufmann2013information}
Kaufmann, E. and Kalyanakrishnan, S. (2013).
\newblock Information complexity in bandit subset selection.
\newblock In {\em Conference on Learning Theory}, pages 228--251.

\bibitem[Lai and Robbins, 1985]{lai1985asymptotically}
Lai, T.~L. and Robbins, H. (1985).
\newblock Asymptotically efficient adaptive allocation rules.
\newblock {\em Advances in applied mathematics}, 6(1):4--22.

\bibitem[Lattimore, 2018]{lattimore2018refining}
Lattimore, T. (2018).
\newblock Refining the confidence level for optimistic bandit strategies.
\newblock {\em The Journal of Machine Learning Research}, 19(1):765--796.

\bibitem[Lattimore and Szepesv{\'a}ri, 2020]{lattimore2018bandit}
Lattimore, T. and Szepesv{\'a}ri, C. (2020).
\newblock {\em Bandit algorithms}.
\newblock Cambridge University Press.

\bibitem[Loertscher et~al., 2022]{loertscher2022optimal}
Loertscher, S., Muir, E.~V., and Taylor, P.~G. (2022).
\newblock Optimal market thickness.
\newblock {\em Journal of Economic Theory}, 200:105383.

\bibitem[Luca, 2017]{luca2017designing}
Luca, M. (2017).
\newblock Designing online marketplaces: Trust and reputation mechanisms.
\newblock {\em Innovation Policy and the Economy}, 17(1):77--93.

\bibitem[Mannor and Tsitsiklis, 2004]{mannor2004sample}
Mannor, S. and Tsitsiklis, J.~N. (2004).
\newblock The sample complexity of exploration in the multi-armed bandit
  problem.
\newblock {\em Journal of Machine Learning Research}, 5(Jun):623--648.

\bibitem[Massouli{\'e} and Xu, 2018]{massoulie2018capacity}
Massouli{\'e}, L. and Xu, K. (2018).
\newblock On the capacity of information processing systems.
\newblock {\em Operations Research}, 66(2):568--586.

\bibitem[Perchet and Rigollet, 2013]{perchet2013multi}
Perchet, V. and Rigollet, P. (2013).
\newblock The multi-armed bandit problem with covariates.
\newblock {\em The Annals of Statistics}, 41(2):693--721.

\bibitem[Perchet et~al., 2016]{perchet2016batched}
Perchet, V., Rigollet, P., Chassang, S., and Snowberg, E. (2016).
\newblock Batched bandit problems.
\newblock {\em Annals of Statistics}, 44(2):660--681.

\bibitem[Russo et~al., 2018]{russo2018tutorial}
Russo, D.~J., Van~Roy, B., Kazerouni, A., Osband, I., Wen, Z., et~al. (2018).
\newblock A tutorial on thompson sampling.
\newblock {\em Foundations and Trends{\textregistered} in Machine Learning},
  11(1):1--96.

\bibitem[Settle and Golden, 1974]{settle1974consumer}
Settle, R.~B. and Golden, L.~L. (1974).
\newblock Consumer perceptions: Overchoice in the market place.
\newblock {\em ACR North American Advances}.

\bibitem[Shah et~al., 2020]{shah2020adaptive}
Shah, V., Gulikers, L., Massouli{\'e}, L., and Vojnovi{\'c}, M. (2020).
\newblock Adaptive matching for expert systems with uncertain task types.
\newblock {\em Operations Research}, 68(5):1403--1424.

\bibitem[Slivkins, 2019]{slivkins2019introduction}
Slivkins, A. (2019).
\newblock Introduction to multi-armed bandits.
\newblock {\em arXiv preprint arXiv:1904.07272}.

\bibitem[Streeter and Smith, 2006a]{streeter2006asymptotically}
Streeter, M.~J. and Smith, S.~F. (2006a).
\newblock An asymptotically optimal algorithm for the max k-armed bandit
  problem.
\newblock In {\em AAAI}, pages 135--142.

\bibitem[Streeter and Smith, 2006b]{streeter2006simple}
Streeter, M.~J. and Smith, S.~F. (2006b).
\newblock A simple distribution-free approach to the max k-armed bandit
  problem.
\newblock In {\em International Conference on Principles and Practice of
  Constraint Programming}, pages 560--574. Springer.

\bibitem[Sun and Zhao, 2022]{suncongestion}
Sun, X. and Zhao, J. (2022).
\newblock Congestion-aware matching and learning for service platforms.

\bibitem[Thompson, 1933]{thompson1933likelihood}
Thompson, W.~R. (1933).
\newblock On the likelihood that one unknown probability exceeds another in
  view of the evidence of two samples.
\newblock {\em Biometrika}, 25(3/4):285--294.

\bibitem[Vaidhiyan and Sundaresan, 2017]{vaidhiyan2017learning}
Vaidhiyan, N.~K. and Sundaresan, R. (2017).
\newblock Learning to detect an oddball target.
\newblock {\em IEEE Transactions on Information Theory}, 64(2):831--852.

\bibitem[Villar et~al., 2015]{villar2015multi}
Villar, S.~S., Bowden, J., and Wason, J. (2015).
\newblock Multi-armed bandit models for the optimal design of clinical trials:
  benefits and challenges.
\newblock {\em Statistical science: a review journal of the Institute of
  Mathematical Statistics}, 30(2):199.

\bibitem[Zhou et~al., 2014]{zhou2014optimal}
Zhou, Y., Chen, X., and Li, J. (2014).
\newblock Optimal pac multiple arm identification with applications to
  crowdsourcing.
\newblock In {\em International Conference on Machine Learning}, pages
  217--225.

\end{thebibliography}
